\def\year{2022}\relax
\documentclass[letterpaper]{article} 
\usepackage{aaai22}  
\usepackage{times}  
\usepackage{helvet}  
\usepackage{courier}  
\usepackage[hyphens]{url}  
\usepackage{graphicx} 
\usepackage{natbib}  
\usepackage{caption} 
\DeclareCaptionStyle{ruled}{labelfont=normalfont,labelsep=colon,strut=off} 
\usepackage{algorithm}
\usepackage{algorithmic}
\usepackage{subcaption}
\usepackage{multirow}
\usepackage{xcolor}
\usepackage{amssymb}
\usepackage[percent]{overpic}

\usepackage{newfloat}
\usepackage{listings}
\floatstyle{ruled}
\newfloat{listing}{tb}{lst}{}
\floatname{listing}{Listing}
\title{SelfReformer: Self-Refined Network with Transformer for Salient Object Detection}
\author {
    Yi Ke Yun,
    Weisi Lin\thanks{Corresponding author.}
}
\affiliations {
    School of Computer Science and Engineering, Nanyang Technological University, Singapore\\
    yunyikeyyk@gmail.com, wslin@ntu.edu.sg
}

\usepackage{bibentry}

\begin{document}
\frenchspacing
\maketitle

\begin{abstract}
The global and local contexts significantly contribute to the integrity of predictions in Salient Object Detection (SOD). Unfortunately, existing methods still struggle to generate complete predictions with fine details. There are two major problems in conventional approaches: first, for global context, CNN-based encoders cannot effectively catch long-range dependencies, resulting in incomplete predictions. Second, downsampling the ground truth to fit the size of predictions will introduce inaccuracy as the ground truth details are lost during interpolation or pooling. To address the abovementioned problems, we employed a Transformer as our encoder backbone for better long-range dependency modeling.
Meanwhile, we developed a branch and framed a patch-wise SOD task to learn the global context instead of assuming they are the high-level features in the encoder. Besides, for better details, we adopt Pixel Shuffle from Super-Resolution (SR) to reshape the predictions in each decoder stage back to the size of ground truth instead of the reverse. Furthermore, we developed a Context Refinement Module (CRM) to fuse global context with decoder features and automatically locate and refine the local details. The proposed network can guide and correct itself based on the global and local context generated (Fig.\ref{HIGHLIGHT}), thus is named, Self-Refined Transformer (SelfReformer). Extensive experiments and evaluation results on five benchmark datasets demonstrate the outstanding performance of the network, and we achieved the state-of-the-art. Code will be released at https://github.com/BarCodeReader/SelfReformer.
\end{abstract}

\section{Introduction}
SOD aims to locate and segment the object that catches human attention in a visual scene. Due to its wide applications, such as AR/VR \cite{U2NET} and image captioning \cite{IMCAP1, IMCAP2}, it has gained growing interest in recent years. Most of the state-of-the-art models are CNN-based and often have an architecture of encoder-decoder where images are firstly encoded into multi-level features, followed by a decoder for feature fusion and saliency prediction. To further improve the accuracy, most of the work tries to develop better fusion modules \cite{F3NET, MINET}, extra refinement networks \cite{BASNET, U2NET}, utilizing different modalities like depth or contour \cite{SSF, EGNET}, and adopting attention modules \cite{PAGE, PAGRN}. These methods achieved remarkable results in the SOD task. However, CNN-based networks are limited in learning long-range relationships, resulting in a lack of global structural consistency in predictions. 


\begin{figure}[!t]
\centering
\begin{overpic}[width=0.32\linewidth]{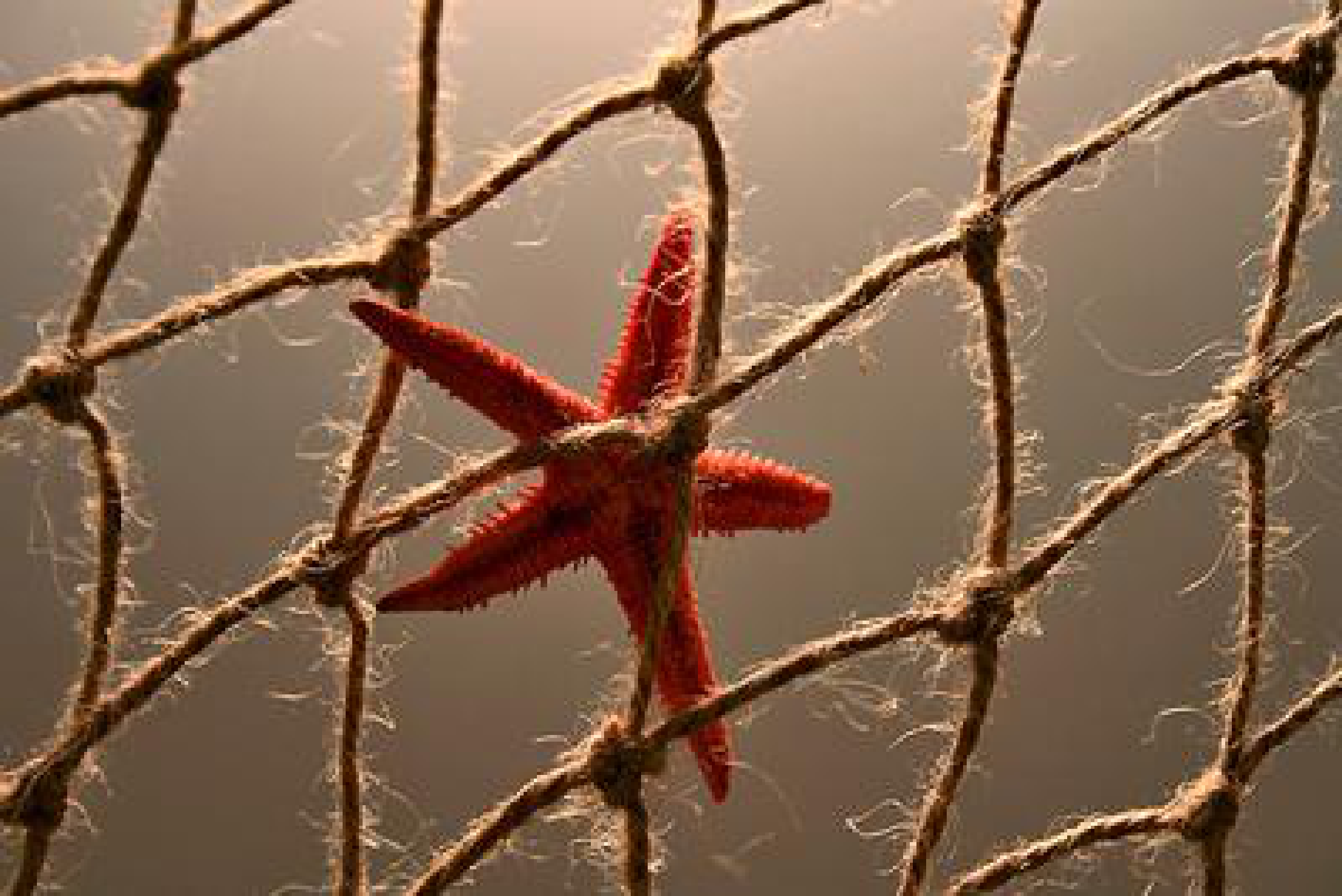}
 \put (85,55) {\textcolor{white}{(a)}}
\end{overpic}
\begin{overpic}[width=0.32\linewidth]{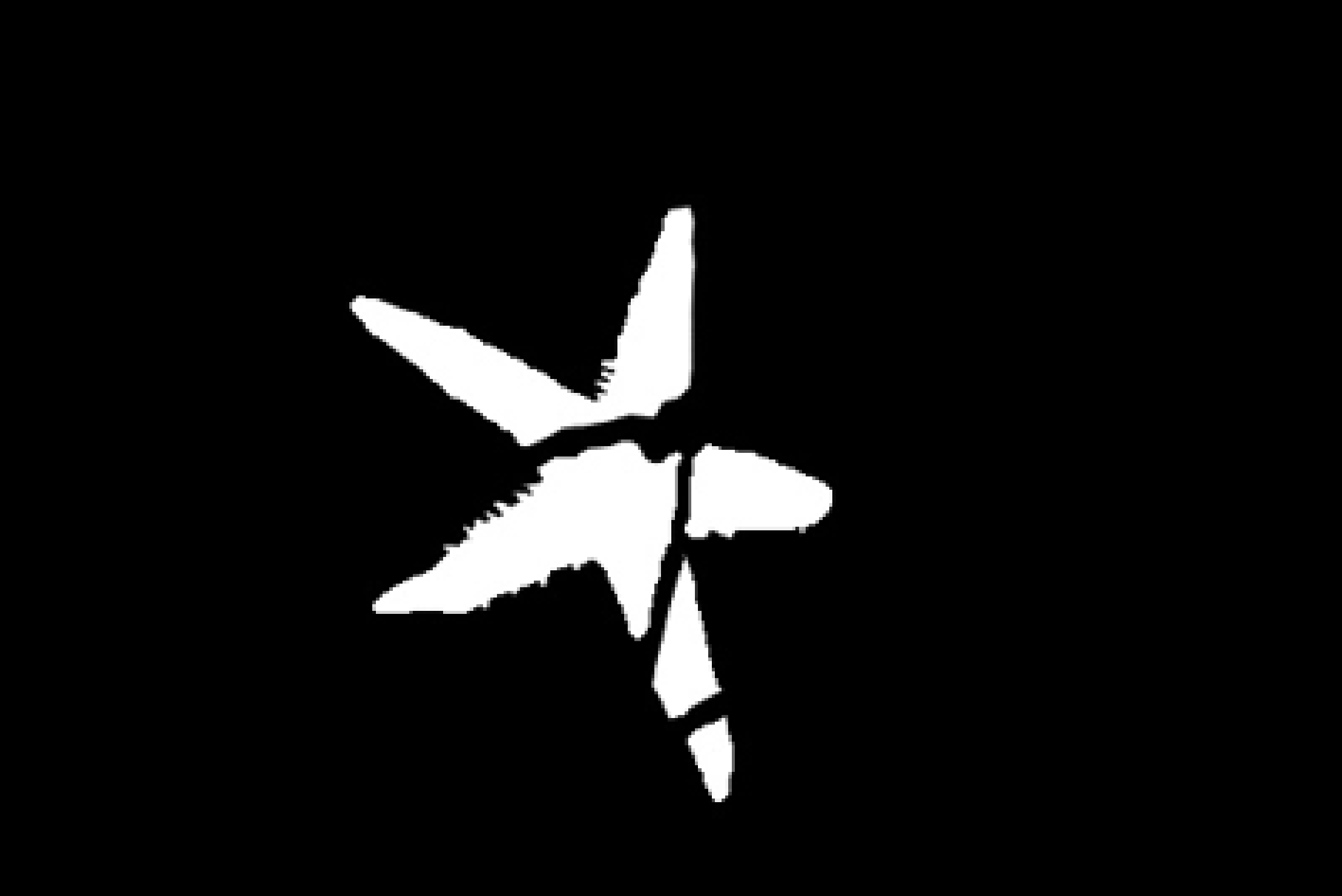}
 \put (85,55) {\textcolor{white}{(b)}}
\end{overpic}
\begin{overpic}[width=0.32\linewidth]{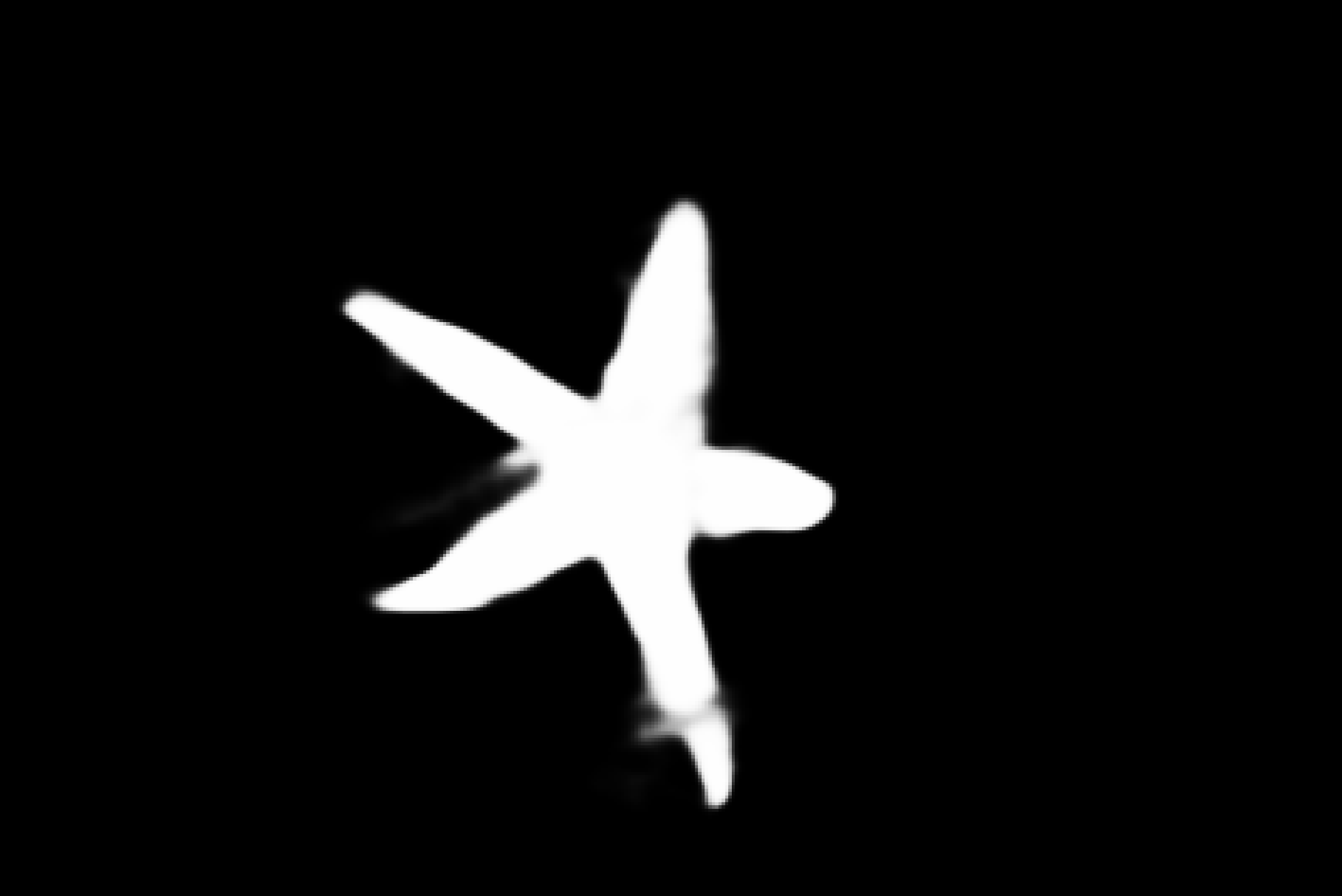}
 \put (85,55) {\textcolor{white}{(c)}}
\end{overpic}
\begin{overpic}[width=0.32\linewidth]{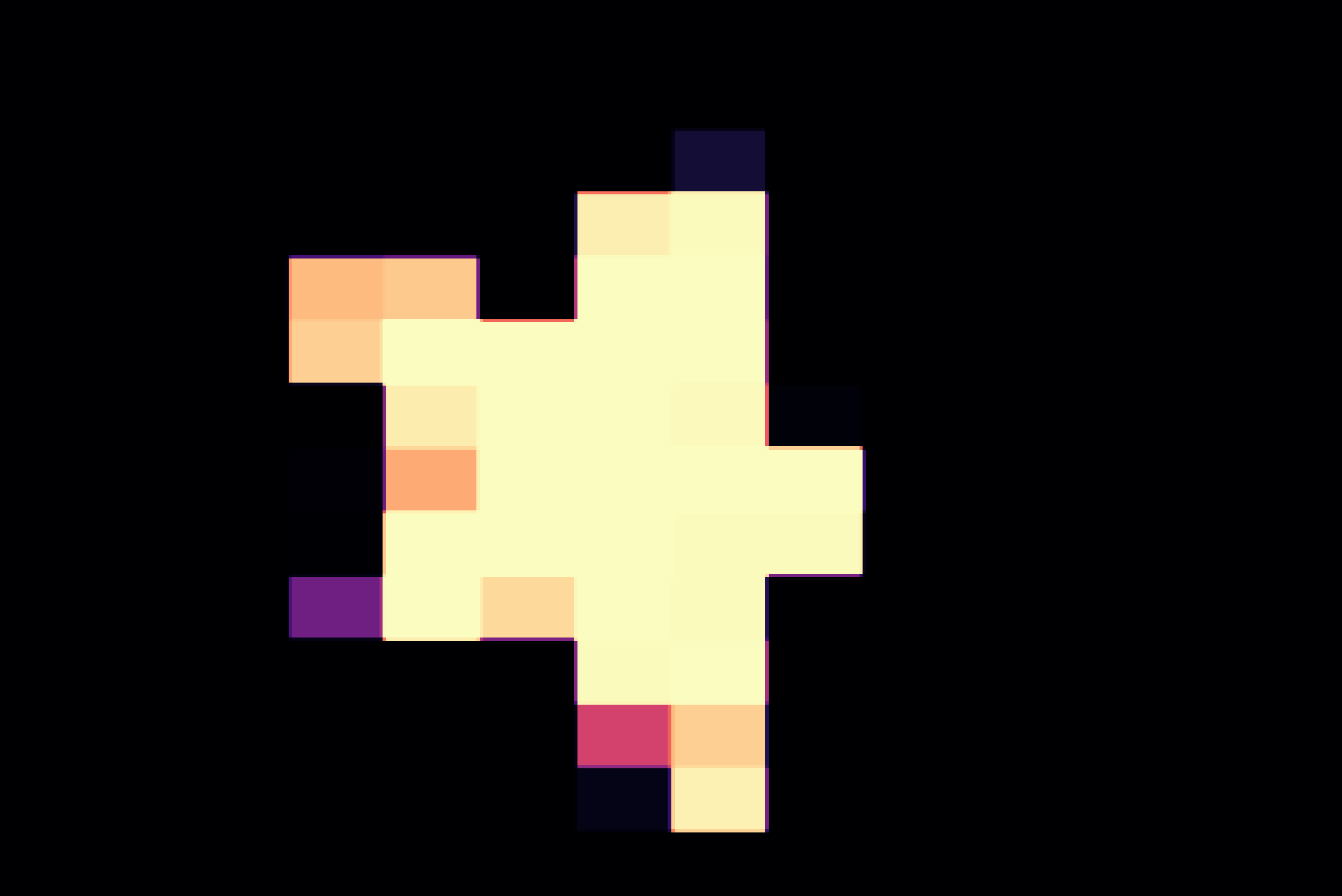}
 \put (85,55) {\textcolor{white}{(d)}}
\end{overpic}
\begin{overpic}[width=0.32\linewidth]{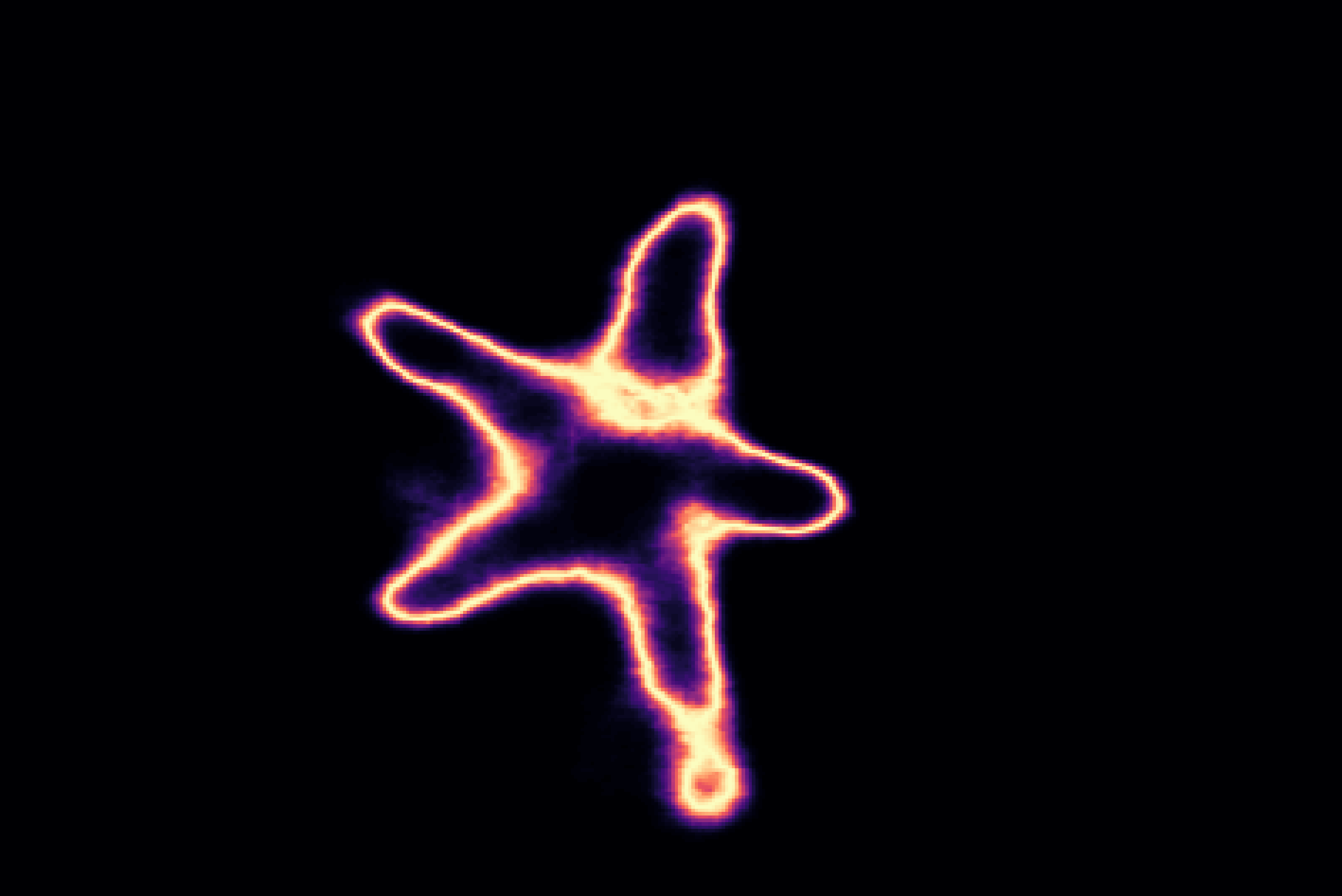}
 \put (85,55) {\textcolor{white}{(e)}}
\end{overpic}
\begin{overpic}[width=0.32\linewidth]{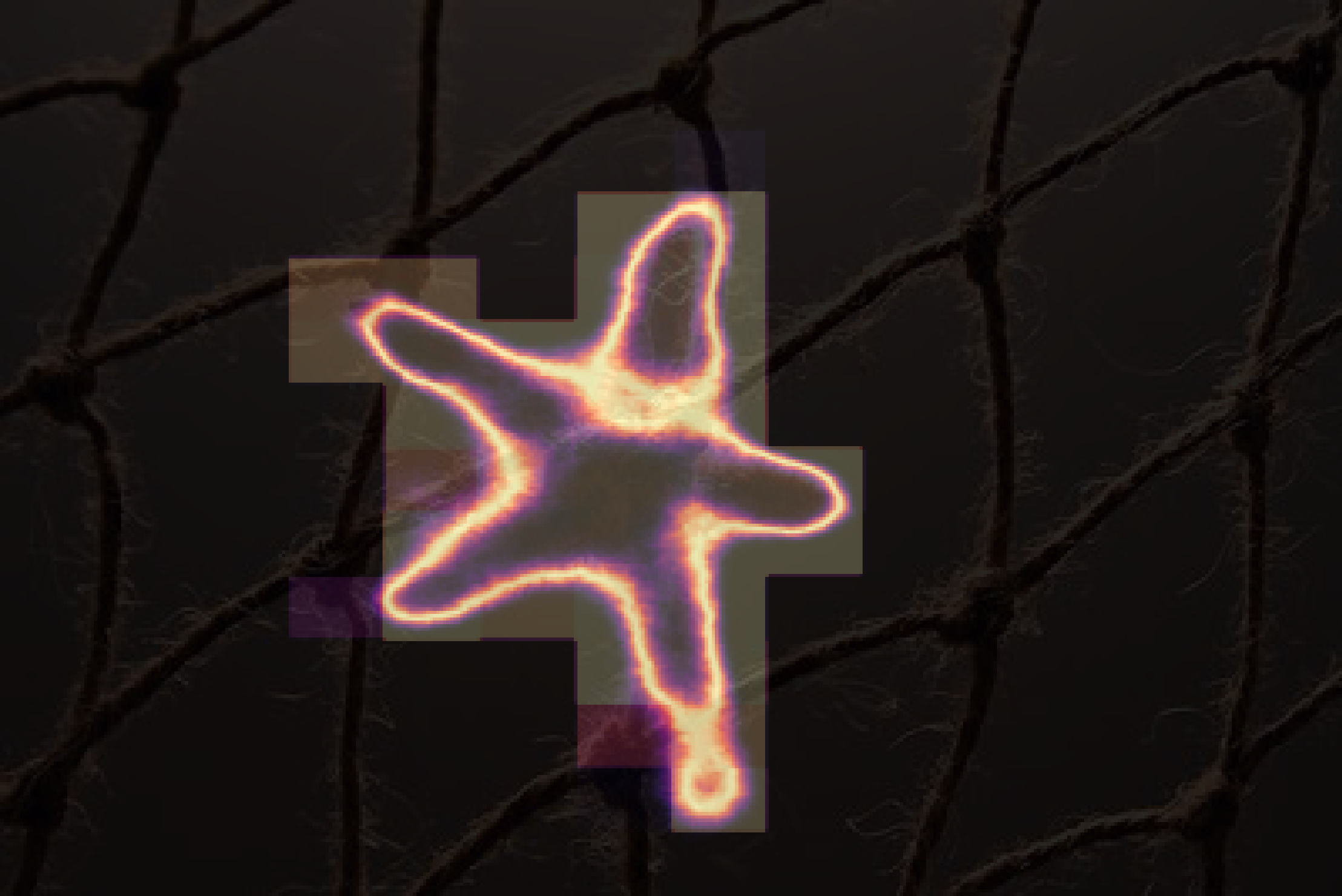}
 \put (85,55) {\textcolor{white}{(f)}}
\end{overpic}
\caption{Visualization of global and local context obtained by our network. (a) Input, (b) Ground truth, (c) Prediction, (d) Patch-wise global context map, (e) Local context map, (f) Illustration of how global and local context benefits the prediction. Best view in color.}
\label{HIGHLIGHT}
\vspace{-0.03\linewidth}
\end{figure}
In recent years, Transformer \cite{ATTN} was proposed to model long-range dependencies in language processing and was further extended to vision tasks. The vision transformers (ViT) \cite{T2TVIT, TWINS, PVT} split the image into patches then apply multi-head self-attention and multi-layer perceptrons to capture long-range dependencies. When applied to SOD, the transformer-based networks \cite{VST, GLSTR} are effective in modeling global context, thus generating predictions with better structural integrity.

However, there are still two big challenges for better SOD. First, SOD is a densely supervised task that requires the ground truth in different resolutions for each decoder stage. Using interpolation or pooling, fine features in the ground truth are lost, and the decoder is trained against inaccurate ground truths, resulting in poor details in predictions. Noteworthily, for input size of $224\times224$, transformers like T2T-ViT \cite{T2TVIT} and PVT \cite{PVT} usually have much smaller size of feature maps ($56\times56$ max), and how to restore the fine features from this small size for accurate SOD still remain unsolved. Second, though existing studies utilizing global and local contexts like feedback network \cite{F3NET} and multi-level fusion \cite{MINET}, the concept of global and local context is still intuitive as we do not have a method to obtain and qualify them explicitly. Thus, finding a better representation of global and local contexts and obtaining them in a controllable manner is still an open question.

\begin{figure}[!t]
\centering
\includegraphics[width=0.9\linewidth]{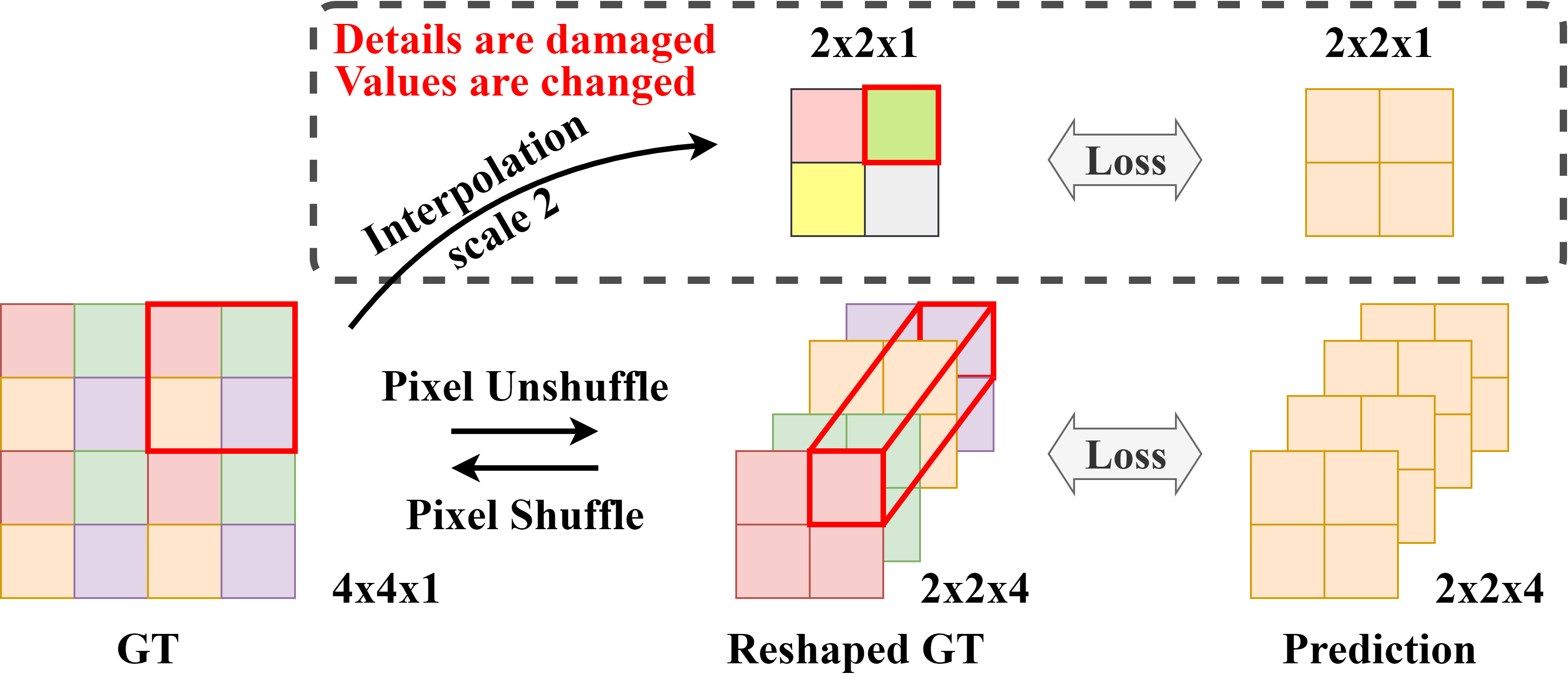} 
\caption{Illustration of conventional approach (grey box) and the Pixel Shuffle (PS). Details in the GT is preserved after PS but lost using interpolation. Best view in color.}
\label{the_ps}
\vspace{-0.03\linewidth}
\end{figure}

We address the abovementioned problems from three aspects. First, to preserve the structural properties of ground truth, we adopt Pixel Shuffle (PS) \cite{ESPCN} from Super-Resolution (SR) as the up/downsampling method. Unlike pooling or interpolation, Pixel Shuffle can reshape a high resolution (HR) image into groups of stacked low resolution (LR) images without changing the pixel values (Fig.\ref{the_ps}). Thus each decoder stage will have the same ground truth instead of multiple inconsistent LR images.
Second, to obtain the global context more precisely, we reframe the SOD task into a patch-wise saliency detection problem and supervise a branch to learn the information explicitly (Fig.\ref{HIGHLIGHT}d). We split input images into $n\times n$ non-overlapping patches whereby the developed branch identifies which patch contains saliency. Compared with existing approaches where global contexts are assumed to be the high-level encoder features, in our work, the obtained global context is learned via a supervised task.
Lastly, we developed a Context Refinement Module (CRM) to fuse global context features and refine local unconfident regions. The CRM will firstly fuse global contexts with decoder features for better detection completeness and generate a prediction. Then based on the unsure regions in the prediction, a local context feature map (Fig.\ref{HIGHLIGHT}e) is generated to guide the network for fine structure segmentation. Thus the CRM is a two-stage module where the final predictions with a better quality were obtained based on the refinement map generated from its first predictions.

To sum up, our contributions are as follows:
\begin{itemize}
\item We proposed an end-to-end Transformer-based network equipped with a global and local context branch for better structural integrity and local details in predictions.
\item We demonstrated that Pixel Shuffle yields better training results than interpolation and pooling methods in preserving fine structures. This is the first work that applied Pixel Shuffle on the SOD task to our best knowledge.
\item We introduced a measurable method to obtain global-context by framing a supervised patch-wise SOD. For local-context, we developed a CRM to automatically locate and refine unsure regions for better details.
\end{itemize}
\section{Related Work}
\subsection{Global and Local Context Fusion}
In encoder-decoder structure, features in deeper layers have a global view but local details are diluted because of too many convolution operations and resolution reduction, while shallower layers contain more local contexts. Many studies in SOD demonstrated that global and local context fusion can boost model performance. In SCNet \cite{SCNET}, Hou et al. showed that global context is capable of locating the salient object while local context is for preserving details. Thus they introduced short-connection to fuse global and local information in deeper and shallower layers. Similarly, PoolNet \cite{POOLNET} introduced a pyramid pooling module to further capture the global semantic information from the encoder, followed by a feature aggregation module to recover the diluted information in the encoder. In PFSNet \cite{PFSNET}, Ma et al. developed a pyramid feature shrinking module to fuse local and global features progressively. Compared with conventional encoder-decoders, where deeper stages are only fused with the next shallower stage, the pyramidal mechanism gradually fuses features across all stages, resulting in better global and local context fusion. Recently, in PA-KRN \cite{PAKRN}, Xu et al. proposed a two-stage model by firstly generating a coarse global context map to locate the salient object, followed by an attention-based sampler to zoom in the target region, and lastly refining the details by fusing local context features from encoders. It has been proved that context fusion will improve the predictions.

\subsection{Vision Transformers}
Transformers were firstly introduced in natural language processing \cite{ATTN, BERT, ALBERT} and were extended to computer vision tasks such as image classification \cite{16X16} and semantic segmentation \cite{SETR} due to their capability of modeling long-range dependencies. Networks like DETR \cite{DETR} and its variants \cite{UPDETR, HOTR} used a combination of CNN and Transformer for various computer vision tasks \cite{TMT, PAT}. Following the Visual Transformer's (ViT) success in image classification, some studies extend the Transformer for dense prediction tasks, e.g., semantic segmentation or depth estimation. SETR \cite{SETR} and PVT \cite{PVT} employ ViT as the encoder and use several convolutional layers to upsample encoder features for dense prediction. In SOD, VST \cite{VST} adopted T2T-ViT \cite{T2TVIT} as the backbone and achieved remarkable results. The input images were unfolded into partially overlapped patches for self-attention, and a reverse T2T (rT2T) mechanism was developed to reconstruct the predictions gradually. The effectiveness of self-attention in modeling long-range dependencies makes Transformer promising in SOD tasks.

\subsection{Pixel Shuffle}
Pixel Shuffle \cite{PS} was originally applied in the task of Single Image Super-Resolution (SISR) to upscale a low-resolution (LR) image $r$ times into a high-resolution (HR) image, and its reverse operation is Pixel-unshuffle (Fig.\ref{the_ps}). Different from interpolation methods, by reshaping the input image from $[r^{2}C, H, W]$ to $[C, rH, rW]$, one can obtain an HR image without changing any pixel values. In SISR, the primary purpose of Pixel Shuffle is to keep the feature map at a small size to achieve a higher inference speed. Another advantage of Pixel Shuffle compared with interpolation or pooling is that we can unshuffle an HR image to LR without losing any structural details since pixels are relocated and values are unchanged. This is useful for encoder-decoder networks as the decoder is usually densely supervised against resized ground truth. When we increase the downsampling scale using interpolation or pooling methods, more and more details are lost (Fig.\ref{PS_COMPARE}b-d); consequently, the decoder is trained against inaccurate ground truths. The shuffle and unshuffle operations ensure that details in HR ground truth remain in its LR form, regardless of scaling factors. In this work, for simplicity, we use the term \textit{Pixel Shuffle} to represent both shuffle and unshuffle operations in between layers with different scales.

\section{Proposed Methods}
\subsection{Overall Architecture}
We adopt PVT as the encoder backbone for better modeling of long-range dependencies. In order to preserve fine structures, Pixel Shuffle is applied across all stages when feature maps need to be scaled. To obtain global context information, we framed a new task by predicting whether a patch of the ground truth contains saliency and developed the global context branch. Sharing the same encoder, the Context Refinement Module (CRM) is developed to fuse global contexts with decoder features and refine local details in the prediction. The network architecture is shown in Fig.\ref{fig:network}.

\subsection{Pyramid Vision Transformer (PVT) Backbone}
\begin{figure}[!t]
\centering
\begin{overpic}[width=0.24\linewidth]{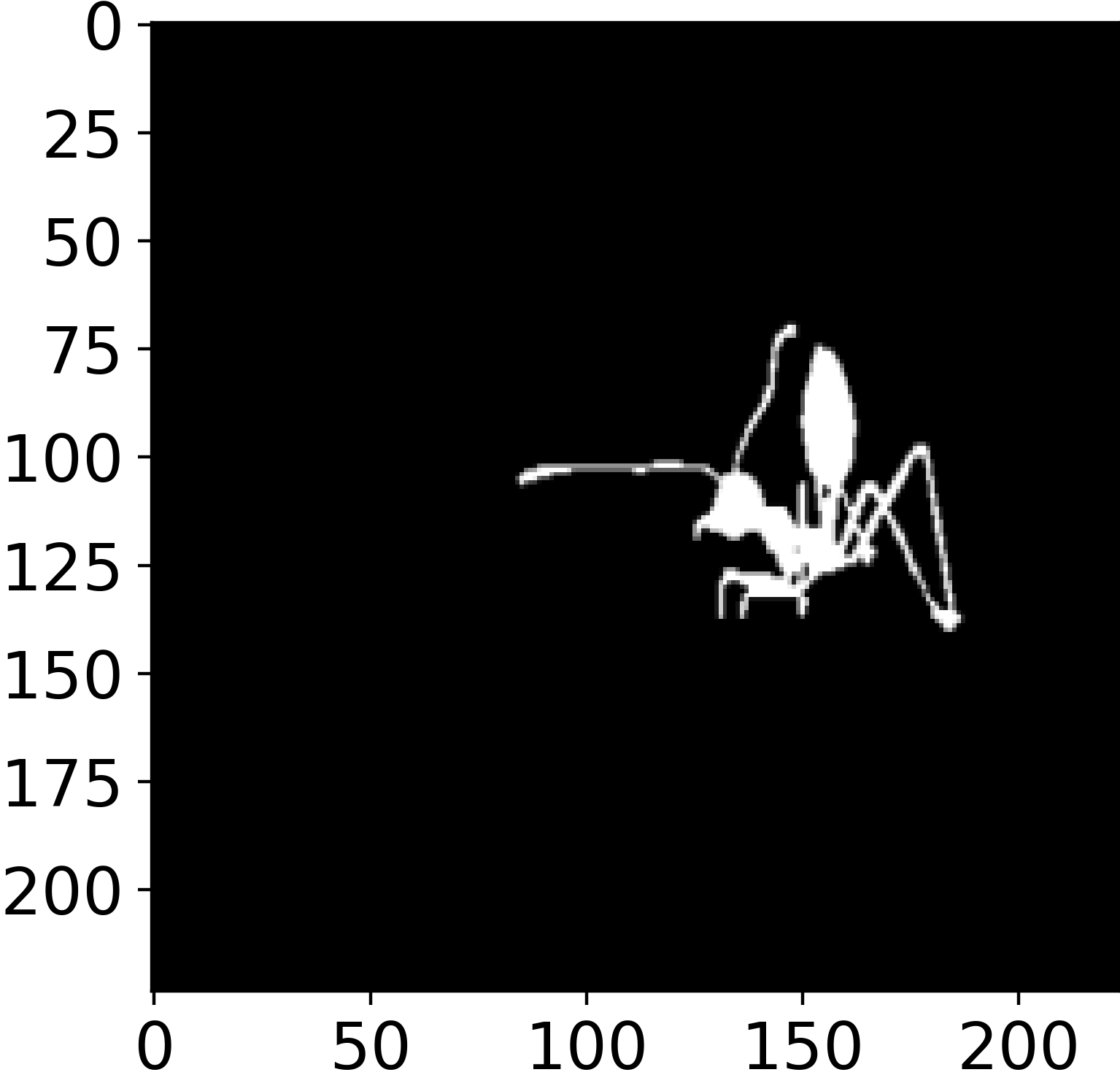}
\put (80,82) {\textcolor{white}{(a)}}
\end{overpic}
\begin{overpic}[width=0.24\linewidth]{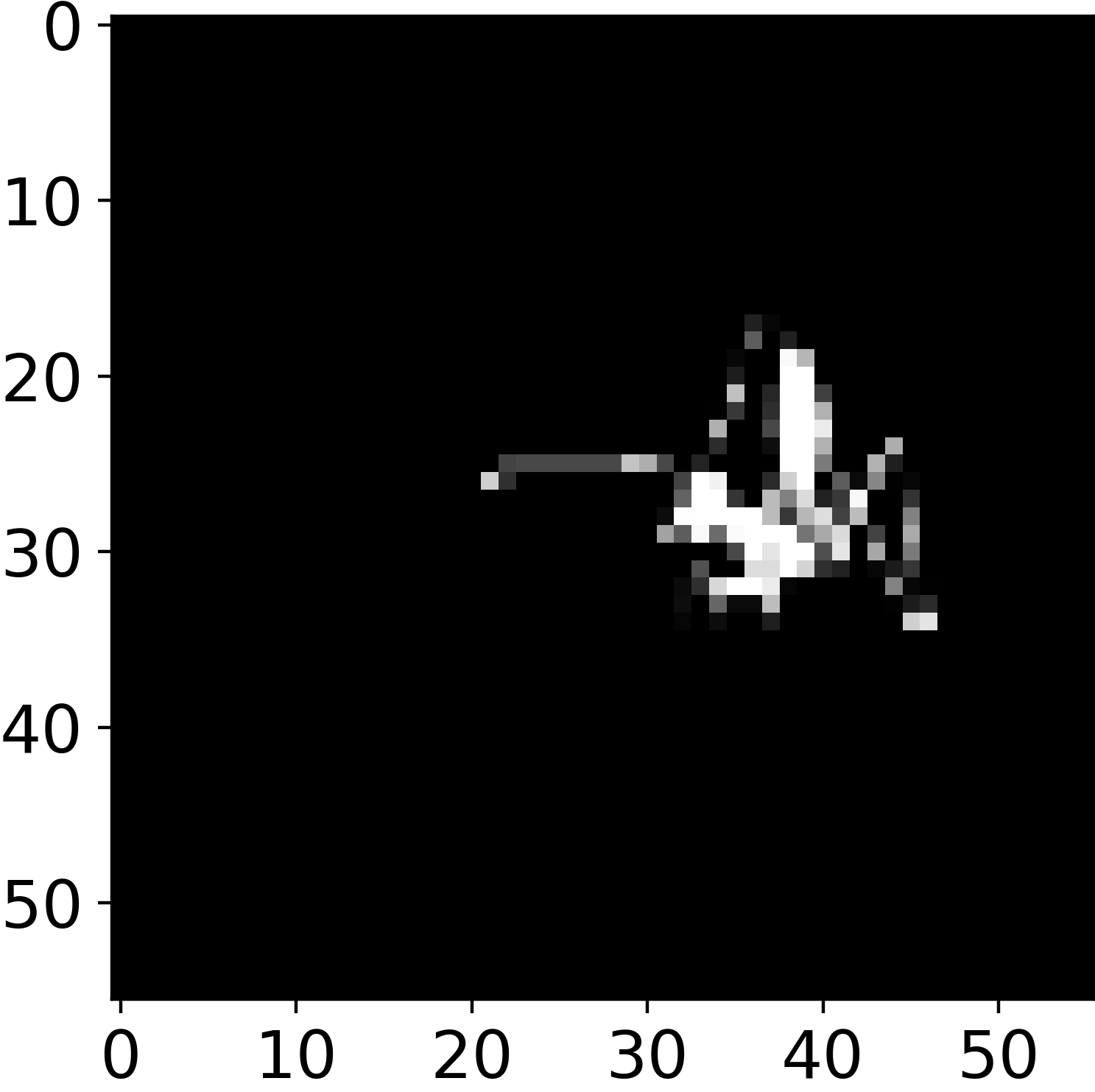}
\put (80,82) {\textcolor{white}{(b)}}
\end{overpic}
\begin{overpic}[width=0.24\linewidth]{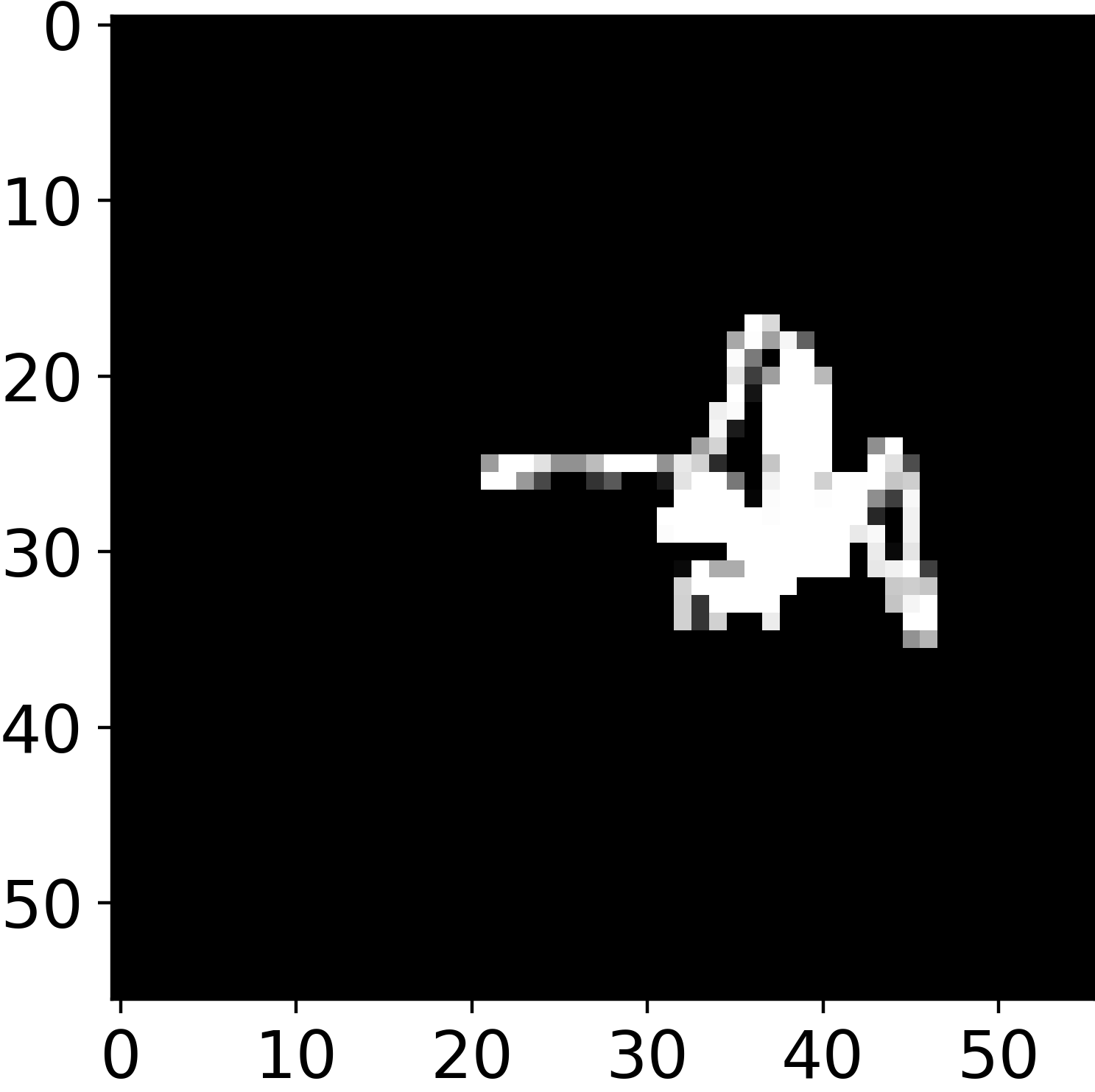}
\put (80,82) {\textcolor{white}{(c)}}
\end{overpic}
\begin{overpic}[width=0.24\linewidth]{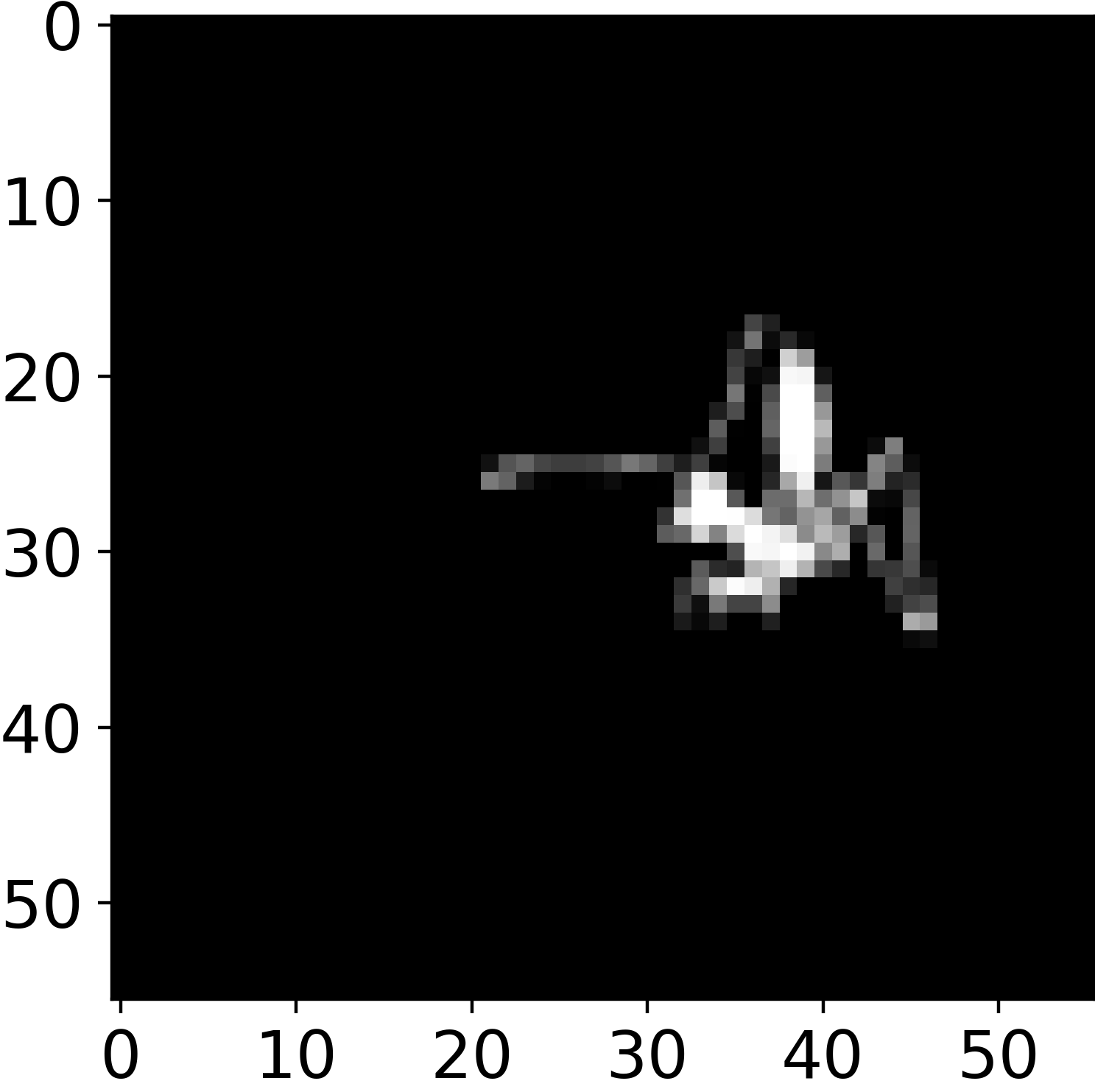}
\put (80,82) {\textcolor{white}{(d)}}
\end{overpic}
\begin{overpic}[width=0.24\linewidth]{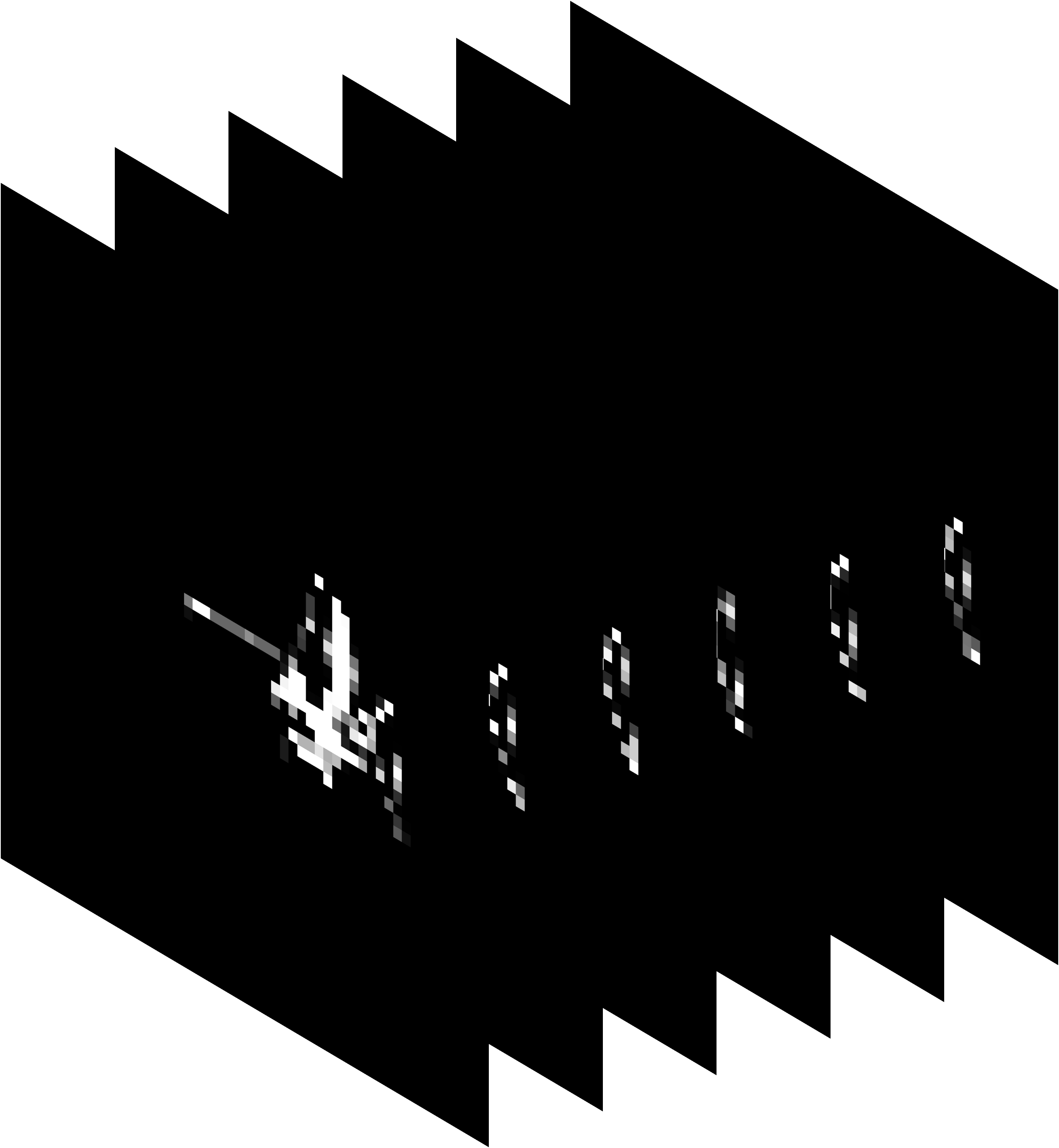}
\put (75,80) {\textcolor{black}{(e)}}
\end{overpic}
\begin{overpic}[width=0.24\linewidth]{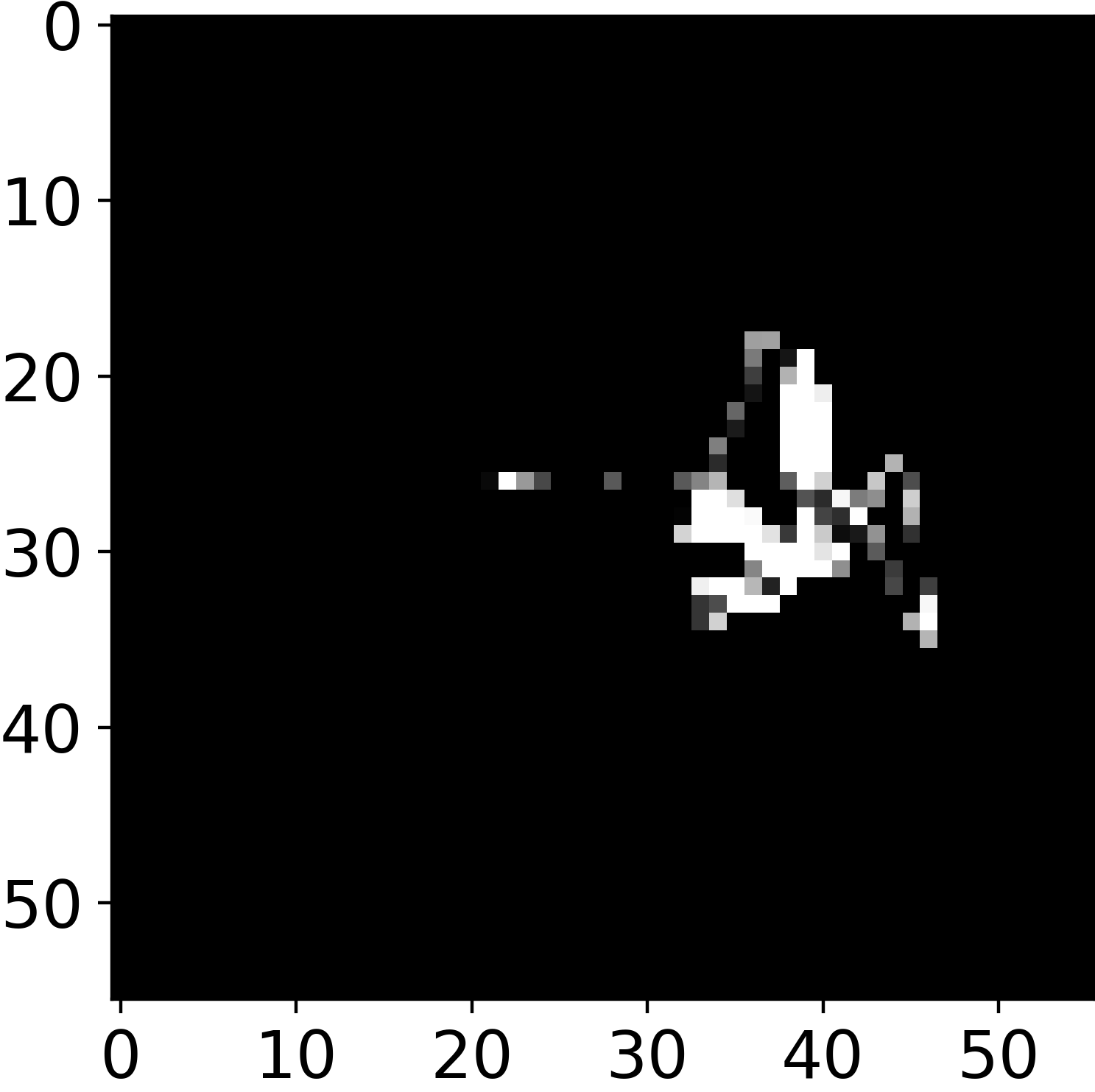}
\put (80,82) {\textcolor{white}{(f)}}
\end{overpic}
\begin{overpic}[width=0.24\linewidth]{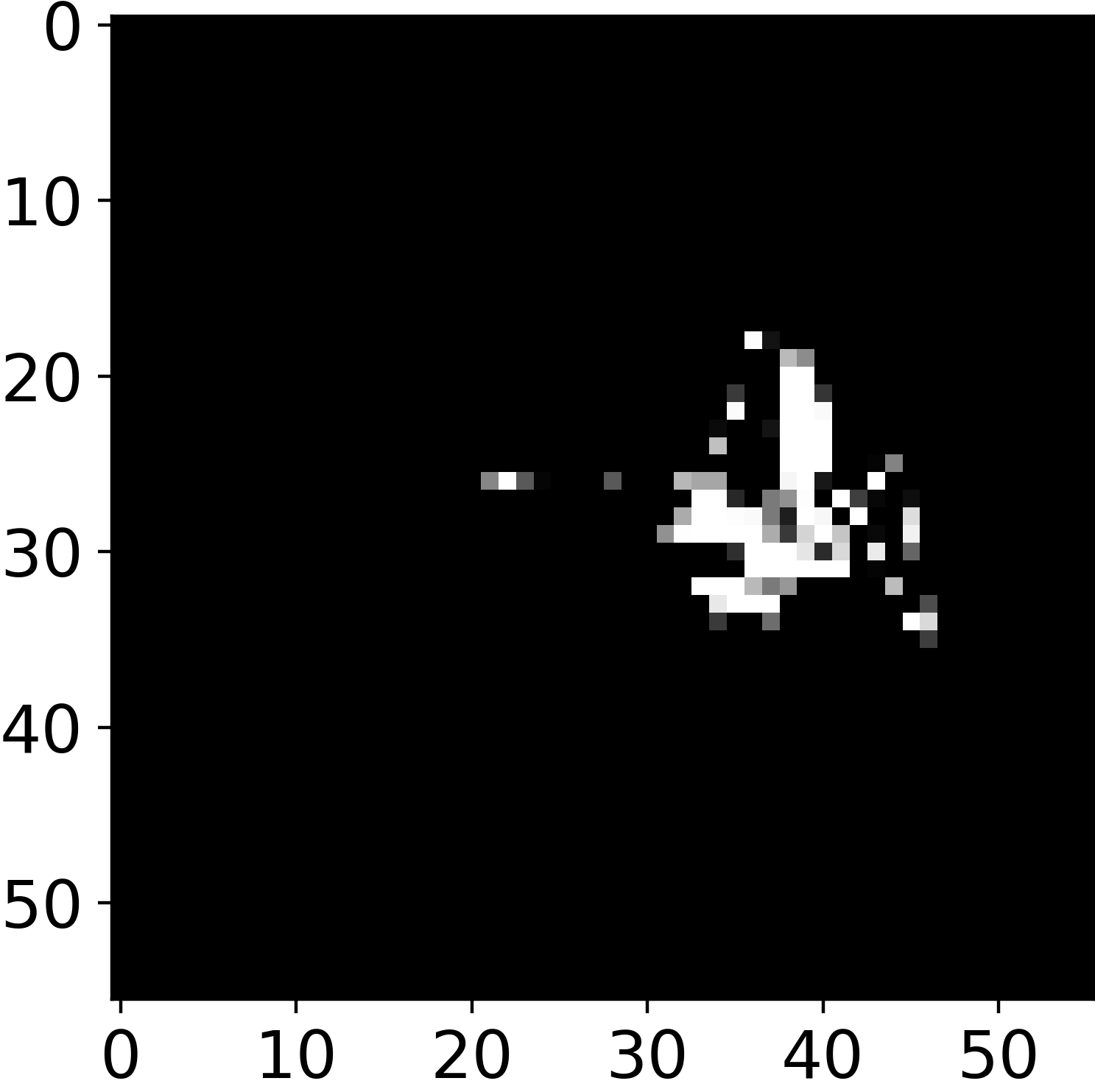}
\put (80,82) {\textcolor{white}{(g)}}
\end{overpic}
\begin{overpic}[width=0.24\linewidth]{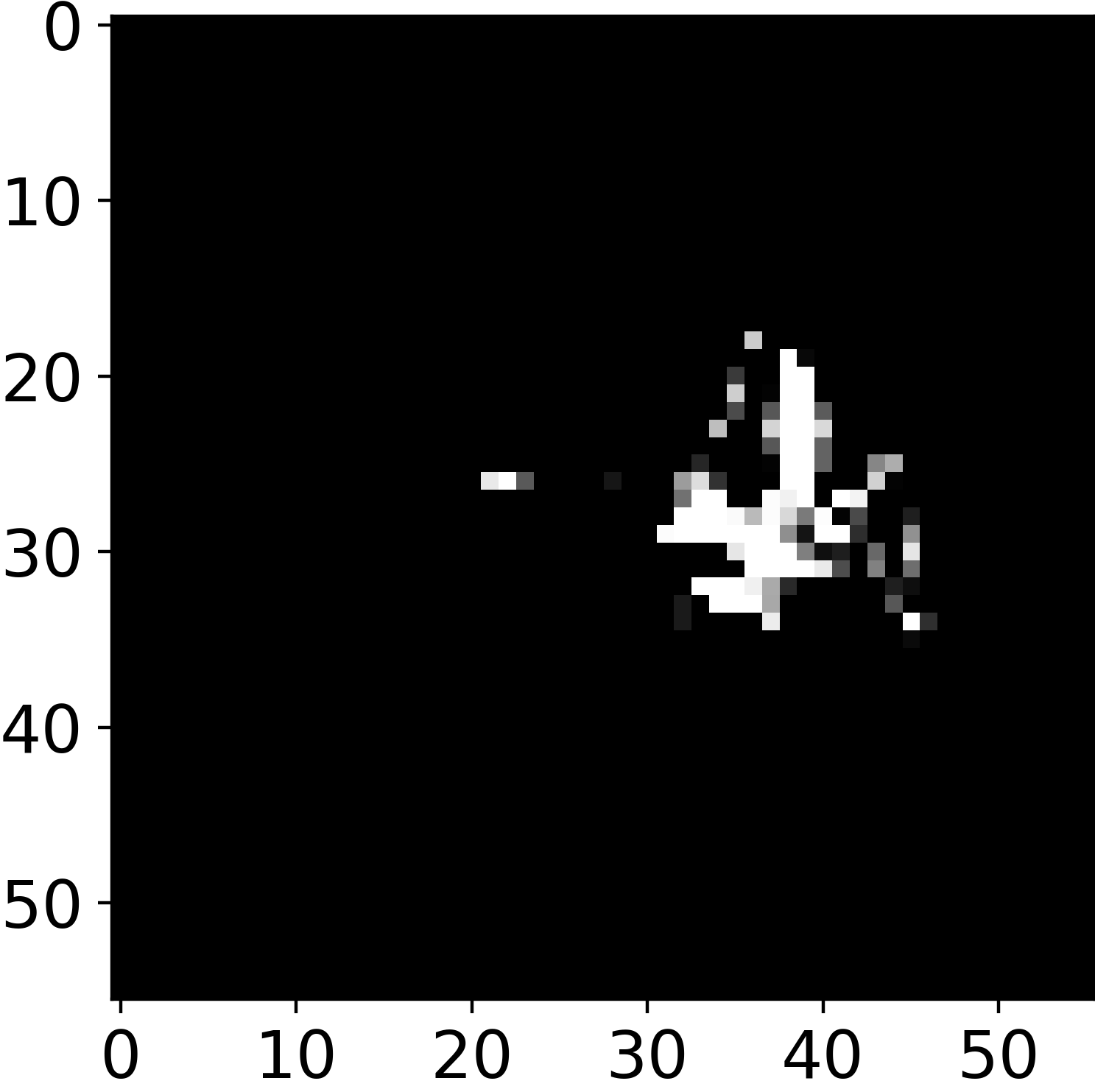}
\put (80,82) {\textcolor{white}{(h)}}
\end{overpic}
\caption{Visual comparisons of vairous downsampling methods from $224\times224$ to $56\times56$. (a) GT, (b) bilinear interpolation, (c) max pooling($kernel=4, stride=4$), (d) average pooling($kernel=4, stride=4$), (e) stacks of pixel shuffled images, (f)-(h) examples of different channels of pixel shuffled image. Best view in zoon-in.}
\label{PS_COMPARE}
\vspace{-0.02\linewidth}
\end{figure}

\begin{figure*}[!t]
\begin{center}
\includegraphics[width=.99\linewidth]{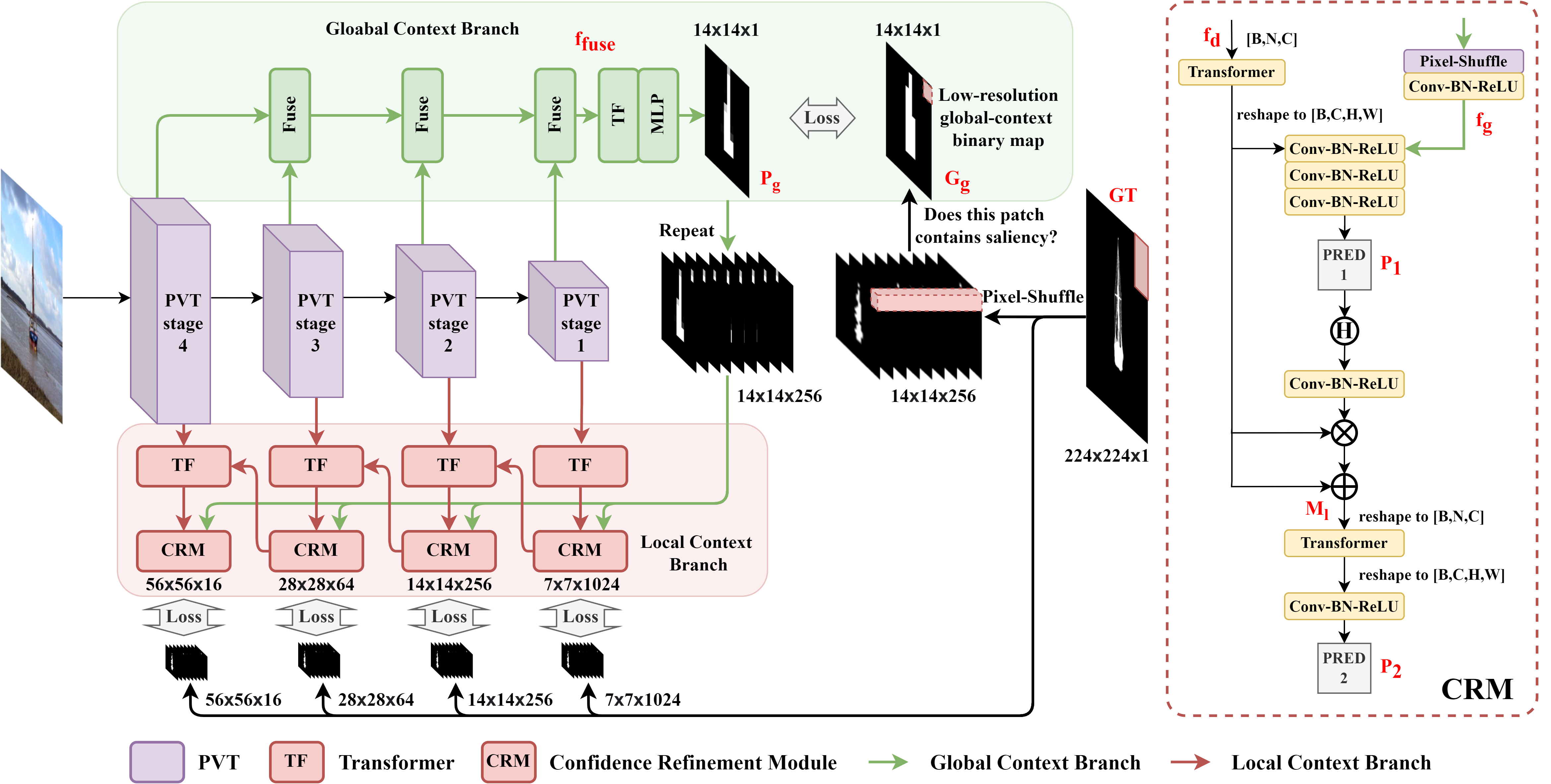}
\end{center}
\caption{Network architecture for SelfReformer. Pre-trained PVT-v2 is employed as the encoder backbone. Encoder features are fed into the global context branch for patch-wise classification to obtain a low-resolution global context map. The map is further fused in CRM to locate the salient object. A CRM is developed for detail refinement where it utilizes its first prediction as a clue and generates features to improve its second prediction. In between each stage, Pixel Shuffle is applied to avoid the loss of details caused by interpolation methods.}
\label{fig:network}
\vspace{-0.02\linewidth}
\end{figure*}

As shown in Fig.\ref{fig:network}, input image in the shape of $224\times 224$ will be cut into patches for self-attention, and PVT will output four groups of features in the shape of $56\times 56$, $28\times 28$, $14\times 14$, and $7\times 7$. For more efficient computation of the multi-head self-attention, PVT introduced a sequence reduction method to reduce the scale of $K$ and $V$ by firstly reshaping the input sequence $X_{i}\in\mathbb{R}^{(HW\times C)}$ into $\hat{X}_{i}\in\mathbb{R}^{(\frac{HW}{r}\times C\times r)}$ then apply an MLP network to reduce the channel of $C\times r$ back to $C$, the process is formulated as:

\begin{equation}
\hat{X}_{i} = LN(MLP(Reshape(X, r)))
\end{equation}
where $LN(\cdot)$ stands for layer normalization \cite{LN}. The self-attention is performed based on the reduced $\hat{K}$ and $\hat{V}$:
\begin{equation}
Attention(Q, \hat{K}, \hat{V}) = Softmax(\frac{Q\hat{K}^{T}}{\sqrt{d_{head}}})\hat{V}
\end{equation}
 As a result, the total computation is reduced $r$ times and hence more efficient.

\subsection{Pixel Shuffle as the Up/down Sampling Method}
SOD is a densely supervised task where the ground truth image needs to be downsampled into multiple LR images to fit the size of each decoder stage. We noticed that given the input size of $224\times224$, PVT's largest feature map size is $56\times 56$, and conventional downsampling methods like interpolation or pooling are no longer viable to generate accurate ground truth images for the decoder. As shown in Fig.\ref{PS_COMPARE}, in (b), fine structures are damaged and inconsistent in the bilinear sampled ground truth. In (c) and (d), the generated GTs via max and average pooling become inaccurate. As we increase the downsampling factor, the methods mentioned above will discard or change more and more pixel values, resulting in different inconsistent GTs for each decoder stage. In contrast, Pixel Shuffle rearranges the GT from $I^{H\times W\times 1}$ into multi-channel LR images $I^{\frac{H}{r} \times \frac{W}{r}\times r^{2}}$, since no pixel is discarded nor changed thus the structural properties are preserved. Though each pixel shuffled channel contains incomplete GT in (e) - (h) due to the reshaping process, the overall image is still the same once we shuffle them back to a single channel. Thus Pixel Shuffle is a more suitable method for downsampling the ground truth owing to its ability to unshuffle an HR image into LR images without changing the value, as illustrated in Fig.\ref{the_ps} previously. Different from all downsampling methods, by using Pixel Shuffle, we will train each decoder stage against the full-scale ground truth instead of its downgraded LR images. Predictions from each decoder stage will now become $P^{\frac{H}{r} \times \frac{W}{r}\times r^{2}}$ instead of $P^{\frac{H}{r} \times \frac{W}{r}\times 1}$, and this training scheme will enable each decoder stage to capture as much information as possible to restore the fine structures of the salient object. To formulate the unshuffle process, given an HR image or feature map $I\in\mathbb{R}^{H\times W\times C} $ and a scaling factor $r$, it can be described as:
\begin{equation}
\newcommand{\floor}[1]{\lfloor #1 \rfloor}
\mathcal PS(I_{x,y,c}, r) = I_{\floor{x/r}, \floor{y/r}, C\cdot r\cdot mod(y,r)+C\cdot mod(x,r)+c}
\end{equation}
where $x, y$ and $c$ represent pixel coordinates and channel index in high-resolution (HR) space.

\subsection{Global-Context Branch}
The global context is the clue indicating where are the salient objects. Though evidence indicates high-level encoder features contain global context and contribute to the completeness of predictions, we still lack a method to evaluate \textit{how much} and \textit{how good} are the global context we obtained from the encoder. Hence, we aim to design a supervised task to explicitly learn the information from the input image and the ground truth pair. Since in Transformer, input images are split into patches, therefore, we frame this supervised task as \textit{which patch contains salient object}. The ground truth of this task can thus be easily obtained from the original ground truth images. For each patch, the branch is only required to predict a single value indicating the likelihood of the presence of the salient object, and the obtained global-context map will be passed to the decoder as guidance to locate the salient object. Since it is a patch-wise prediction instead of full-scale pixel-wise, the designed task is easier than the salient object detection scoped for the decoder. The developed branch will learn a representation of the global context in a controllable manner, and its features will be used as a map to guide the decoder network.

\begin{figure}[!t]
\centering
\includegraphics[width=\linewidth]{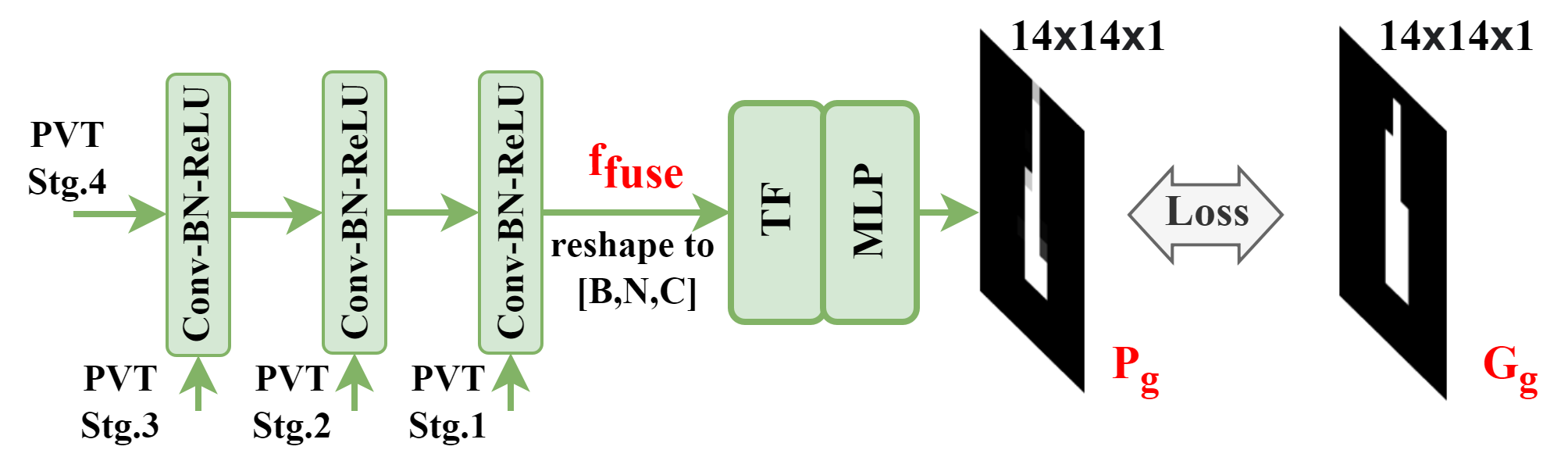} 
\caption{Global Context Branch. A patch-wise SOD task is framed to catch the global context explicitly.}
\label{glc}
\vspace{-0.02\linewidth}
\end{figure}

To build the branch, as shown in Fig.\ref{glc}, we firstly apply Pixel Shuffle to reshape all encoder features to $14\times14$ then concatenate and use a few Conv-BN-ReLU layers for feature fusion. Then a transformer block $\mathcal{TF}$ from the original Transformer and an MLP layer are employed for patch-wise saliency prediction $P_{g}$. For simplicity, let $f_{fuse}$ represent the fused features being passed to the Transformer, and the global context branch can be described as:
\begin{equation}
P_{g} = Sigmoid(MLP(\mathcal{TF}(Reshape(f_{fuse}))))
\end{equation}
where $Reshape$ is the tensor operation from $[B,C,H,W]$ to $[B,H\times W, C]$. The self-attention in $\mathcal{TF}$ is the same as the original transformer:
\begin{equation}
Attention(Q, K, V) = Softmax(\frac{QK^T}{\sqrt{d_{head}}})V
\end{equation}
To obtain the ground truth $G_{g}$ for this branch, we firstly apply Pixel Shuffle to reshape the original ground truth $G$ from $224\times 224\times 1$ to $14\times 14\times 256$, then apply $max(\cdot)$ function along the channel dimension $c$:
\begin{equation}
G_{g} = \max\limits_{c\in C}((\mathcal{PS}(G))_{i,j,c})
\end{equation}
where $i, j$ and $c$ represent pixel coordinates and channel indices. The branch is supervised using Binary Cross-Entropy(BCE) Loss:
\begin{equation}
\mathcal{L}_{g} = BCE(P_{g}, G_{g})
\end{equation}
The obtained map is then passed to Context Refinement Module (CRM) for fusion, and Pixel Shuffle is applied accordingly to match different scales in each decoder stage.

\subsection{Context Refinement Module (CRM)}

\begin{figure}[!t]
\centering
\begin{minipage}{.18\linewidth}
\centering
\includegraphics[width=\linewidth]{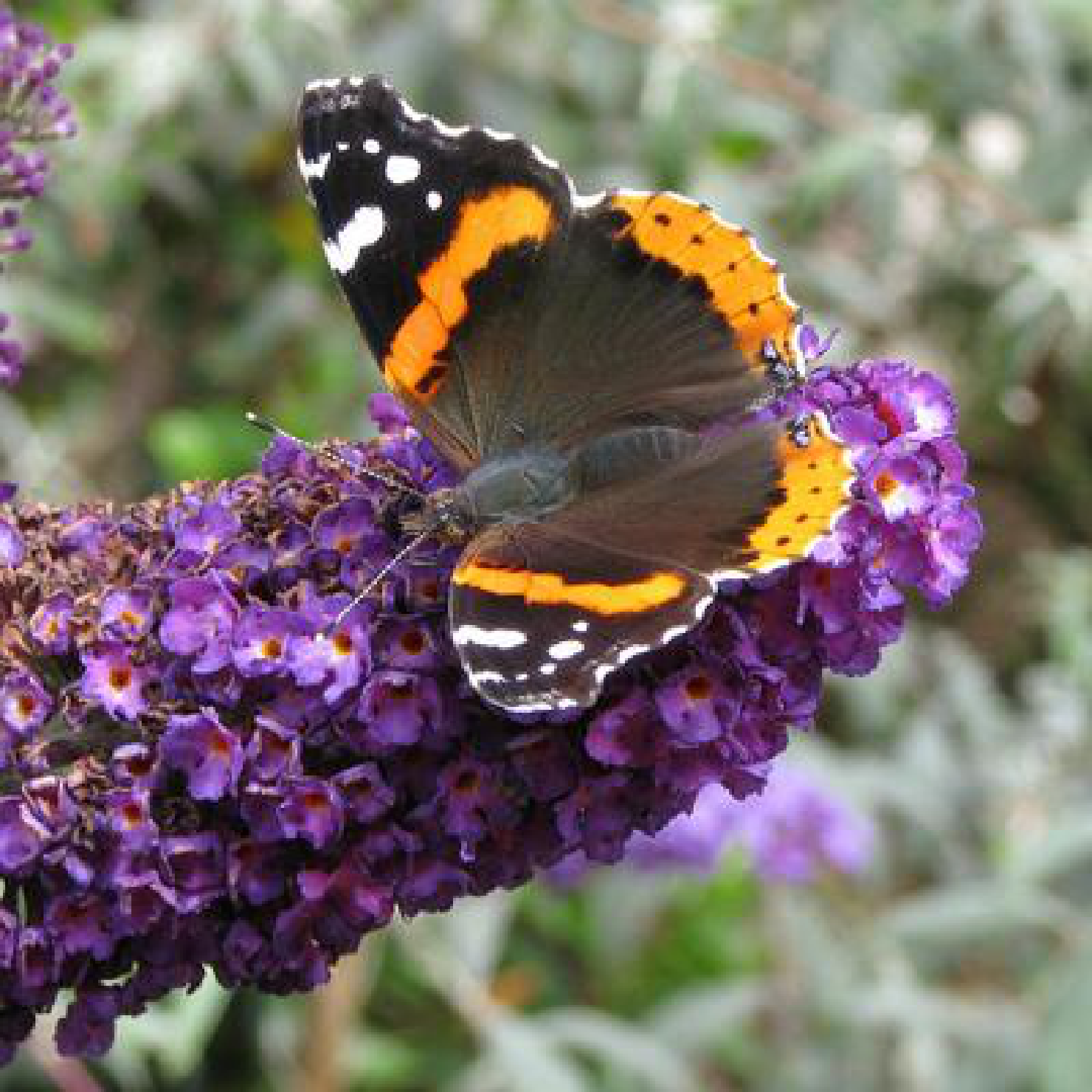} 
\includegraphics[width=\linewidth]{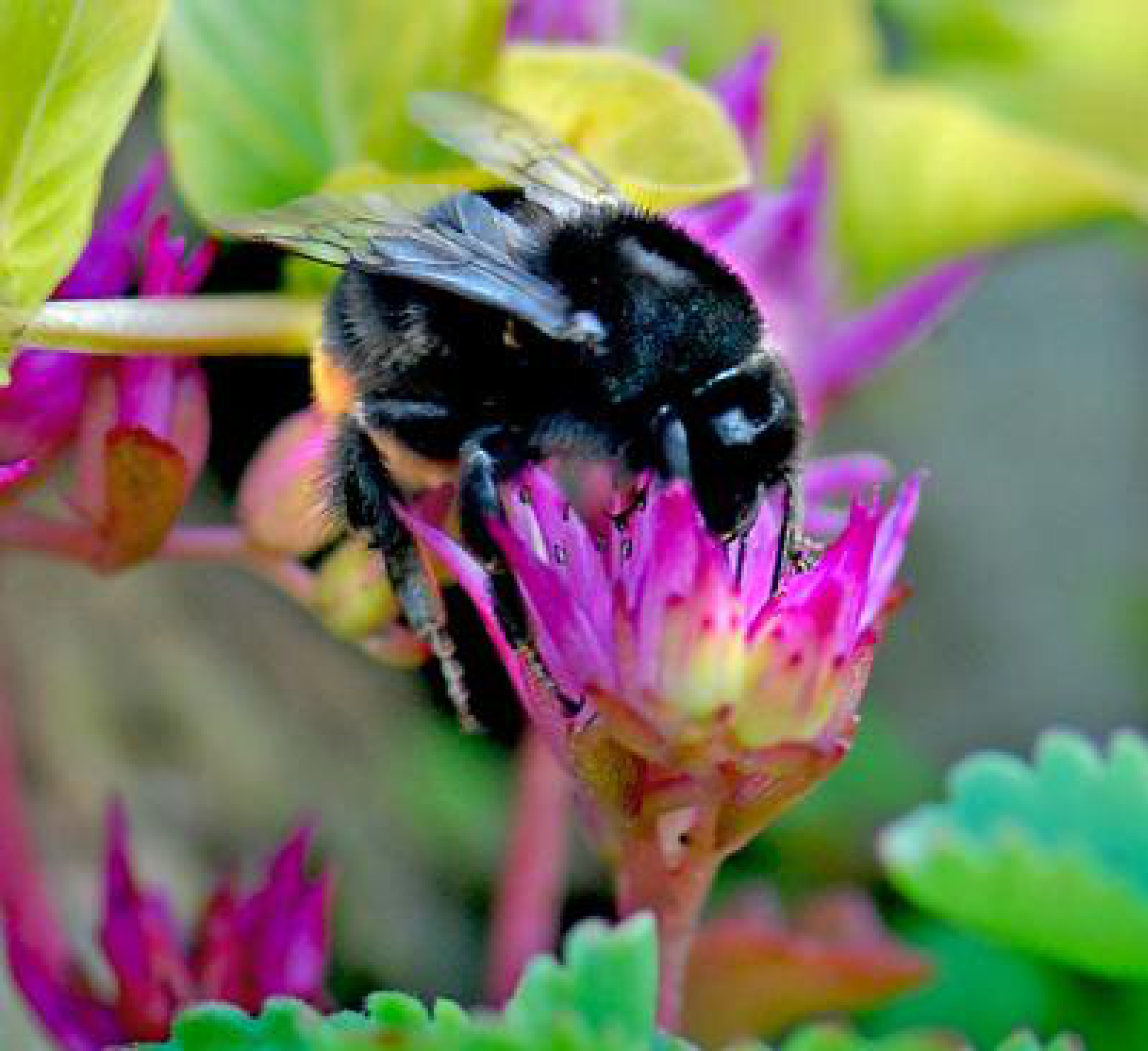} 
\includegraphics[width=\linewidth]{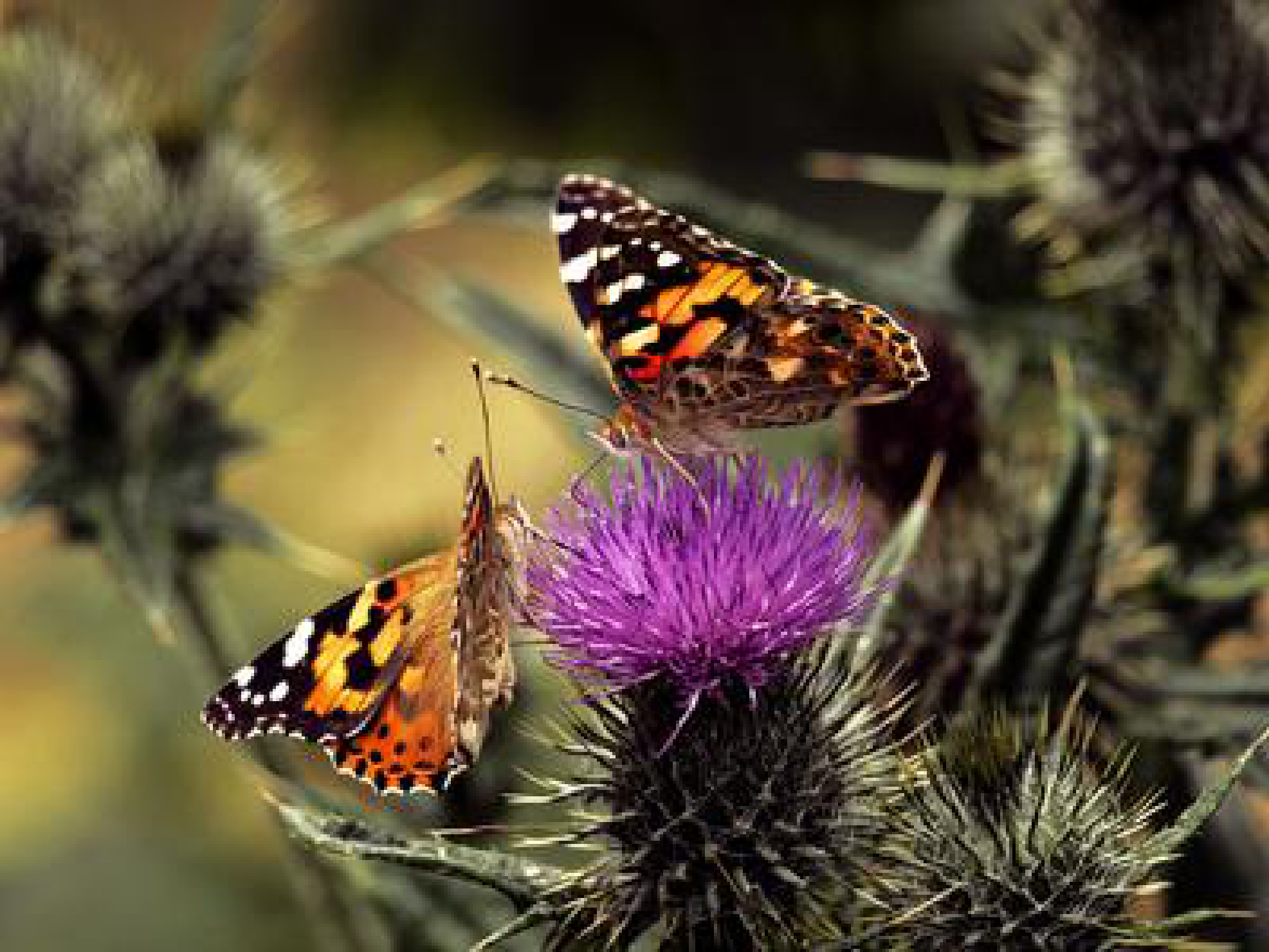}
\end{minipage}
\begin{minipage}{.18\linewidth}
\centering
\includegraphics[width=\linewidth]{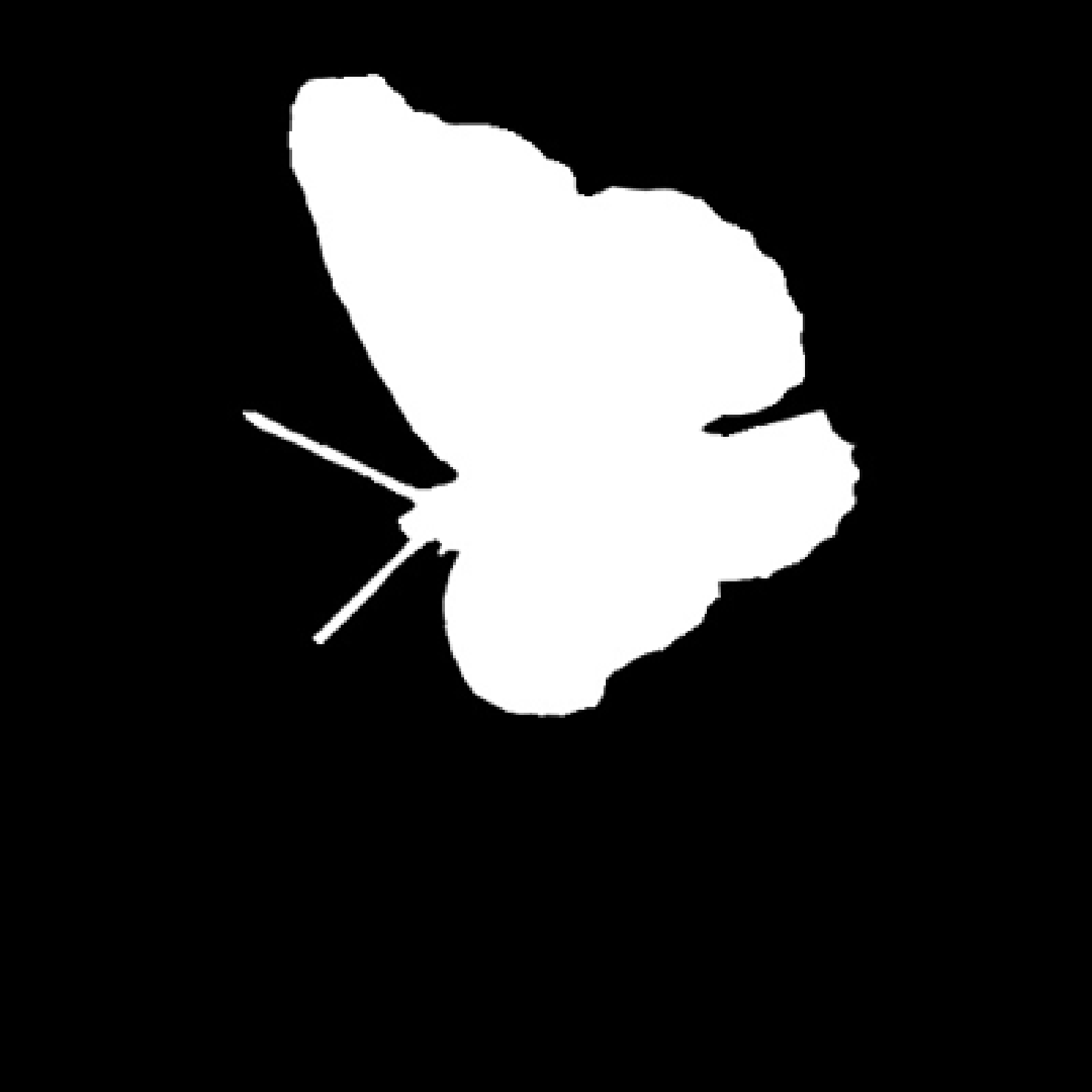} 
\includegraphics[width=\linewidth]{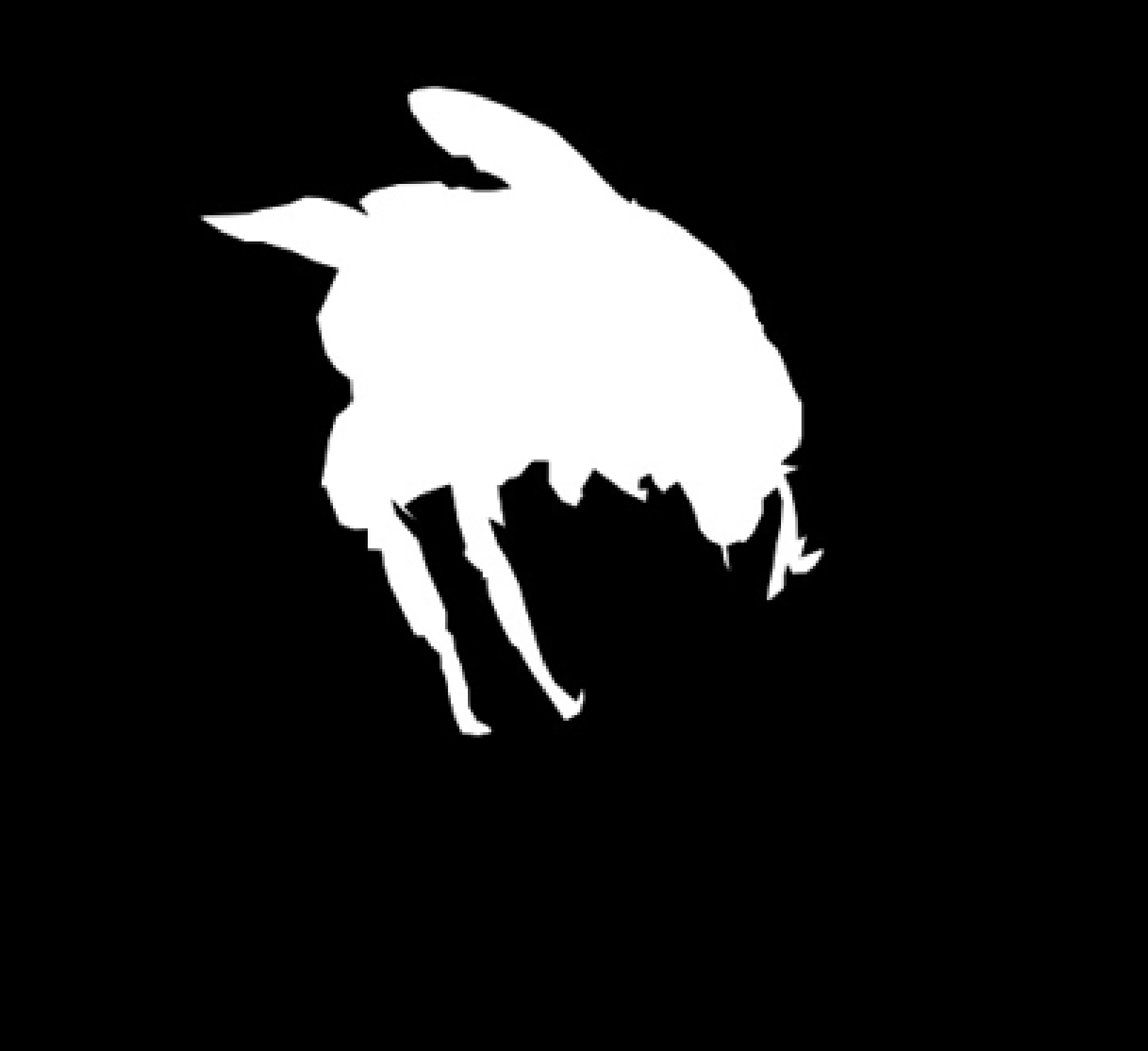} 
\includegraphics[width=\linewidth]{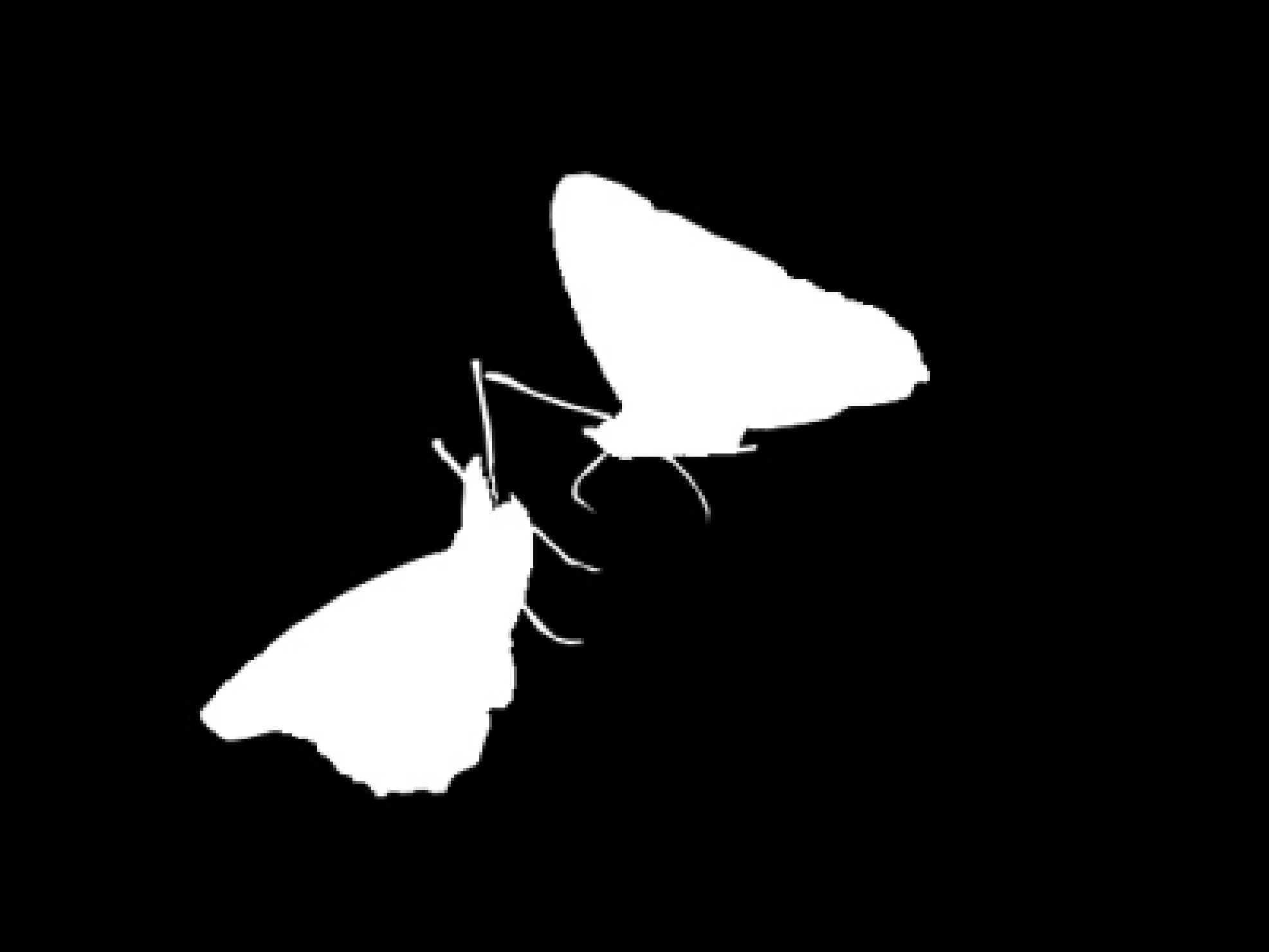}
\end{minipage}
\begin{minipage}{.18\linewidth}
\centering
\includegraphics[width=\linewidth]{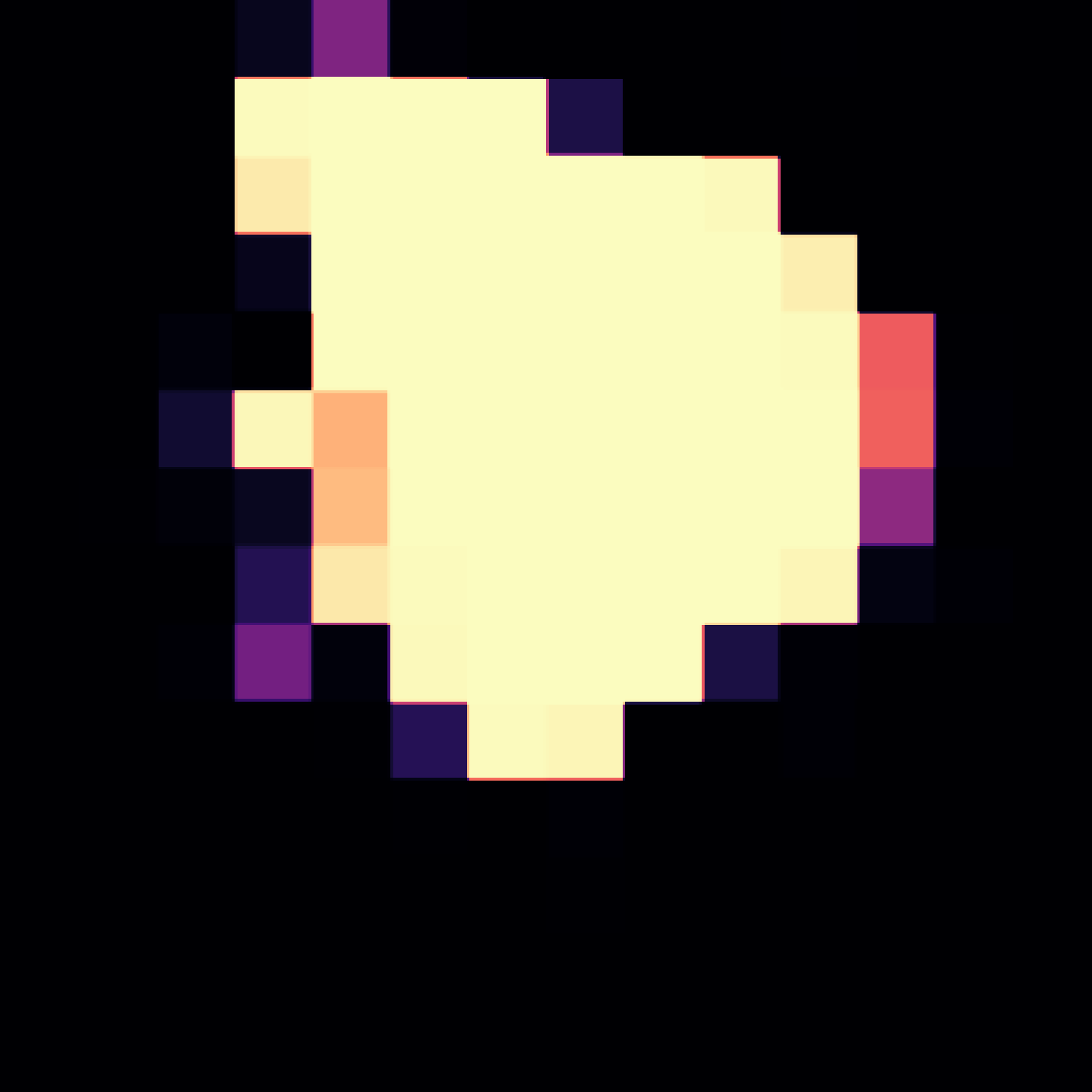} 
\includegraphics[width=\linewidth]{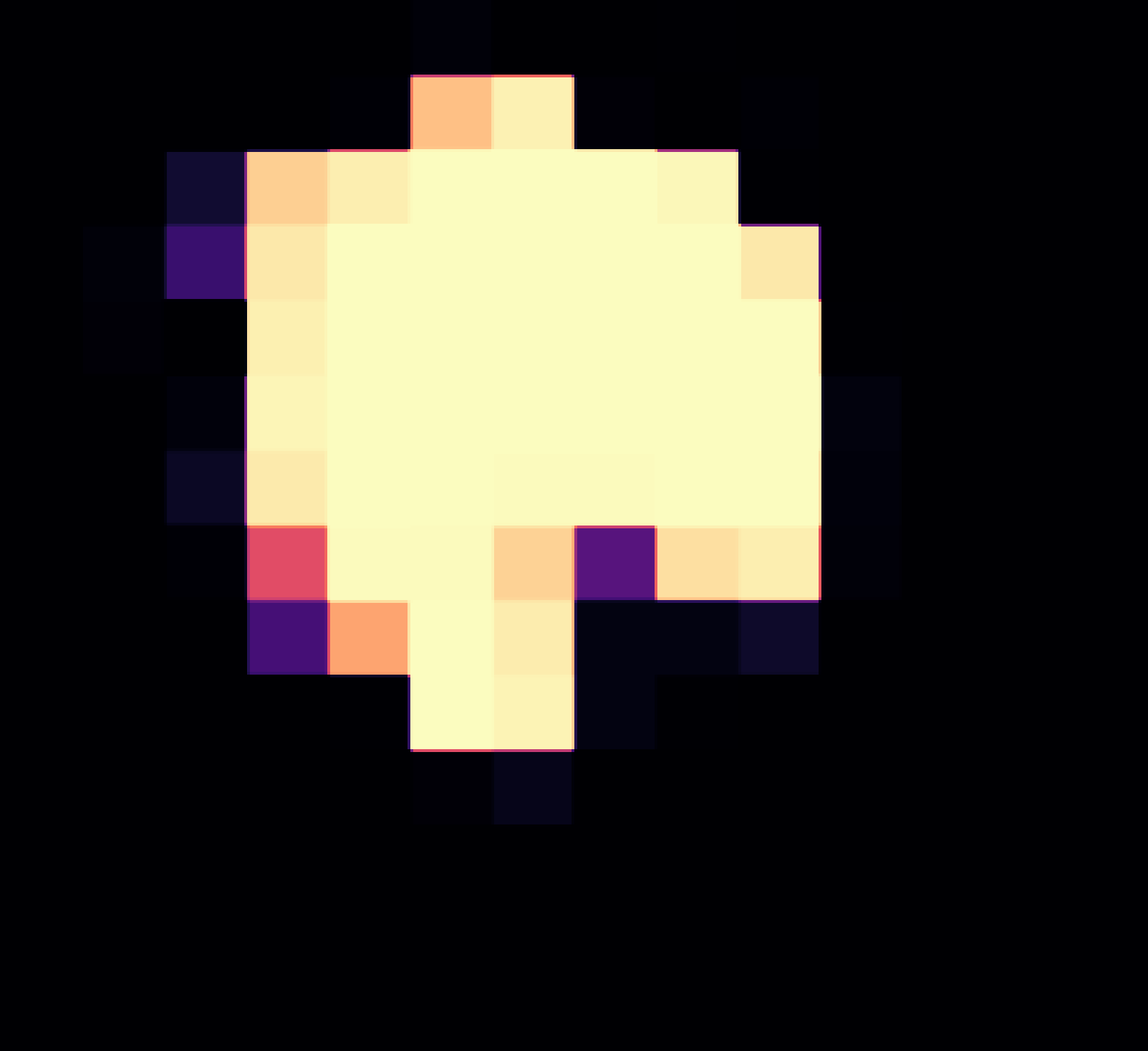} 
\includegraphics[width=\linewidth]{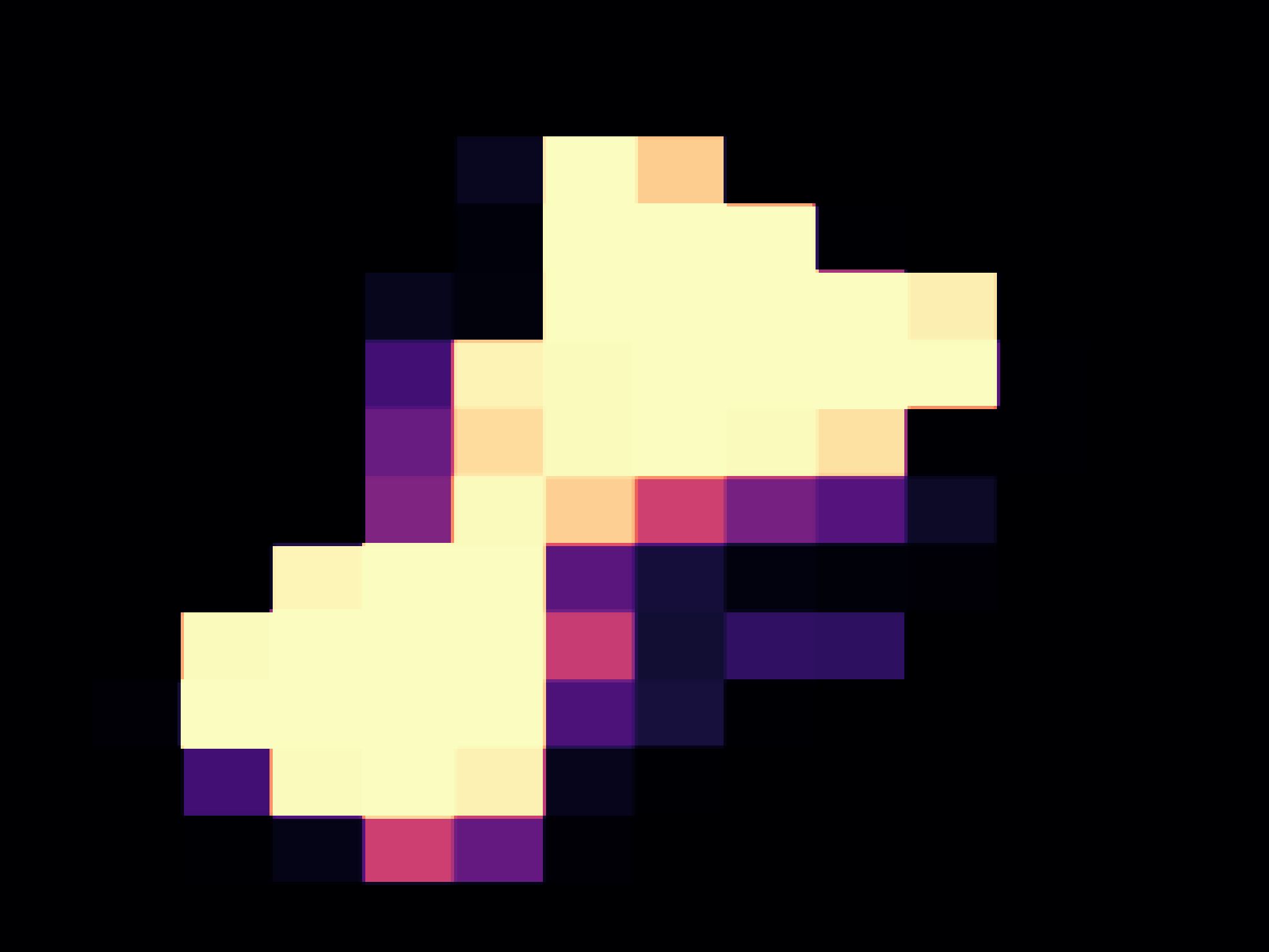}
\end{minipage}
\begin{minipage}{.18\linewidth}
\centering
\includegraphics[width=\linewidth]{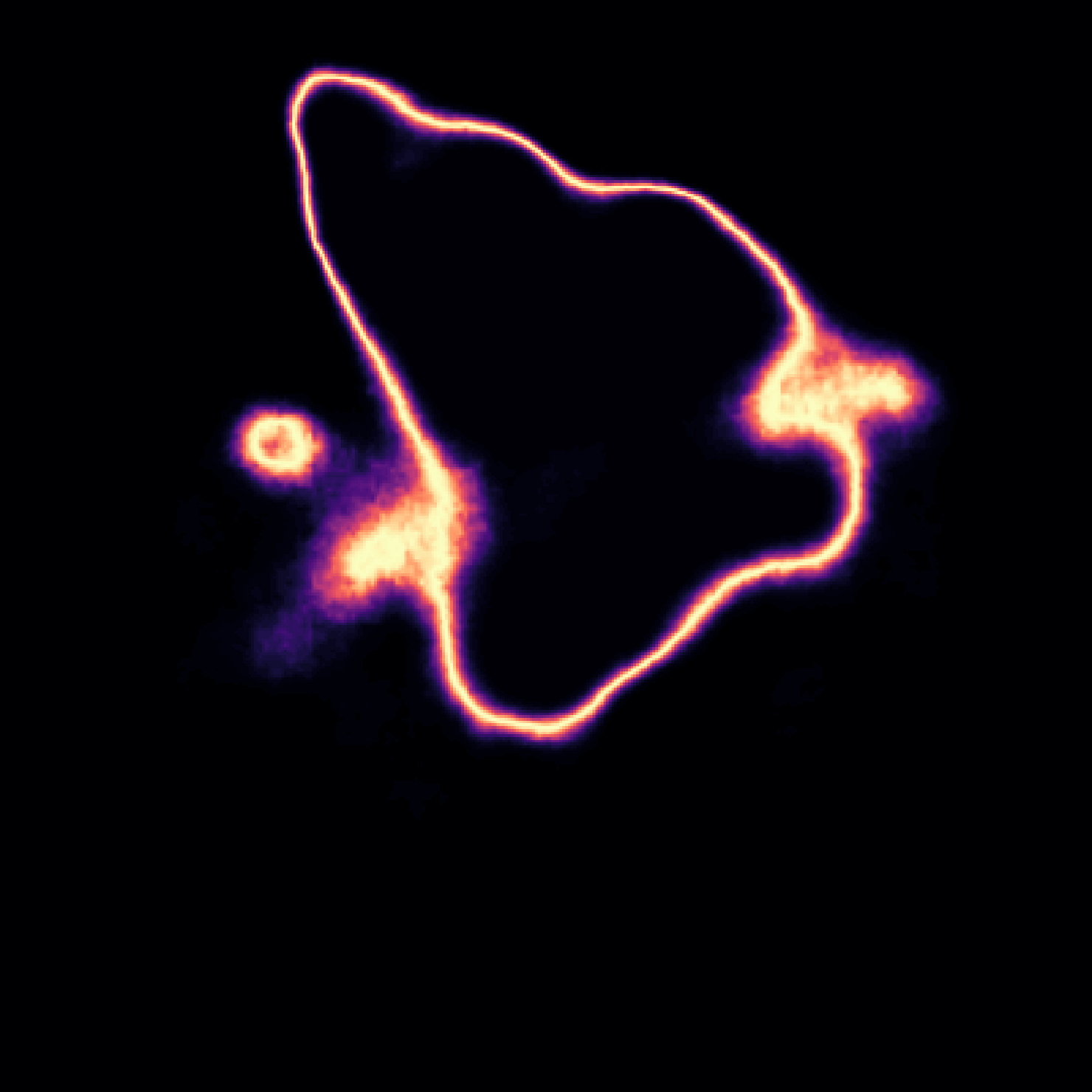} 
\includegraphics[width=\linewidth]{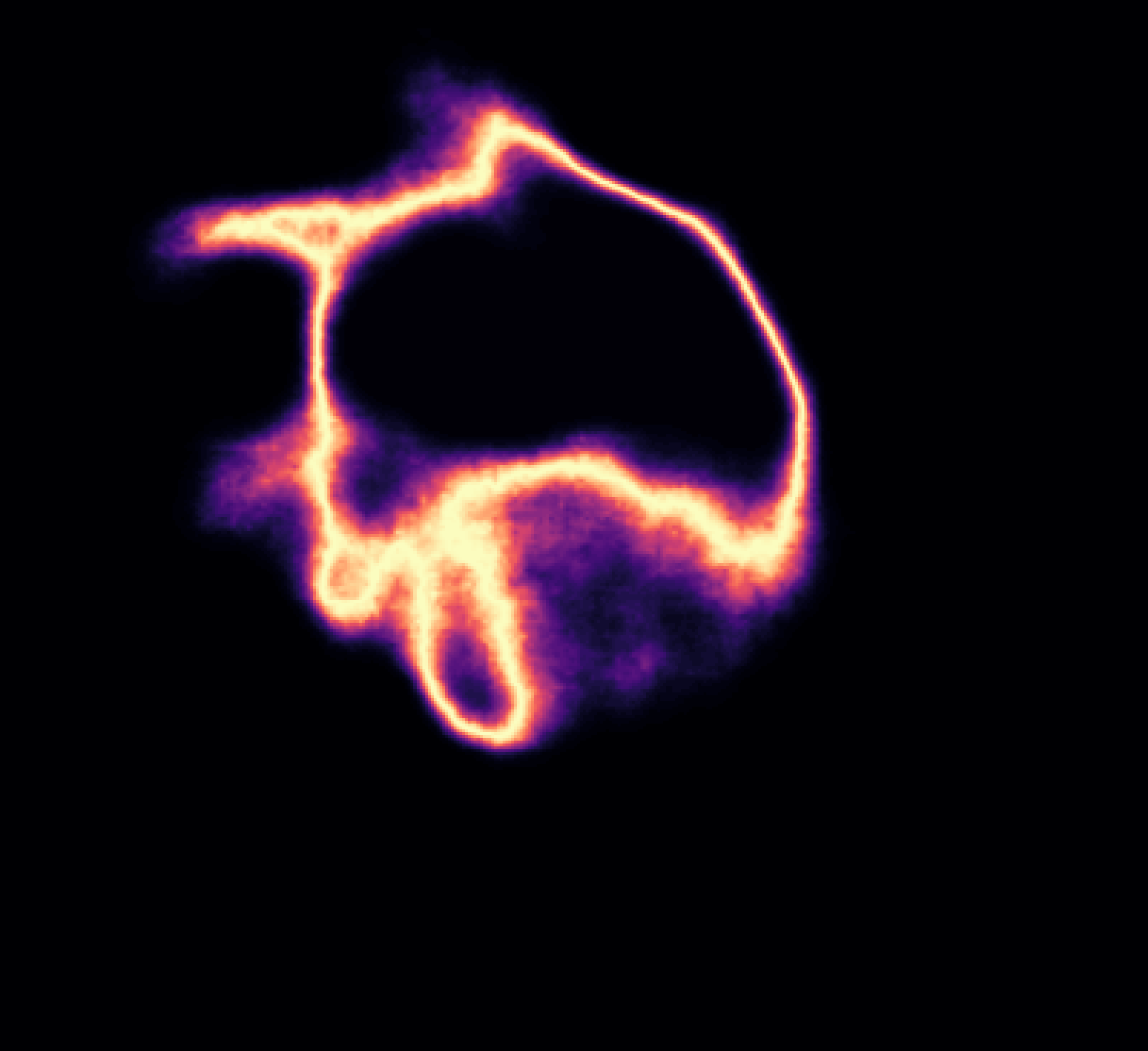} 
\includegraphics[width=\linewidth]{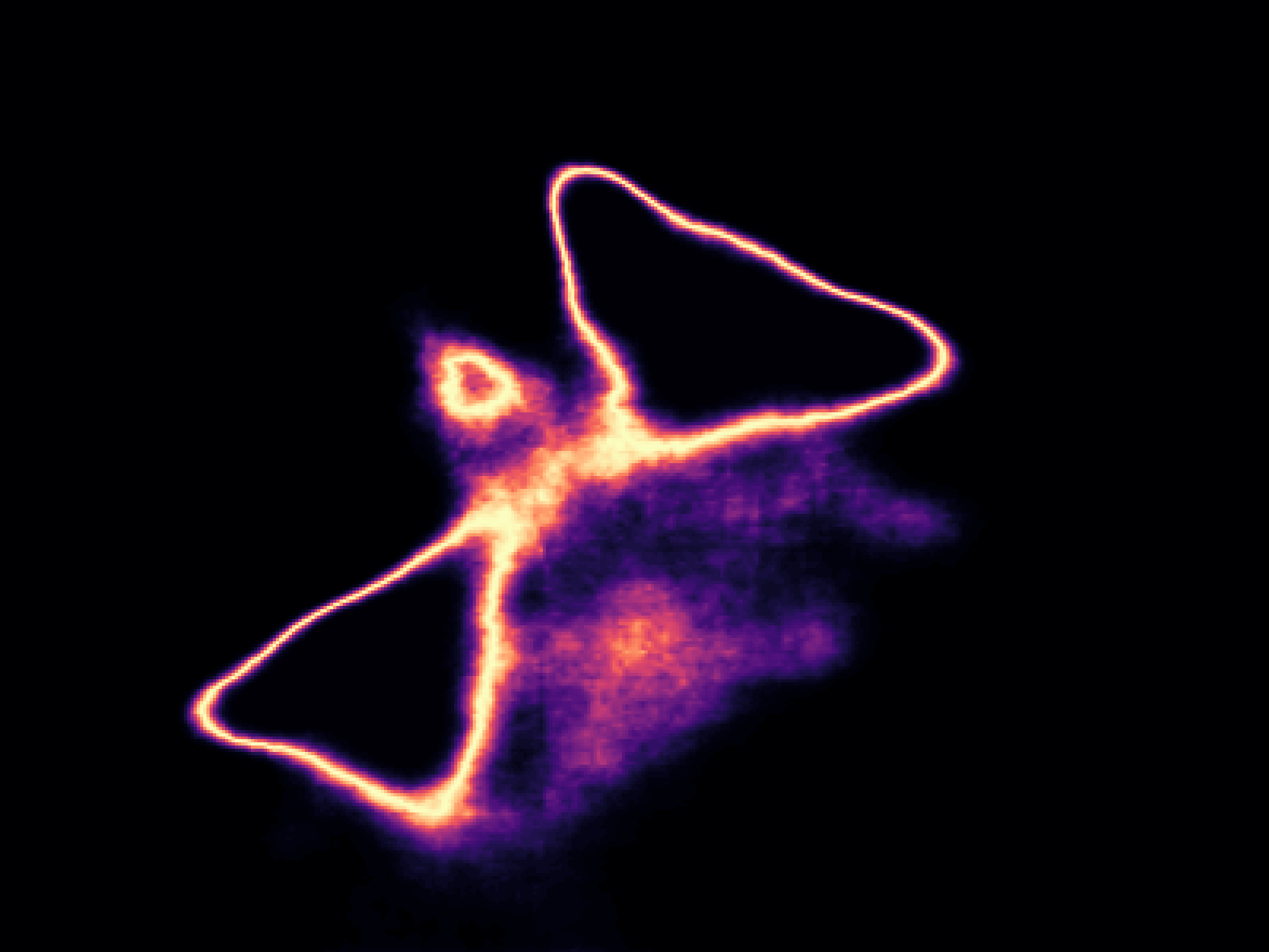}
\end{minipage}
\begin{minipage}{.18\linewidth}
\centering
\includegraphics[width=\linewidth]{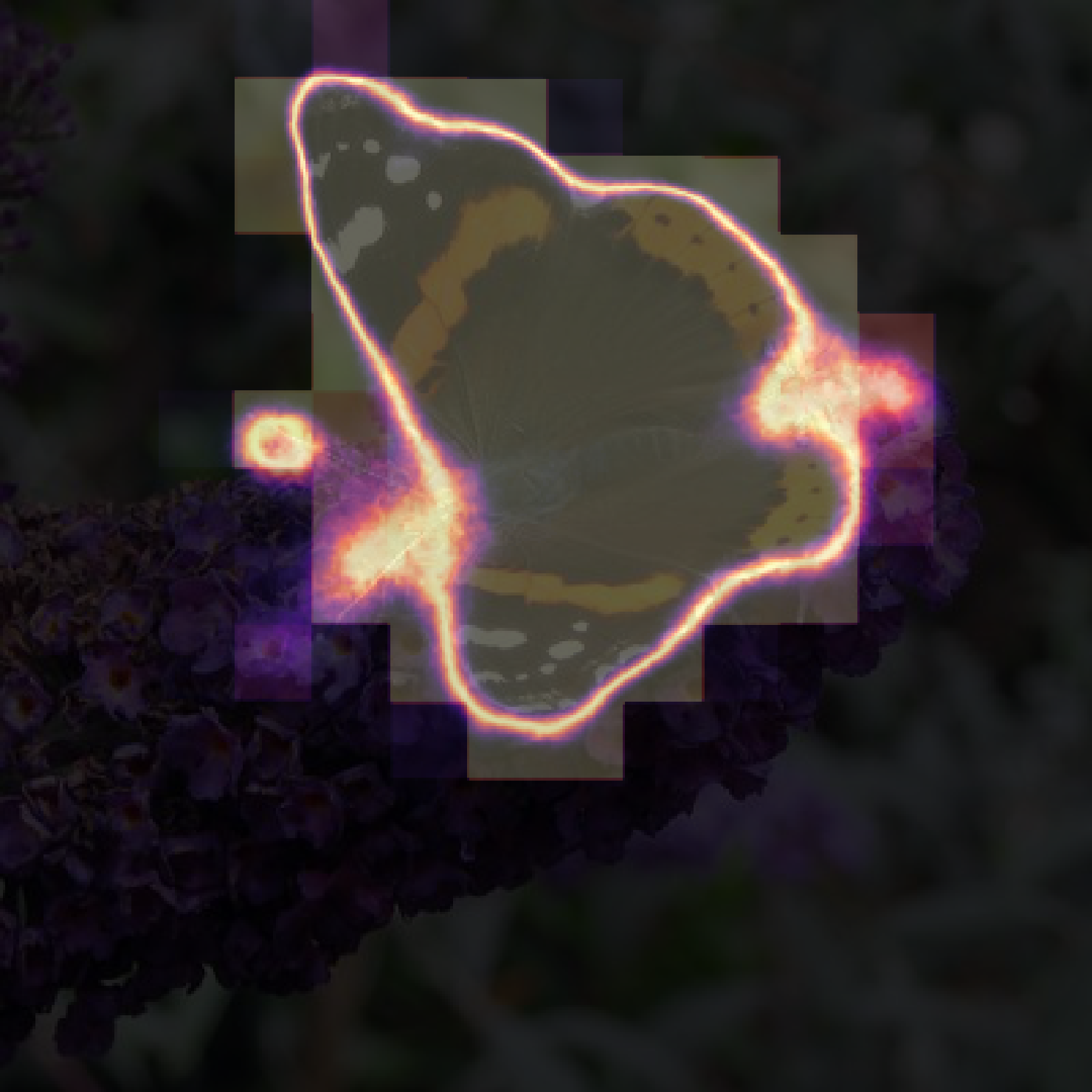} 
\includegraphics[width=\linewidth]{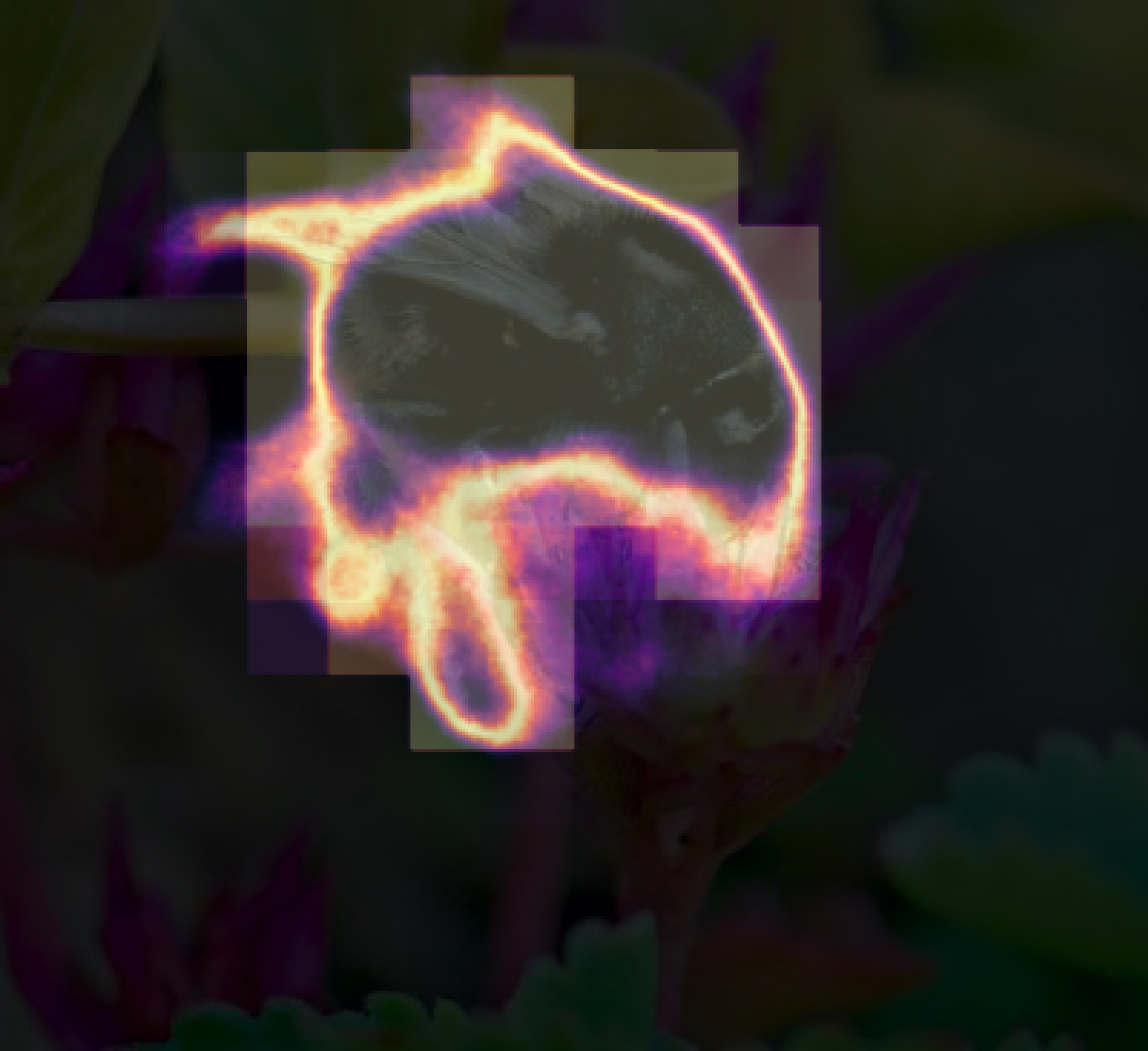} 
\includegraphics[width=\linewidth]{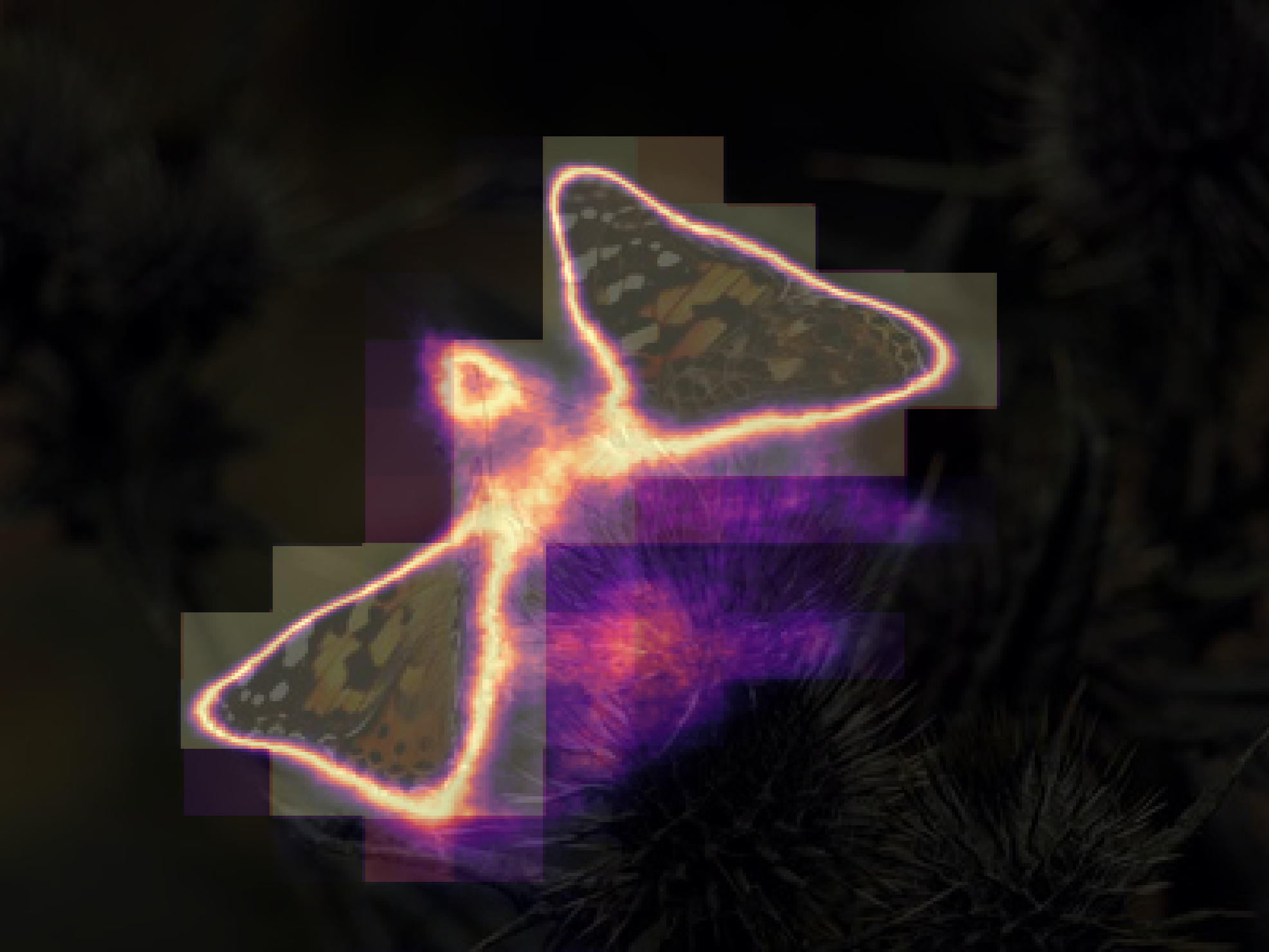}
\end{minipage}
\caption{Illustration of fusion global and local context in CRM. From left to right: Input, Ground Truth, Global-context map, Local-context map, Overlay of global and local context. The global-context map can locate the salient object while the local-context map is to refine details.}
\label{CRM_list}
\vspace{-0.05\linewidth}
\end{figure}

We propose CRM to guide the network for better semantic integrity and refine its predictions for richer details. Key steps and results are shown in Fig.\ref{CRM_list} where the global context map is fused with decoder features; then a local refinement map is generated for fine structure segmentation. Thus CRM is a two-stage module where we handle global and local information separately, as shown in Fig.\ref{fig:network}.

For global information, to match the feature map dimension, Pixel Shuffle $\mathcal PS$ is applied on the global context features $f_{g}$ with different scaling factors $r$ depending on the decoder stage. Decoder features $f_{d}$ are fused with $f_{g}$ via a few Conv-BN-ReLU layers denoted as $\mathcal{F}_{1}$, and the first stage prediction $\mathcal P_{1}$ is obtained and supervised against the ground truth. Mathematically, this process can be described as:
\begin{equation}
P_{1} = Sigmoid(\mathcal{F}_{1}(f_{d}, \mathcal PS( f_{g}, r)))
\end{equation}

Above obtained $P_{1}$ contains unconfident regions in the presence of grey areas in the image. These areas are considered as hard pixels to the network. Noteworthily, due to the property of the Sigmoid function, values of hard pixels are close to $0.5$ while values are close to $0$ or $1$ for confident predictions. By multiplying $P_{1}$ with $1-P_{1}$, the unconfident area are highlighted, and features can be extracted as the local-context map $\mathcal{M}_{l}$ to guide the second stage to focus and refine the unsure regions in $P_{1}$. Denote the designed multiplication as $\mathcal H$, we adopt a single Conv-BN-ReLU layer $\mathcal{F}_{2}$ to obtain the map:
\begin{equation}
\mathcal{H}(P_{1}) = P_{1}*(1-P_{1})
\end{equation}
\begin{equation}
\mathcal{M}_{l} = f_{d}*\mathcal{F}_{2}(\mathcal{H}(P_{1}))+f_{d}
\end{equation}
where * represents element-wise multiplication. We further adopt a transformer block $\mathcal{TF}$ and another Conv-BN-ReLU layer $\mathcal{F}_{3}$ to generate the final prediction $P_{2}$:
\begin{equation}
P_{2} = Sigmoid(\mathcal{F}_{3}(\mathcal{TF}(\mathcal{M}_{l})))
\end{equation}
The obtained $P_{2}$ has better quality in fine structures than $P_{1}$, which will be discussed in ablation studies. The proposed CRM achieves \textit{self-refine} as it adopts global context to guide the decoder for better completeness, and automatically refines the details in the prediction.

We apply the weighted BCE loss ($\mathcal{L}_{w}$) for each decoder stage as used in F3Net \cite{F3NET}:
\begin{equation}
\mathcal{L}_{l} = \sum_{i=1}^{4}\lambda_{i}(\mathcal{L}_{w}^{i}(P_{1}, G)+\mathcal{L}_{w}^{i}(P_{2}, G))
\end{equation}
where subscript $i$ represents each decoder stage as listed in Fig.\ref{fig:network}, and the values of $\lambda_{1-4}$ are $[0.5, 0.7, 0.9, 1.1]$ respectively. The total loss of the network $\mathcal{L}$ is simply the sum of $\mathcal{L}_{g}$ and $\mathcal{L}_{l}$ as described above.

\section{Experiments}
\subsection{Implementation Details}
\textbf{DUTS-TR} \cite{DUTSTE}(10553 images) was used as the training dataset and we resize images resized to 224×224. Random $90^{\circ}$ rotation and horizontal flipping were applied as the data augmentation. Besides, we apply Leaky ReLU \cite{LEAKY} for convolution layers, GeLU \cite{GELU} and Layer Norm \cite{LN} for Transformers. Adam optimizer \cite{ADAM} with default hyperparameters was adopted to train the network. We trained the network for 200 epochs with batch size of 16. Learning rates for the encoder, global context branch, and the decoder were set to $10^{-5}$, $10^{-5}$, and $10^{-4}$, respectively, and were halved every 40 epochs. During testing, images are resized to 224×224, and the predictions from last decoder stage (56×56×16) were firstly upscaled to 224×224×1 using Pixel Shuffle, and bilinear interpolation was applied to further resize it back to its original size.
\begin{table*}[!t]
\centering
\setlength\tabcolsep{2pt}
\caption{Quantitative comparisons between our SelfReformer and other 11 methods on five benchmark datasets. Text in \textbf{bold} indicates the best performance, and superscript * stands for Transformer based network. Postfix \textit{\textbf{BI}} of our work stands for network using bilinear interpolation instead of Pixel Shuffle, and \textit{\textbf{full}} represents our proposed network.}
\begin{tabular}{c|cccc|cccc|cccc|cccc|cccc} 
\hline
\hline
\multirow{2}{*}{Methods} & \multicolumn{4}{c|}{\textbf{DUTS-TE}} & \multicolumn{4}{c|}{\textbf{HKU-IS}} & \multicolumn{4}{c|}{\textbf{PASCAL-S}} & \multicolumn{4}{c|}{\textbf{ECSSD}} & \multicolumn{4}{c}{\textbf{DUT-OMRON}}  \\
                  & \textbf{$F_{\beta}\hspace{-1.5mm}\uparrow$} & \textbf{$M\hspace{-1.5mm}\downarrow$} & \textbf{$E_{\xi}\hspace{-1.5mm}\uparrow$} & \textbf{$S_{\alpha}\hspace{-1.5mm}\uparrow$}
                  & \textbf{$F_{\beta}\hspace{-1.5mm}\uparrow$} & \textbf{$M\hspace{-1.5mm}\downarrow$} & \textbf{$E_{\xi}\hspace{-1.5mm}\uparrow$} & \textbf{$S_{\alpha}\hspace{-1.5mm}\uparrow$}
                  & \textbf{$F_{\beta}\hspace{-1.5mm}\uparrow$} & \textbf{$M\hspace{-1.5mm}\downarrow$} & \textbf{$E_{\xi}\hspace{-1.5mm}\uparrow$} & \textbf{$S_{\alpha}\hspace{-1.5mm}\uparrow$}
                  & \textbf{$F_{\beta}\hspace{-1.5mm}\uparrow$} & \textbf{$M\hspace{-1.5mm}\downarrow$} & \textbf{$E_{\xi}\hspace{-1.5mm}\uparrow$} & \textbf{$S_{\alpha}\hspace{-1.5mm}\uparrow$}
                  & \textbf{$F_{\beta}\hspace{-1.5mm}\uparrow$} & \textbf{$M\hspace{-1.5mm}\downarrow$} & \textbf{$E_{\xi}\hspace{-1.5mm}\uparrow$} & \textbf{$S_{\alpha}\hspace{-1.5mm}\uparrow$}        \\ 
\hline
\hline
F\textsuperscript{3}Net\textsubscript{20}     &.891 &.035 &.901 &.888 &.936 &.028 &.952 &.917 &.871 &.061 &.858 &.854 &.945 &.033 &.927 &.924 &.813 &.052 &.869 &.838    \\
\hline
GateNet\textsubscript{20}  &.887 &.040 &.889 &.885 &.933 &.033 &.949 &.915 &.869 &.067 &.851 &.851 &.945 &.040 &.924 &.919 &.818 &.054 &.862 &.838    \\
\hline
GCPA\textsubscript{20}      &.888 &.038 &.890 &.890 &.938 &.030 &.949 &.920 &.869 &.061 &.846 &.858 &.948 &.034 &.920 &.926 &.811 &.056 &.860 &.838    \\
\hline
MINet\textsubscript{20}     &.883 &.037 &.897 &.884 &.934 &.028 &.953 &.918 &.866 &.063 &.850 &.849 &.947 &.033 &.926 &.924 &.809 &.055 &.864 &.832    \\
\hline
U2Net\textsubscript{20}     &.872 &.044 &.886 &.873 &.935 &.031 &.948 &.915 &.859 &.073 &.842 &.838 &.951 &.033 &.924 &.927 &.822 &.054 &.870 &.846    \\
\hline
LDF\textsubscript{20}        &.897 &.033 &.909 &.892 &.939 &.027 &.953 &.919 &.874 &.059 &.865 &.856 &.950 &.033 &.924 &.924 &.819 &.051 &.873 &.838    \\
\hline
MSFNet\textsubscript{21}  &.877 &.034 &.911 &.875 &.927 &.026 &.953 &.907 &.862 &.060 &.858 &.843 &.941 &.033 &.926 &.914 &.798 &.045 &.862 &.819    \\
\hline
PFSNet \textsubscript{21}  &.896 &.036 &.902 &.892 &.943 &.026 &.956 &.924 &.875 &.063 &.856 &.854 &.952 &.031 &.928 &.930 &.823 &.055 &.875 &.842    \\
\hline
DCN \textsubscript{21}      &.894 &.035 &.903 &.892 &.939 &.027 &.957 &.922 &.872 &.061 &.858 &.855 &.952 &.031 &.929 &.928 &.823 &.051 &.878 &.845    \\
\hline
PAKRN \textsubscript{21}  &.906 &.032 &.916 &.900 &.942 &.027 &.954 &.923 &.873 &.065 &.857 &.851 &.952 &.032 &.923 &.927 &.834 &.049 &.885 &.853    \\
\hline
VST*\textsubscript{21}      &.890 &.037 &.891 &.896 &.942 &.029 &.952 &.928 &.875 &.060 &.837 &.865 &.950 &.032 &.917 &.932 &.824 &.057 &.861 &.850    \\
\hline
\hline
Ours-BI*                          &.905 &.029 &.919 &.904 &.943 &.025 &.956 &.928 &.890 &.052 &.870 &.873 &.953 &.029 &.926 &.933 &.829 &.043 &.877 &.848    \\
\hline
Ours-full*                       &\textbf{.916} &\textbf{.026} &\textbf{.920} &\textbf{.911} &\textbf{.947} &\textbf{.024} &\textbf{.959} &\textbf{.930} &\textbf{.894} &\textbf{.050} &\textbf{.872} &\textbf{.874} &\textbf{.957} &\textbf{.027} &\textbf{.928} &\textbf{.935} &\textbf{.836} &\textbf{.041} &\textbf{.886} &\textbf{.856}    \\
\hline
\hline
\end{tabular}
\label{metric_result}
\end{table*}

\begin{figure*}[!h]
\centering
\begin{minipage}{.19\linewidth}
\centering
\includegraphics[width=\linewidth]{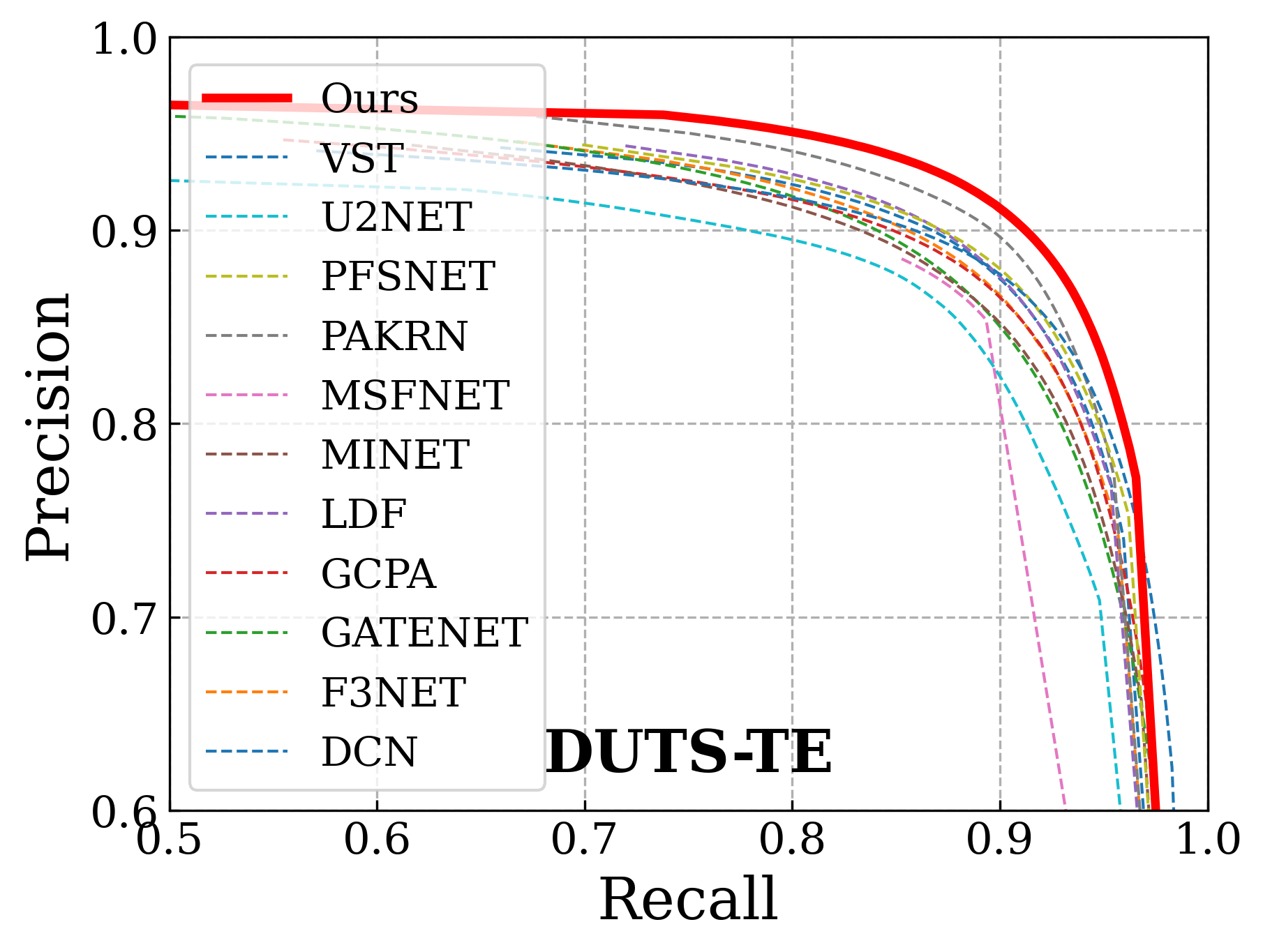} 
\includegraphics[width=\linewidth]{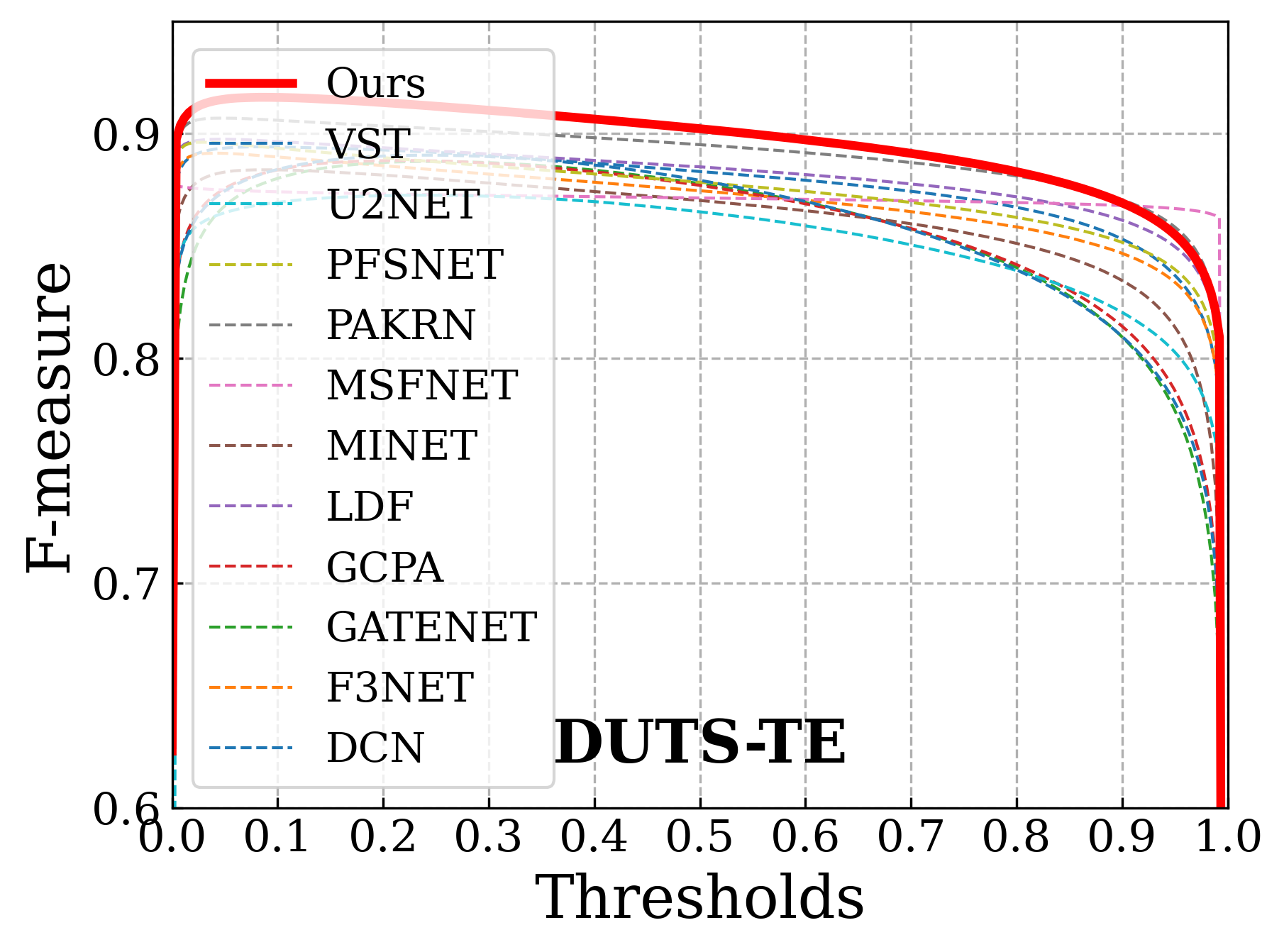}
\end{minipage}
\begin{minipage}{.19\linewidth}
\centering
\includegraphics[width=\linewidth]{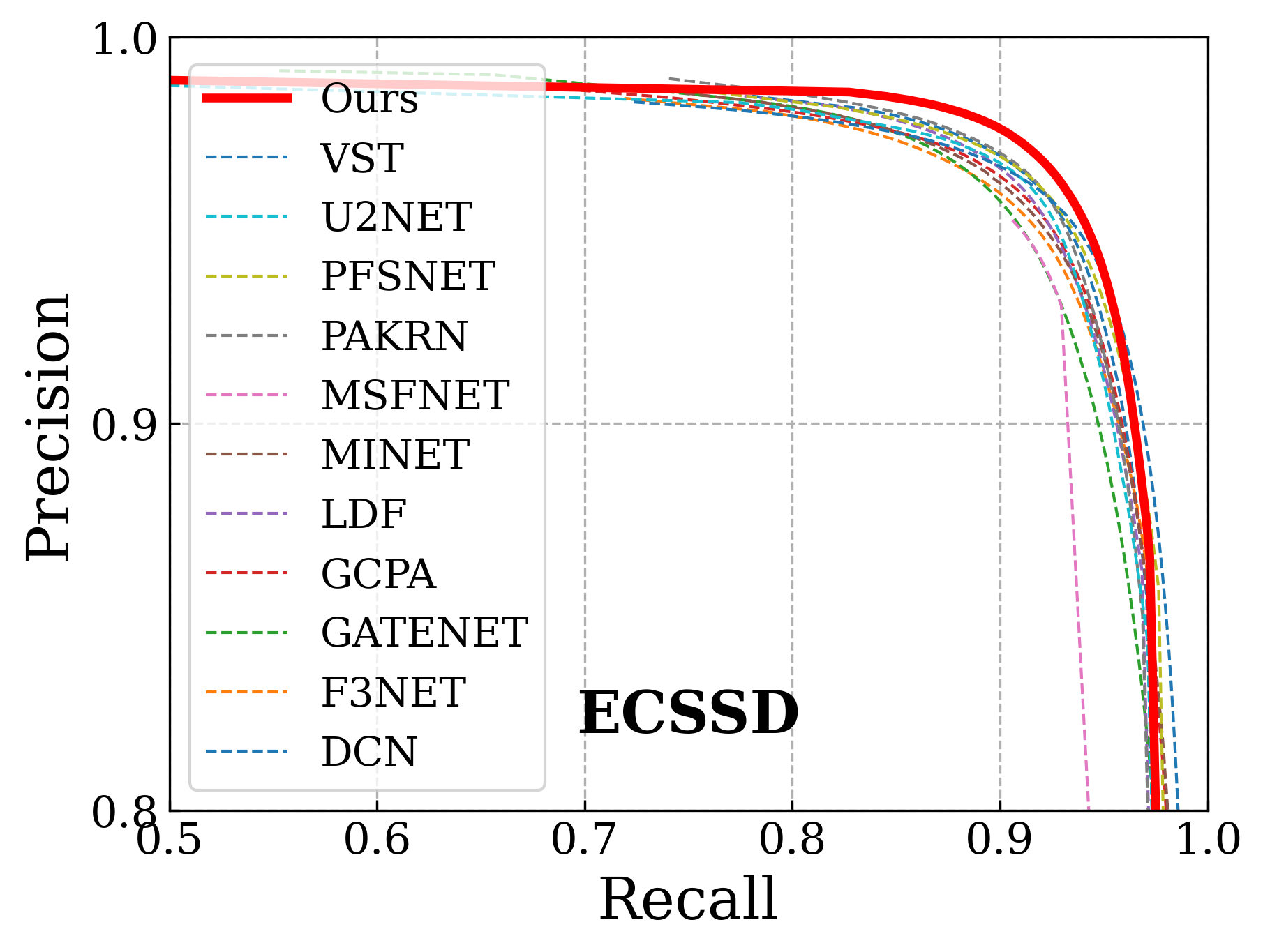} 
\includegraphics[width=\linewidth]{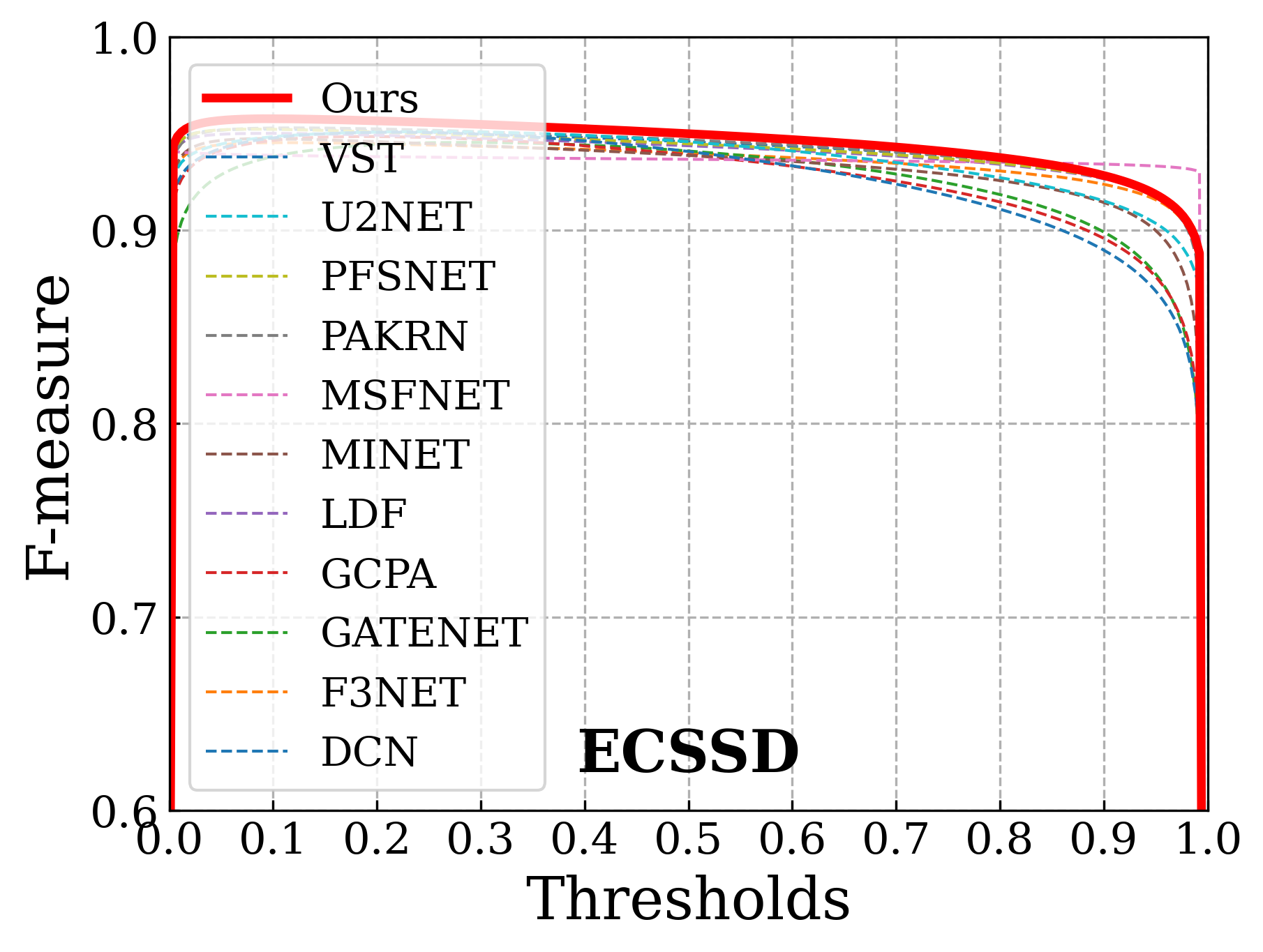}
\end{minipage}
\begin{minipage}{.19\linewidth}
\centering
\includegraphics[width=\linewidth]{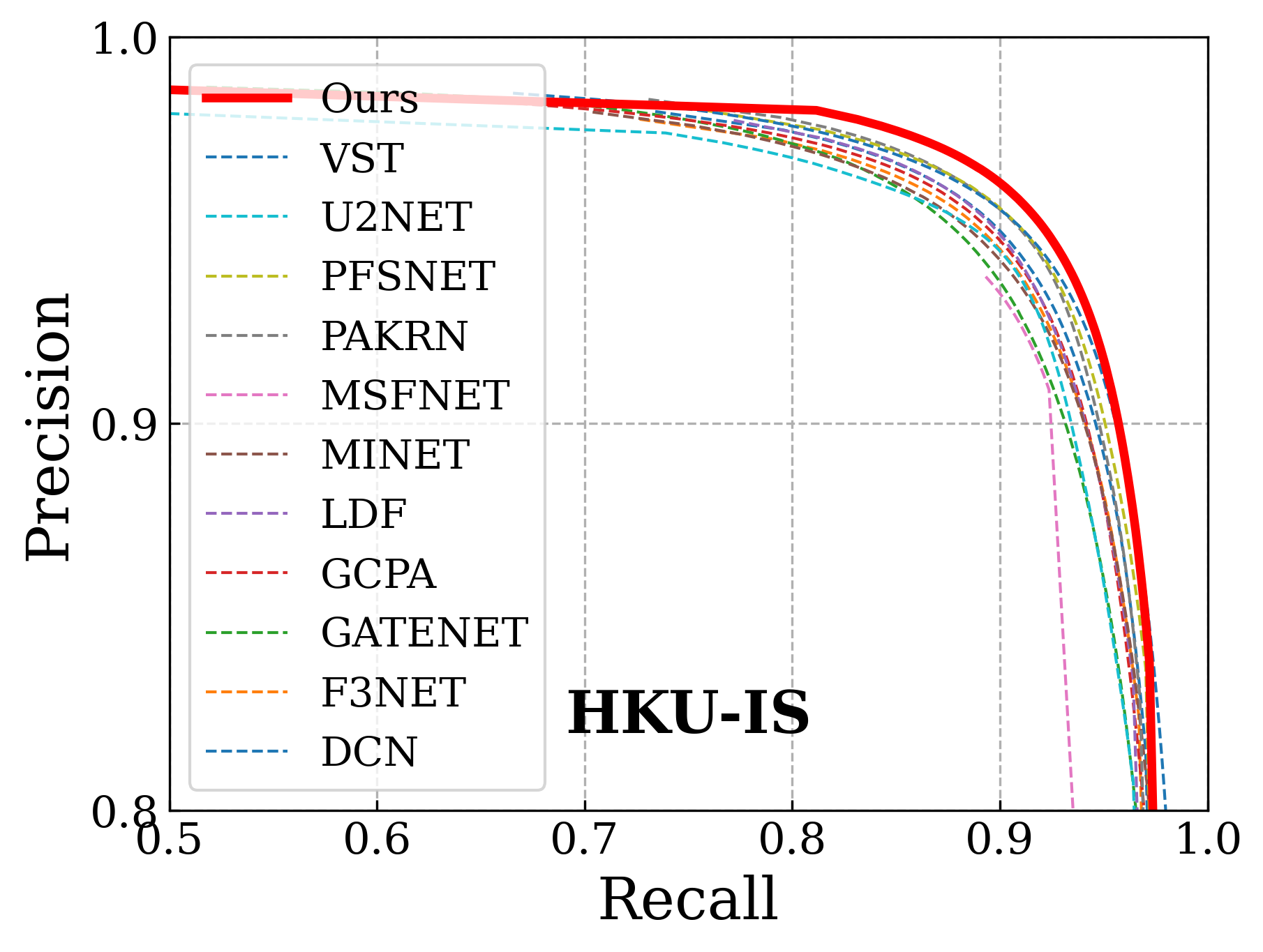} 
\includegraphics[width=\linewidth]{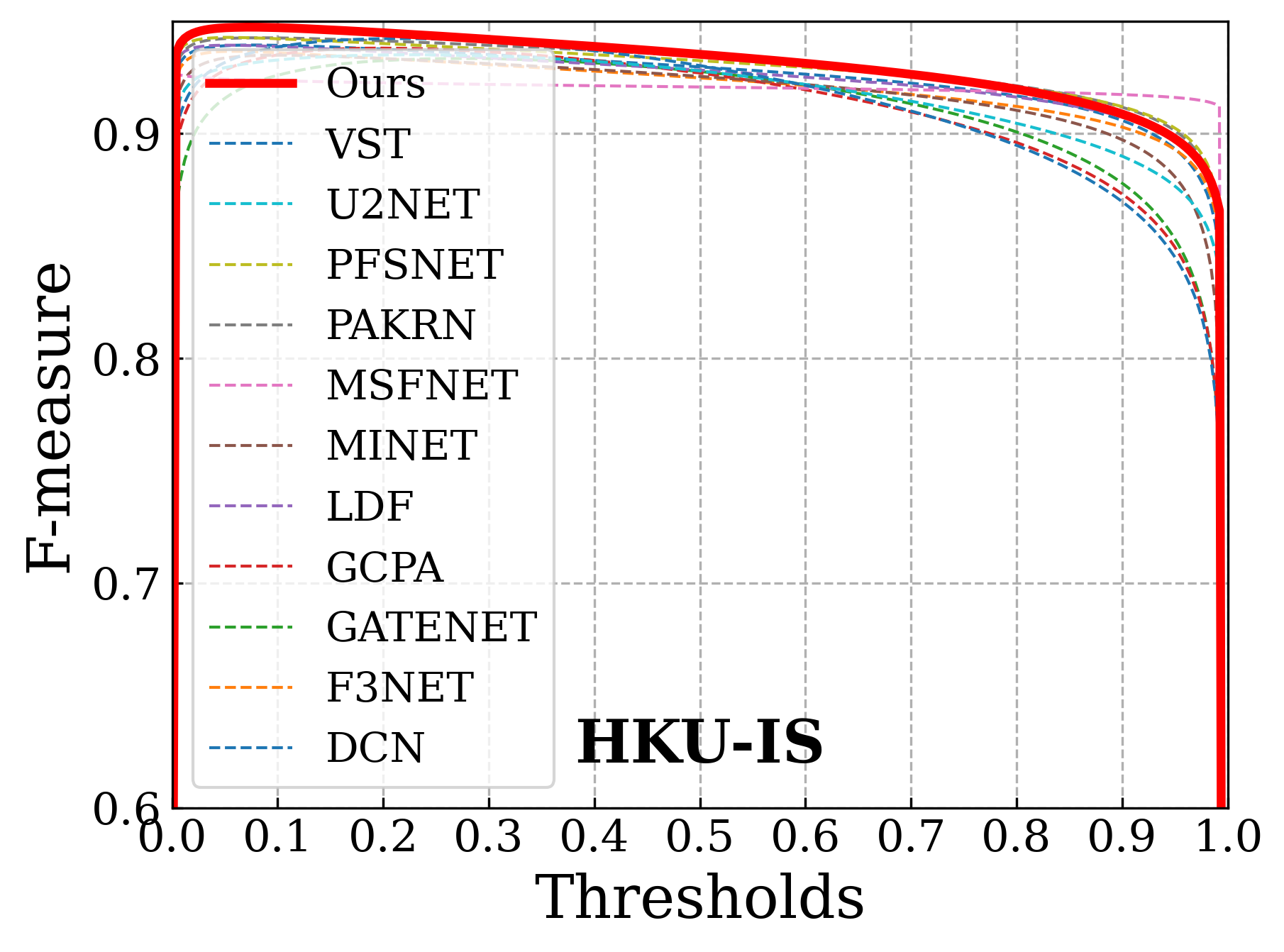}
\end{minipage}
\begin{minipage}{.19\linewidth}
\centering
\includegraphics[width=\linewidth]{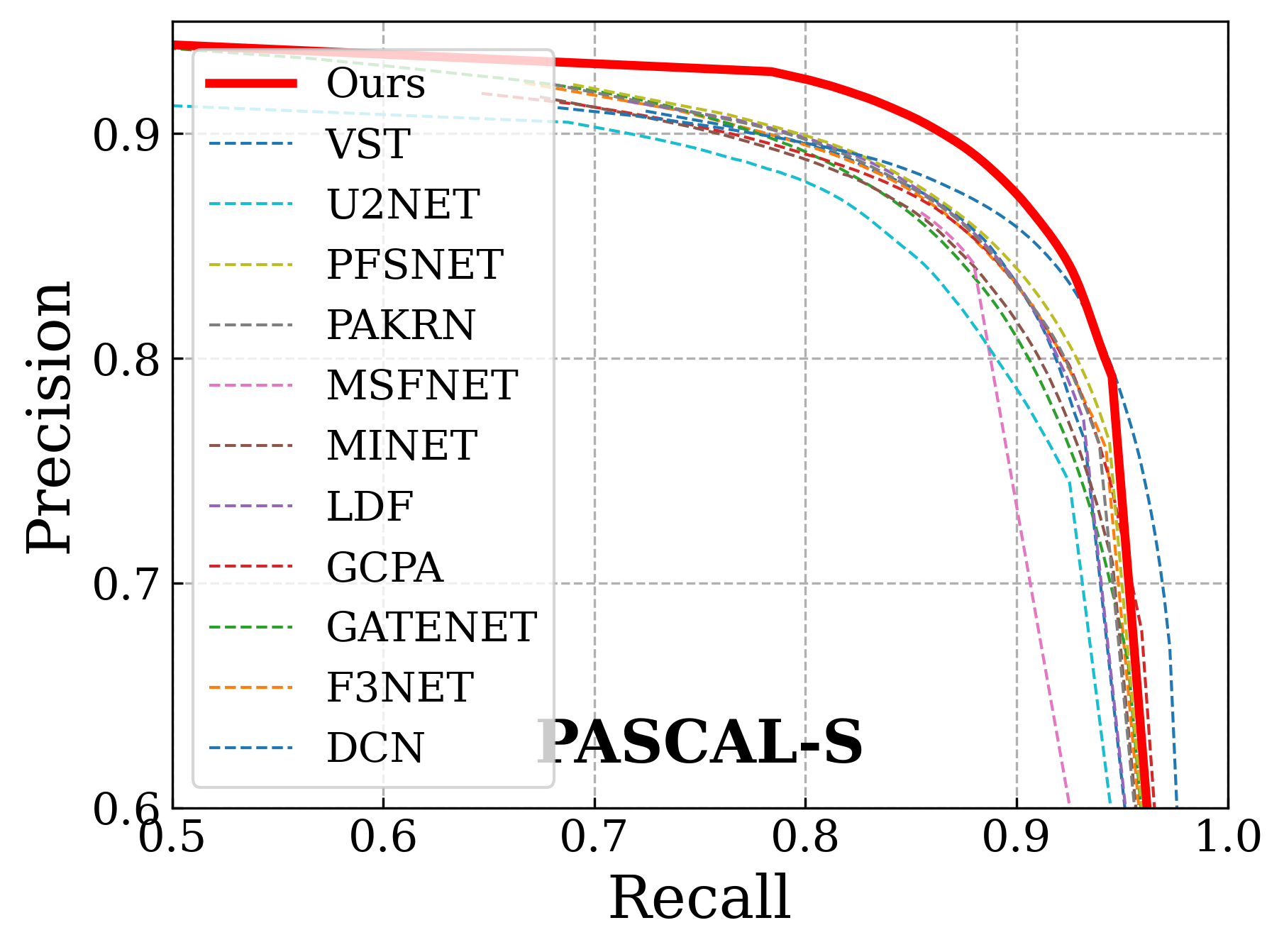} 
\includegraphics[width=\linewidth]{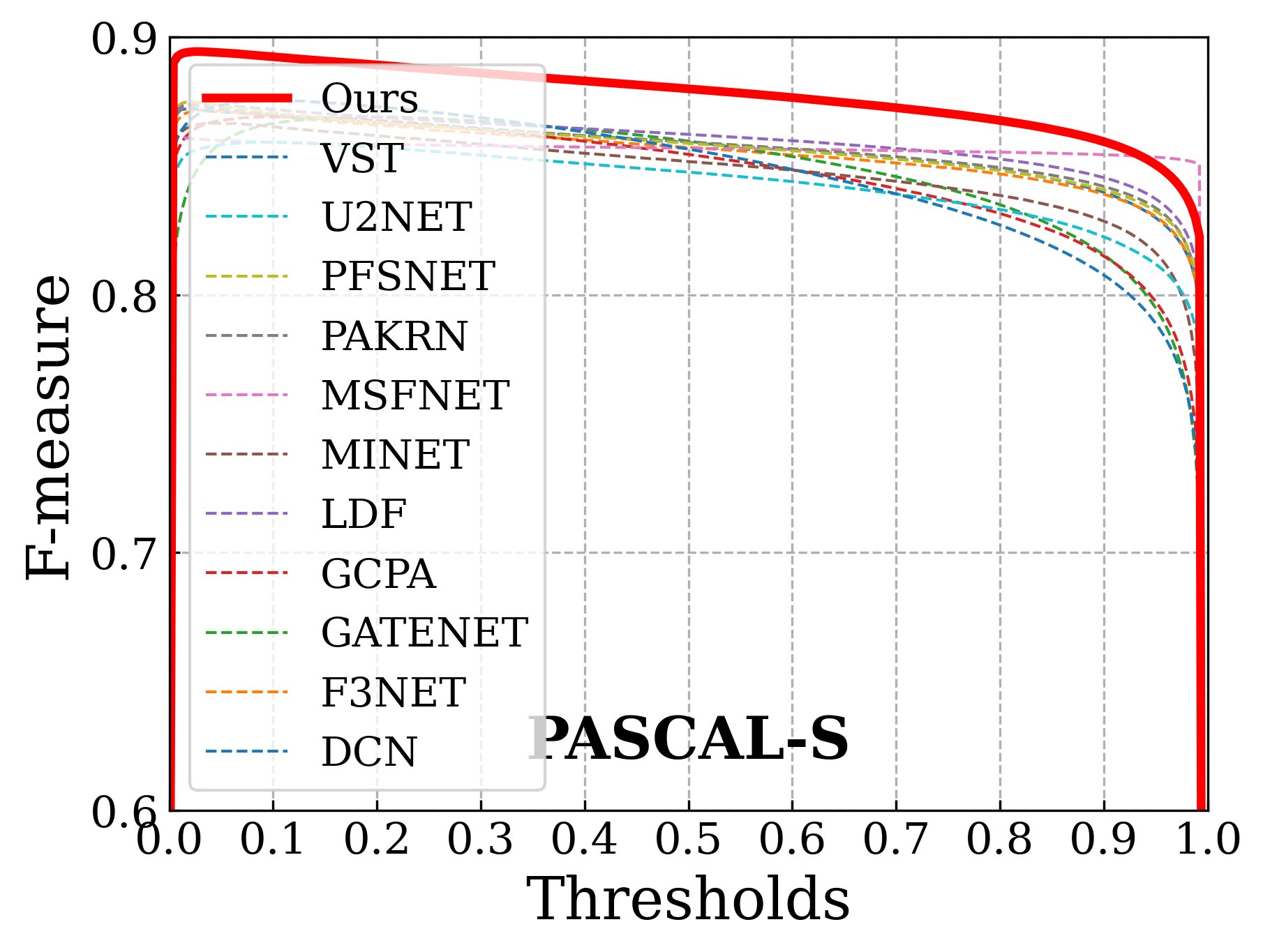}
\end{minipage}
\begin{minipage}{.19\linewidth}
\centering
\includegraphics[width=\linewidth]{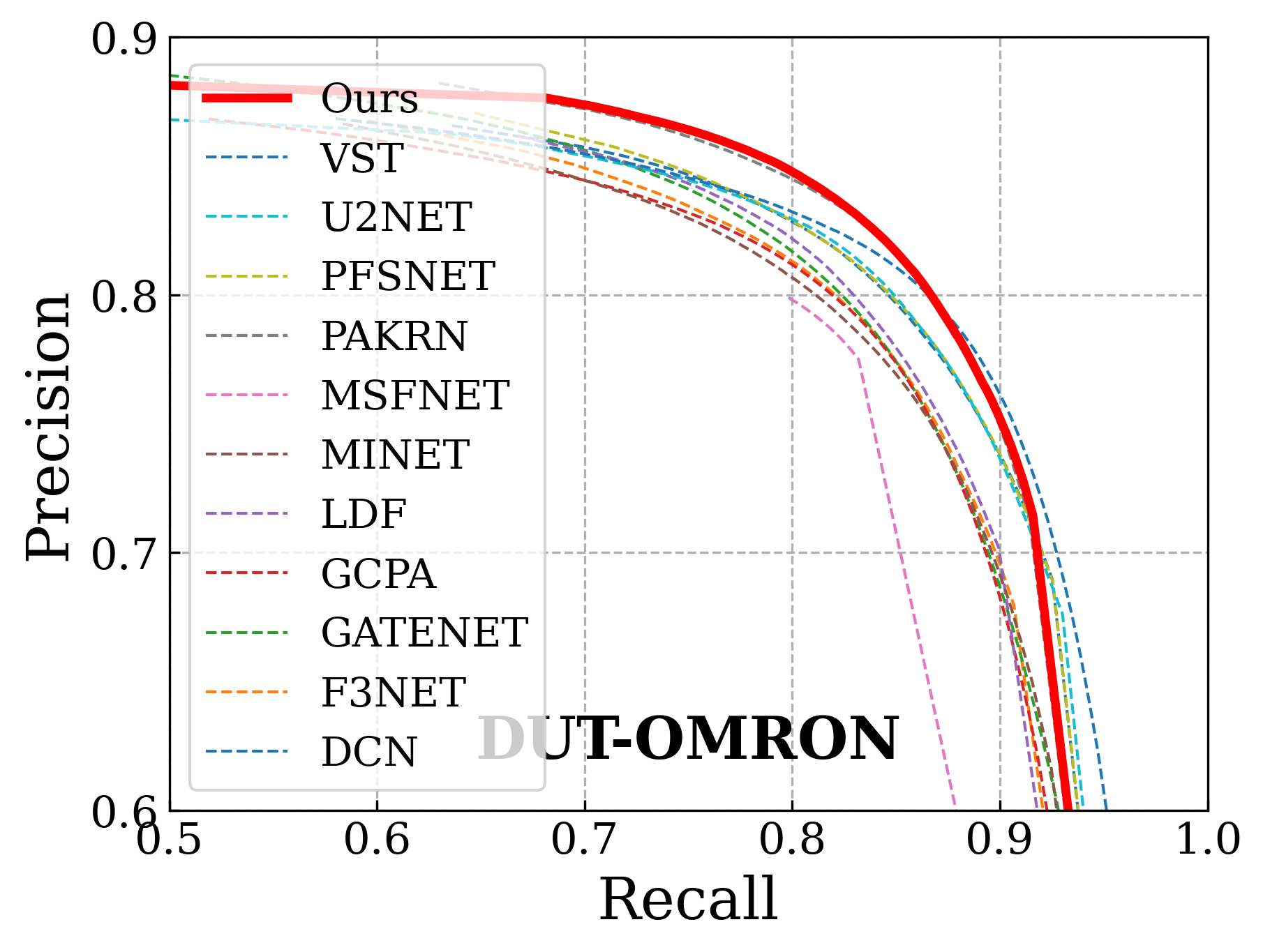} 
\includegraphics[width=\linewidth]{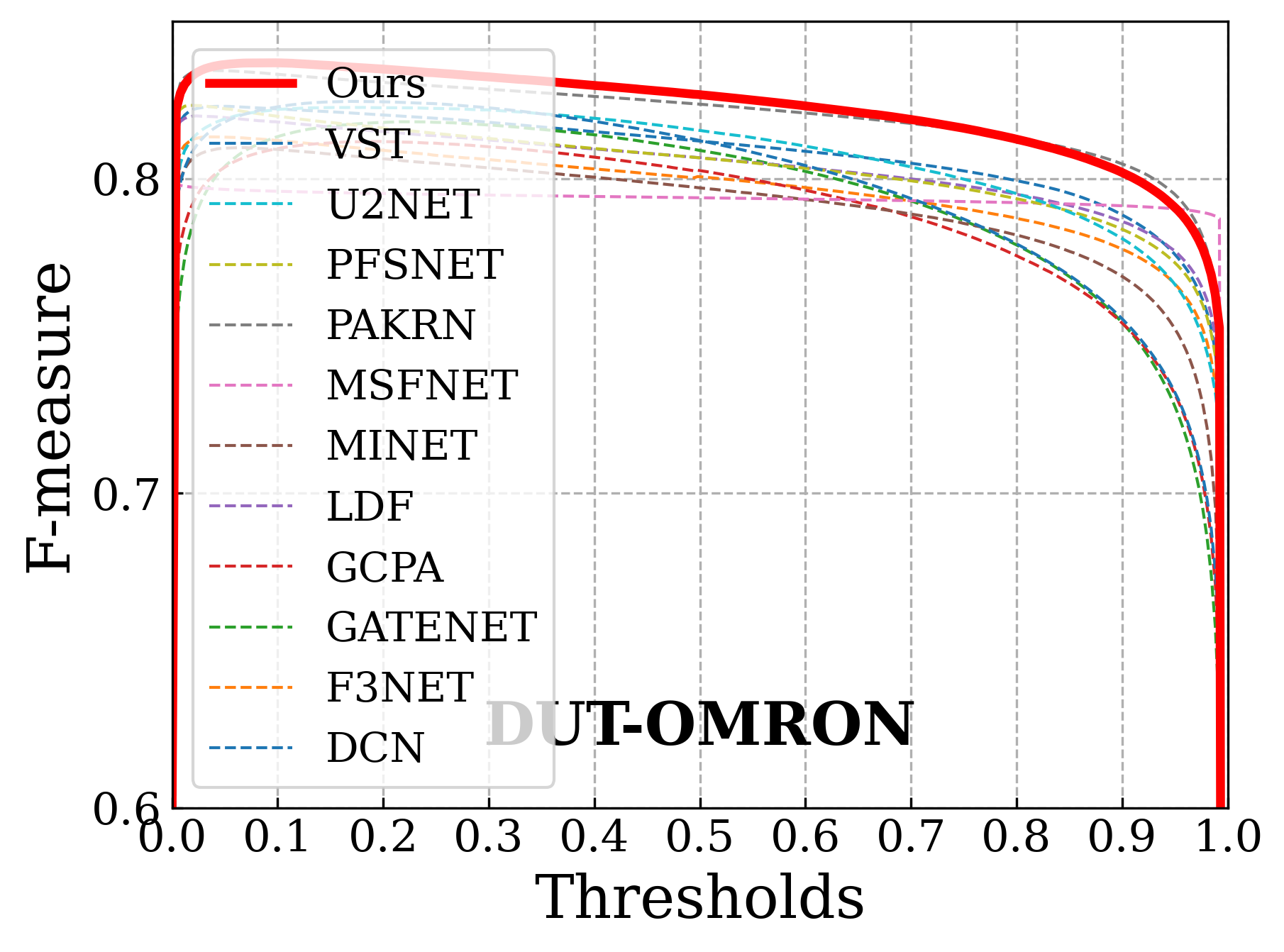}
\end{minipage}
\caption{Precision-Recall Curves (first row) and F-measure Curves (second row) comparison on five saliency benchmark datasets. As shown above, our network achieved the best results among all networks across five datasets.}
\label{fig:PR}
\vspace{-0.02\linewidth}
\end{figure*}

\subsection{Evaluation Datasets and Metrics}
\textbf{DUT-OMRON} \cite{DUTO}(5168 images), \textbf{ECSSD} \cite{ECSSD}(1000 images), \textbf{PASCAL-S} \cite{PASCALS}(850 images), \textbf{HKUIS} \cite{HKUIS}(4447 images), and \textbf{DUTS-TE} \cite{DUTSTE}(5019 Images) are our evaluation datasets, and the evaluation metrics are as follows:


\textbf{F\textsubscript{$\beta$}-measure}. The F\textsubscript{$\beta$}-measure\cite{FBETA} is calculated based on the precision and recall value of saliency maps:
$F_{\beta} = \frac{(1+\beta^2)\times Precision \times Recall}{\beta^2 \times Precision + Recall}$
where $\beta^2$ is set to 0.3 \cite{FBETA}.

\textbf{MAE}. MAE is the mean absolute element-wise differences between ground truths $\hat{y}$ and predictions $x$:  $MAE = \frac{1}{n}\sum^{n}_{i=1}|x_i - \hat{y}_i|$.

\textbf{E-measure}. By using local pixel values and the image-wise mean, $E_{\xi}$\cite{EMEASURE} calculates the similarity between the ground truth and predictions.

\textbf{S-measure}. $S_{\alpha}$\cite{SMEASURE} aims to measure the region and object level of structural similarities between the ground truth and the prediction, denoted as $S_{o}$ and $S_{r}$. It is defined as $S_{\alpha} = \alpha S_{o} + (1-\alpha)S_{r}$ with $\alpha$ = 0.5.

\subsection{Comparisons with state-of-the-art}
We compare our method against 11 state-of-the-art networks in the field, namely, \textbf{F\textsuperscript{3}Net} \cite{F3NET}, \textbf{GateNet} \cite{GATENET}, \textbf{GCPA} \cite{GCPA}, \textbf{MINet} \cite{MINET}, \textbf{U\textsuperscript{2}Net} \cite{U2NET}, \textbf{LDF} \cite{LDF}, \textbf{MSFNet} \cite{MSFNET}, \textbf{PFSNet} \cite{PFSNET}, \textbf{DCN} \cite{DCN}, \textbf{PAKRN} \cite{PAKRN}, and \textbf{VST} \cite{VST}. Results were calculated using the code provided by F\textsuperscript{3}Net.
\subsubsection{Quantitative Evaluation.}
As shown in Table.\ref{metric_result}, our network achieved the best results in all metrics calculated across the five benchmark datasets. It demonstrates outstanding performances of the proposed SelfReformer. Besides, Fig.\ref{fig:PR} shows the precision-recall curve of the above-listed networks, and our network consistently outperformed all other methods.

\begin{figure*}[!t]
\centering
\begin{subfigure}{\linewidth}
\centering
\newcommand\id{1769}
\includegraphics[width=.07\linewidth]{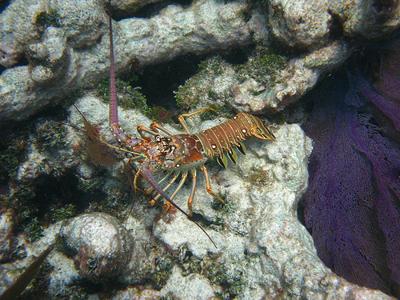}  
\includegraphics[width=.07\linewidth]{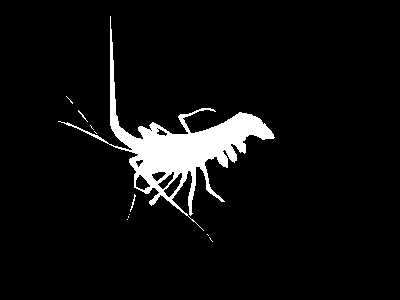}  
\includegraphics[width=.07\linewidth]{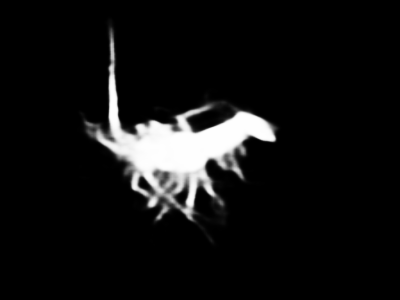}  
\includegraphics[width=.07\linewidth]{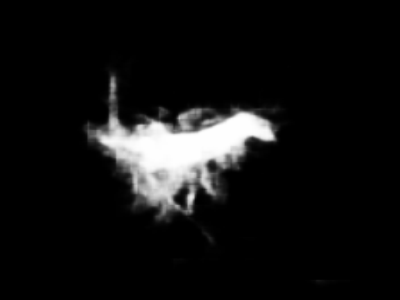}  
\includegraphics[width=.07\linewidth]{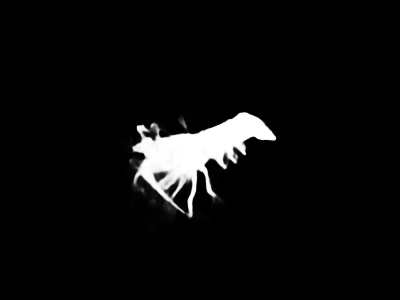}  
\includegraphics[width=.07\linewidth]{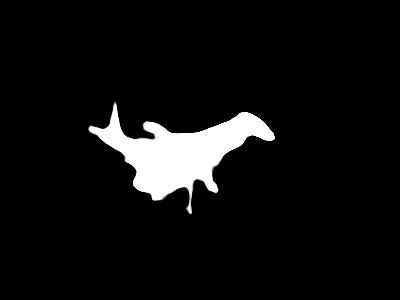}  
\includegraphics[width=.07\linewidth]{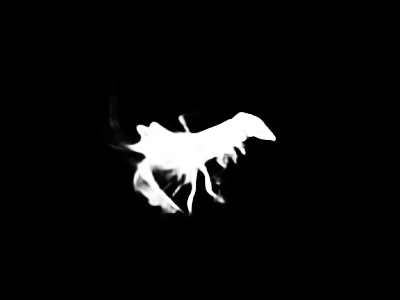}  
\includegraphics[width=.07\linewidth]{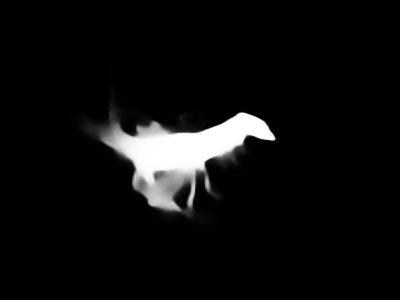}  
\includegraphics[width=.07\linewidth]{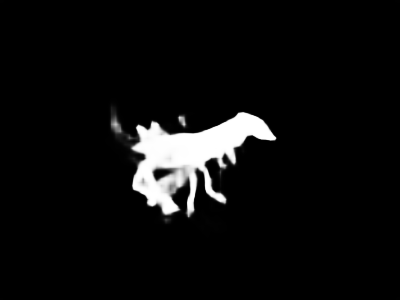}  
\includegraphics[width=.07\linewidth]{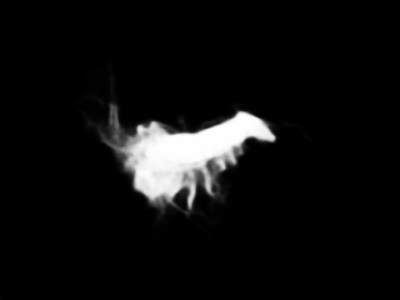}  
\includegraphics[width=.07\linewidth]{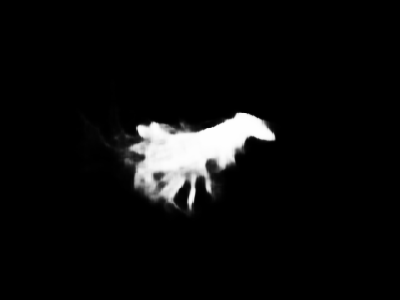}  
\includegraphics[width=.07\linewidth]{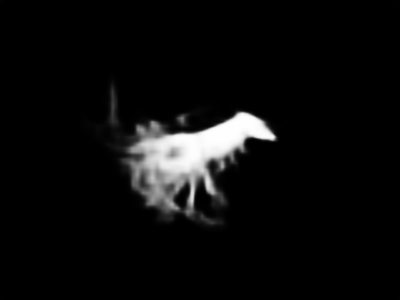}  
\includegraphics[width=.07\linewidth]{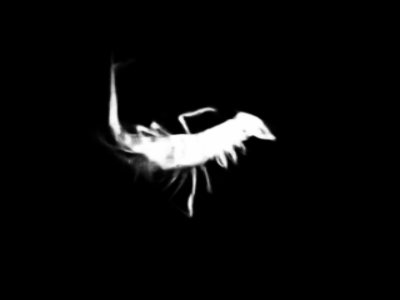}  
\end{subfigure}
\begin{subfigure}{\linewidth}
\centering
\newcommand\id{2103}
\includegraphics[width=.07\linewidth]{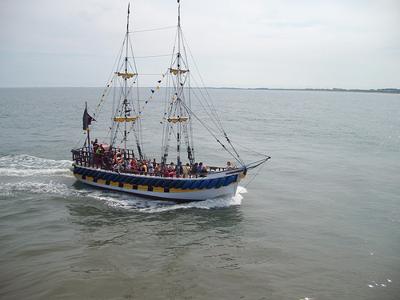}  
\includegraphics[width=.07\linewidth]{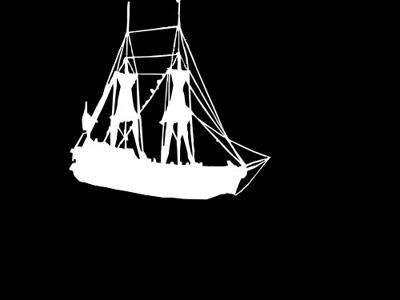}  
\includegraphics[width=.07\linewidth]{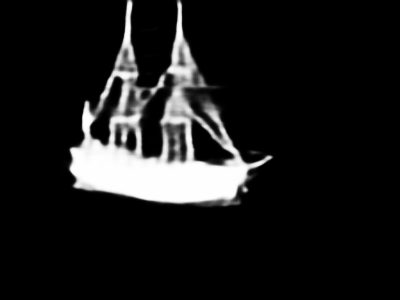}  
\includegraphics[width=.07\linewidth]{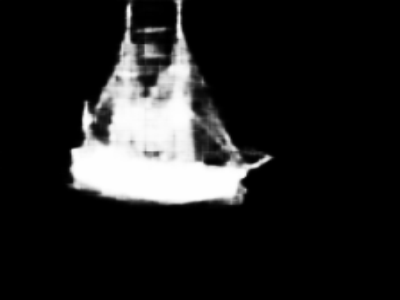}  
\includegraphics[width=.07\linewidth]{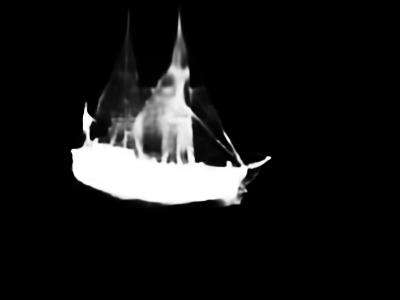}  
\includegraphics[width=.07\linewidth]{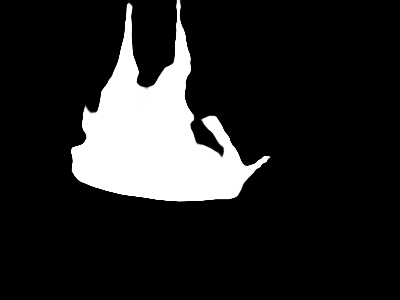}  
\includegraphics[width=.07\linewidth]{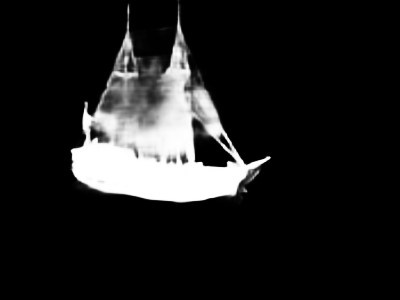}  
\includegraphics[width=.07\linewidth]{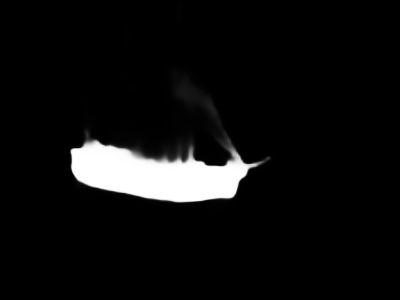}  
\includegraphics[width=.07\linewidth]{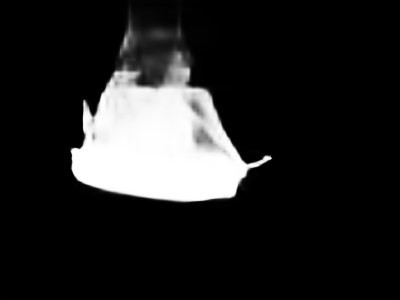}  
\includegraphics[width=.07\linewidth]{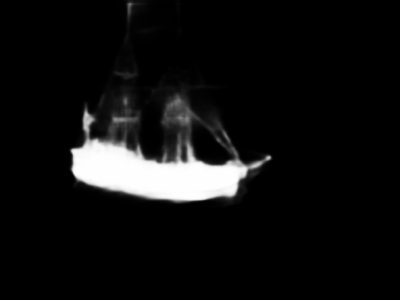}  
\includegraphics[width=.07\linewidth]{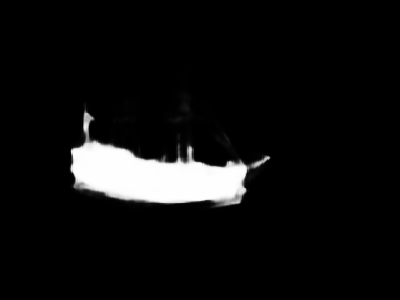}  
\includegraphics[width=.07\linewidth]{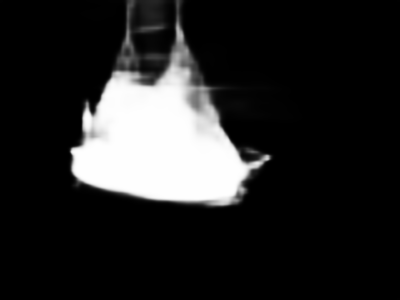}  
\includegraphics[width=.07\linewidth]{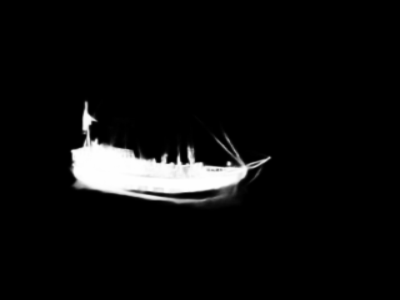}  
\end{subfigure}
\begin{subfigure}{\linewidth}
\centering
\newcommand\id{1106}
\includegraphics[width=.07\linewidth]{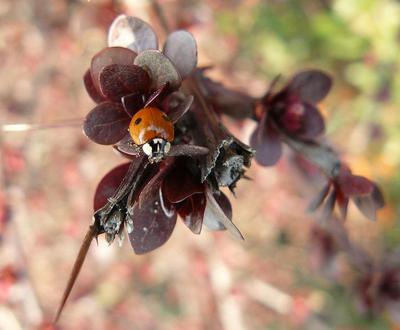}  
\includegraphics[width=.07\linewidth]{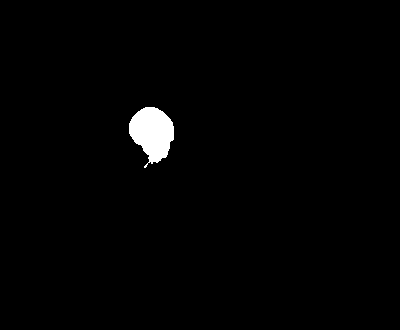}  
\includegraphics[width=.07\linewidth]{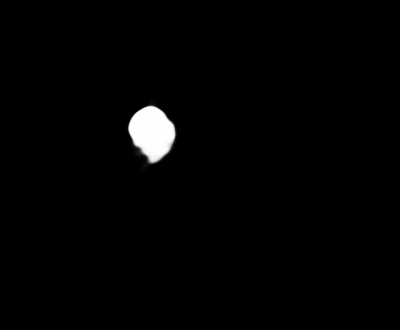}  
\includegraphics[width=.07\linewidth]{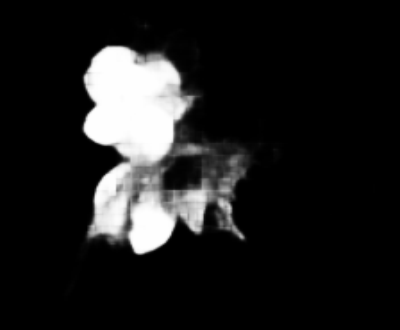}  
\includegraphics[width=.07\linewidth]{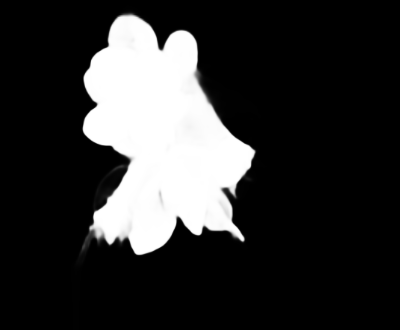}  
\includegraphics[width=.07\linewidth]{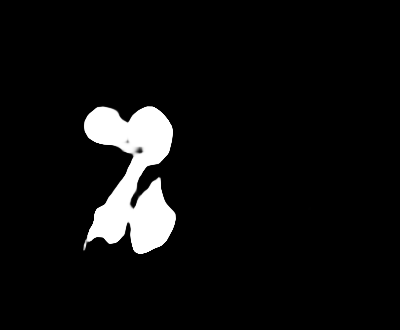}  
\includegraphics[width=.07\linewidth]{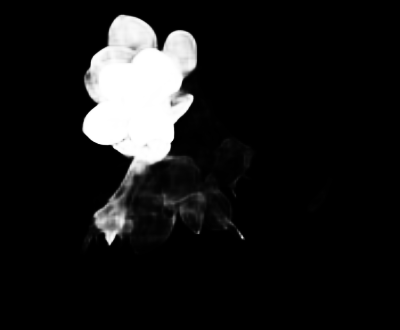}  
\includegraphics[width=.07\linewidth]{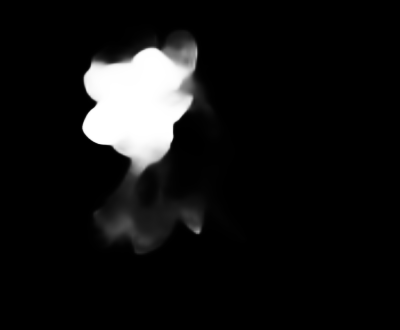}  
\includegraphics[width=.07\linewidth]{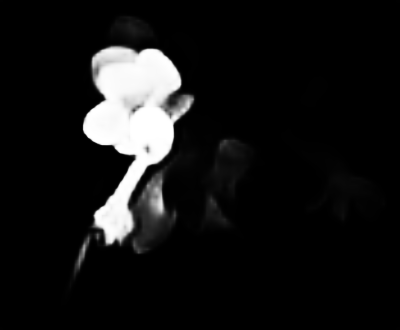}  
\includegraphics[width=.07\linewidth]{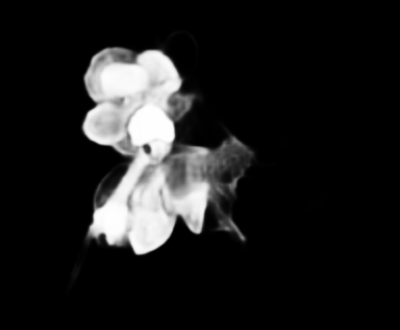}  
\includegraphics[width=.07\linewidth]{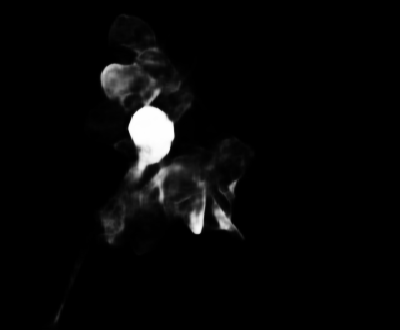}  
\includegraphics[width=.07\linewidth]{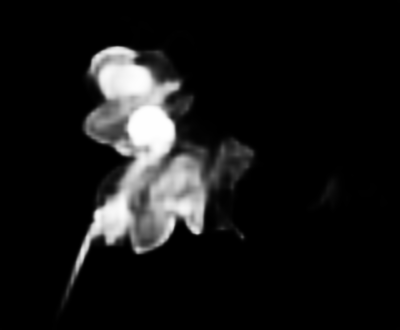}  
\includegraphics[width=.07\linewidth]{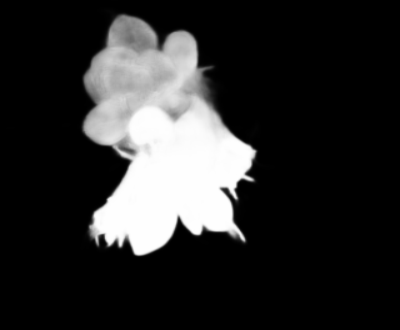} 
\end{subfigure}
\begin{subfigure}{\linewidth}
\centering
\newcommand\id{1502}
\includegraphics[width=.07\linewidth]{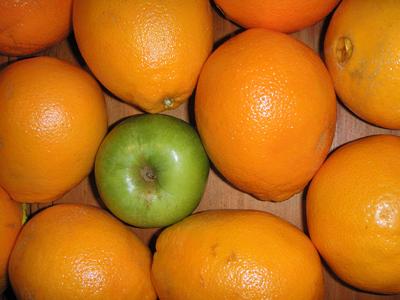}  
\includegraphics[width=.07\linewidth]{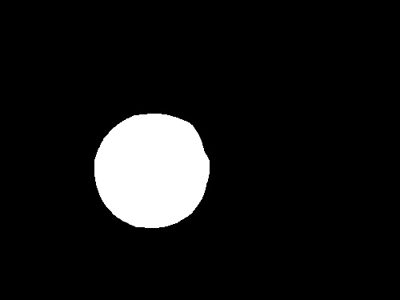}  
\includegraphics[width=.07\linewidth]{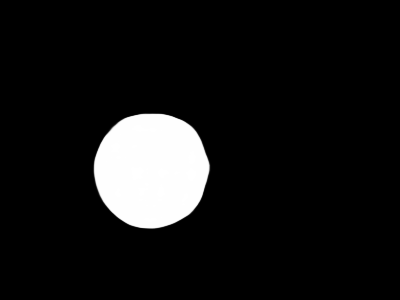}  
\includegraphics[width=.07\linewidth]{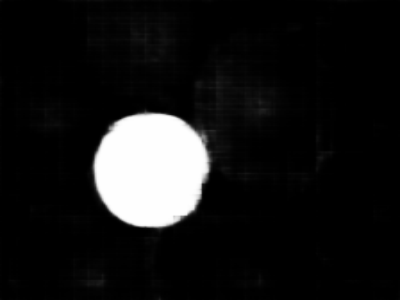}  
\includegraphics[width=.07\linewidth]{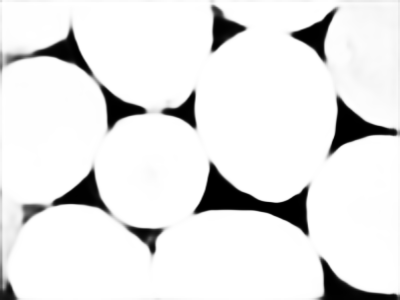}  
\includegraphics[width=.07\linewidth]{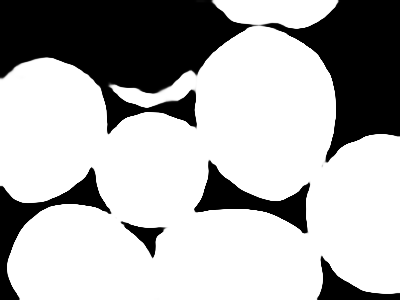}  
\includegraphics[width=.07\linewidth]{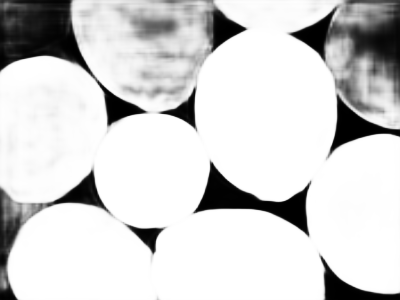}  
\includegraphics[width=.07\linewidth]{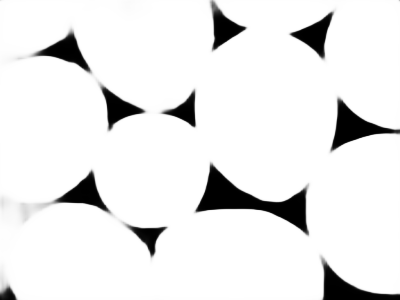}  
\includegraphics[width=.07\linewidth]{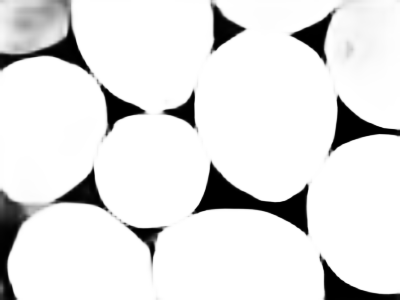}  
\includegraphics[width=.07\linewidth]{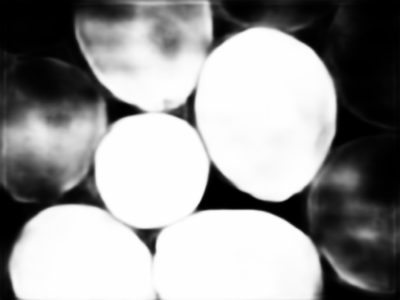}  
\includegraphics[width=.07\linewidth]{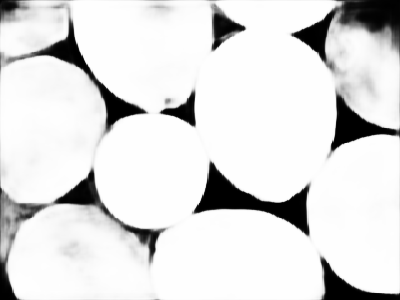}  
\includegraphics[width=.07\linewidth]{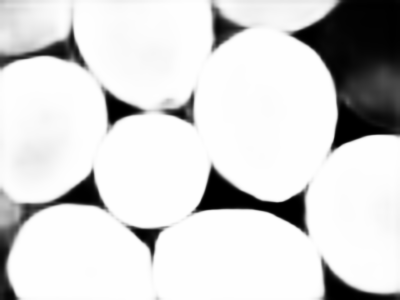}  
\includegraphics[width=.07\linewidth]{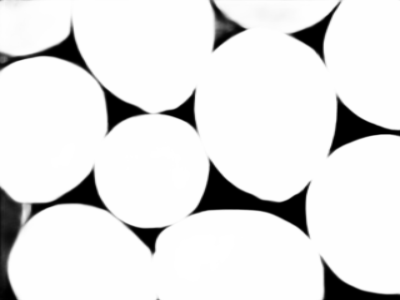}  
\end{subfigure}
\begin{subfigure}{\linewidth}
\centering
\newcommand\id{2404}
\includegraphics[width=.07\linewidth]{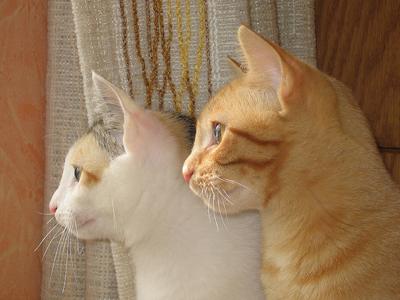}  
\includegraphics[width=.07\linewidth]{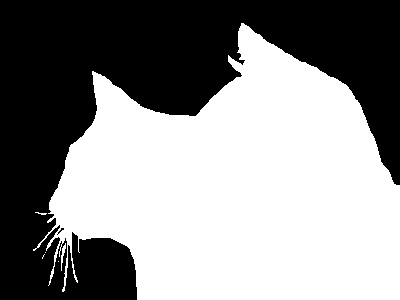}  
\includegraphics[width=.07\linewidth]{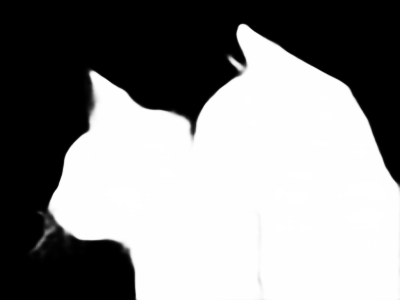}  
\includegraphics[width=.07\linewidth]{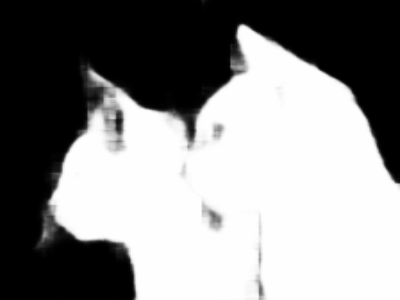}  
\includegraphics[width=.07\linewidth]{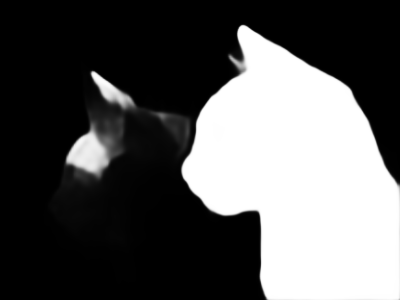}  
\includegraphics[width=.07\linewidth]{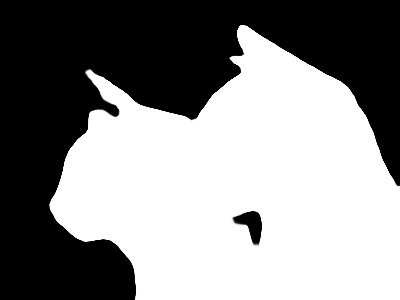}  
\includegraphics[width=.07\linewidth]{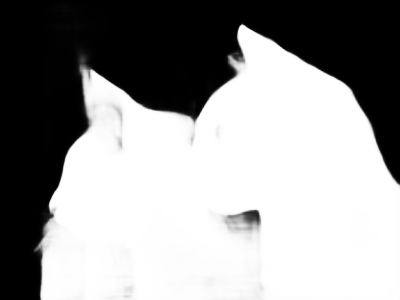}  
\includegraphics[width=.07\linewidth]{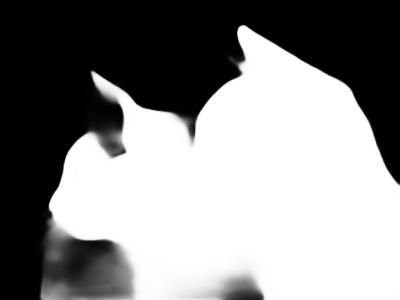}  
\includegraphics[width=.07\linewidth]{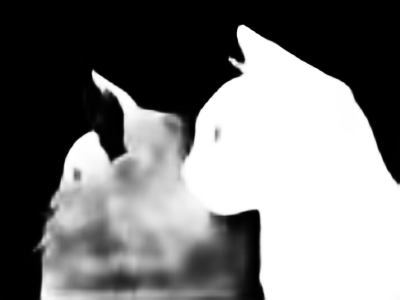}  
\includegraphics[width=.07\linewidth]{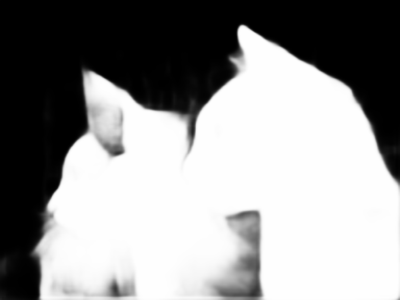}  
\includegraphics[width=.07\linewidth]{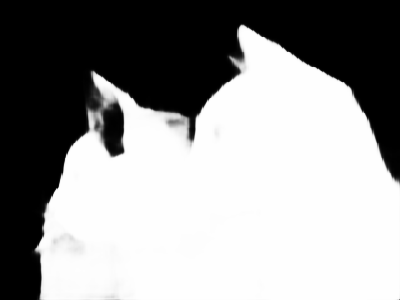}  
\includegraphics[width=.07\linewidth]{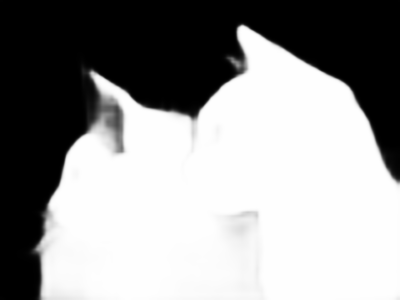}  
\includegraphics[width=.07\linewidth]{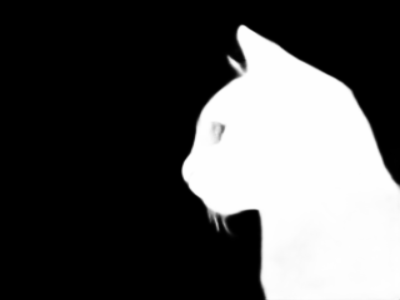}  
\end{subfigure}
\begin{subfigure}{\linewidth}
\centering
\newcommand\id{2019}
\includegraphics[width=.07\linewidth]{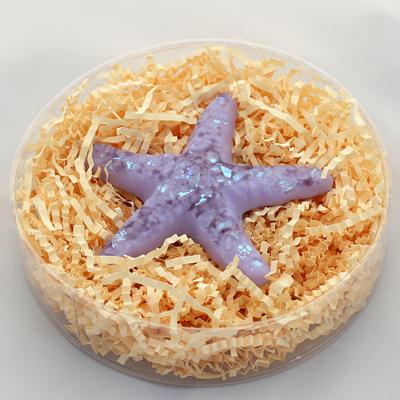}  
\includegraphics[width=.07\linewidth]{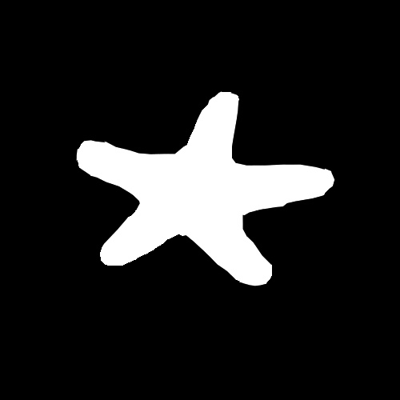}  
\includegraphics[width=.07\linewidth]{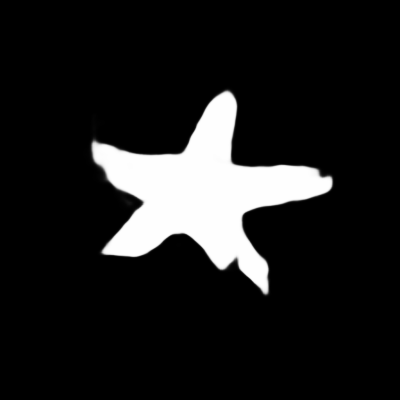}  
\includegraphics[width=.07\linewidth]{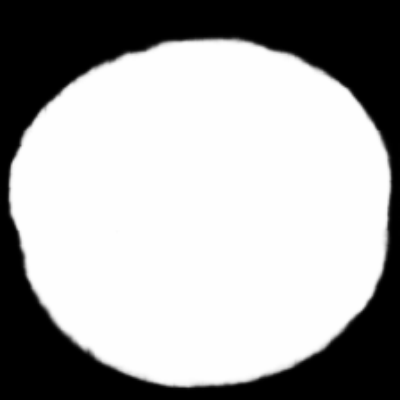}  
\includegraphics[width=.07\linewidth]{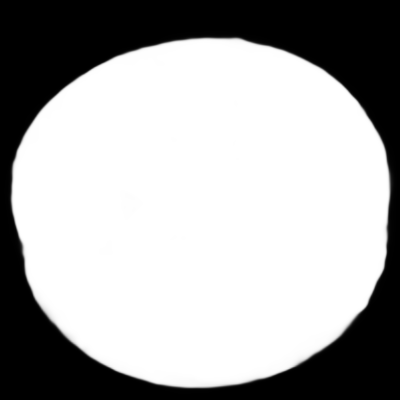}  
\includegraphics[width=.07\linewidth]{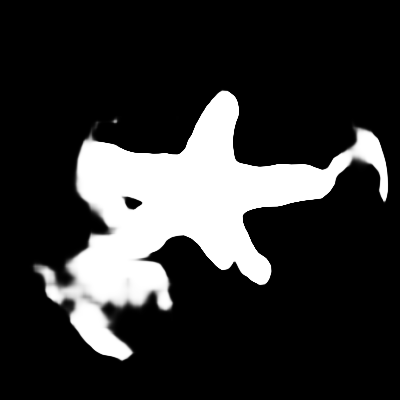}  
\includegraphics[width=.07\linewidth]{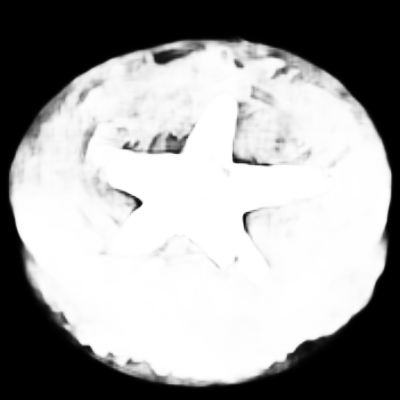}  
\includegraphics[width=.07\linewidth]{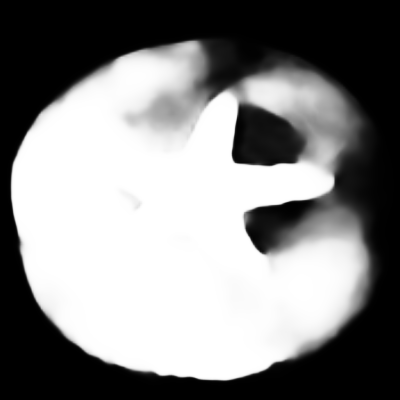}  
\includegraphics[width=.07\linewidth]{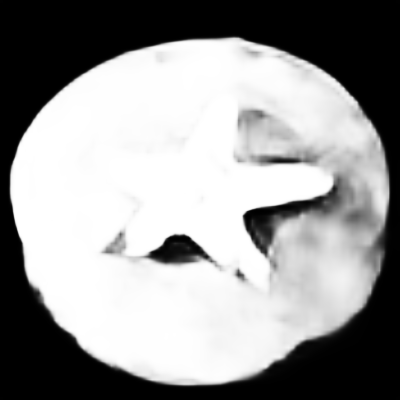}  
\includegraphics[width=.07\linewidth]{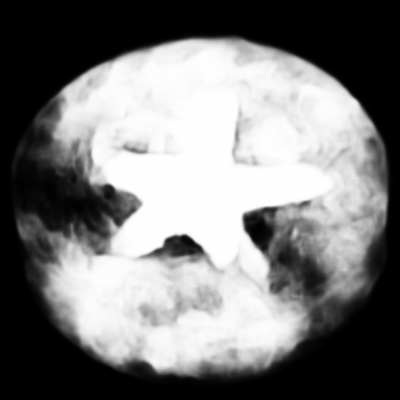}  
\includegraphics[width=.07\linewidth]{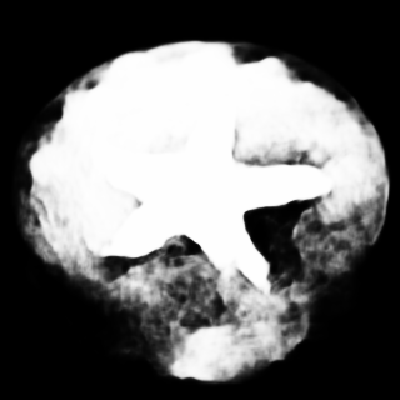}  
\includegraphics[width=.07\linewidth]{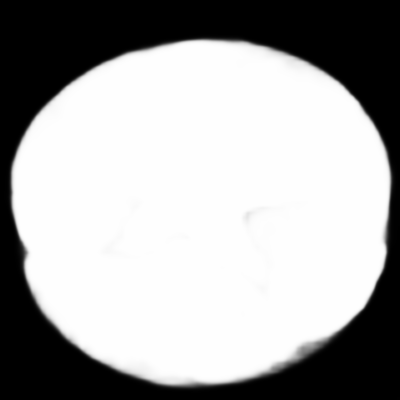}  
\includegraphics[width=.07\linewidth]{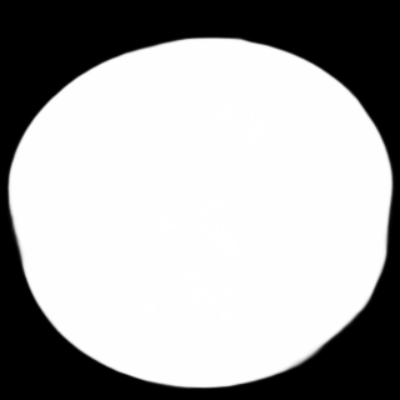}  
\end{subfigure}
\begin{minipage}{\linewidth}
{\footnotesize
\ \ \ \ \ \ \ Image
\ \ \ \ \ \ \ \ GT
\ \ \ \ \ \ \ \ Ours*
\ \ \ \ \ \ \ VST*
\ \ \ \ PAKRN
\ \ MSFNet
\ \ \ PFSNet
\ \ \ \ \ DCN
\ \ \ \ \ \ \ F\textsuperscript{3}Net
\ \ \ GATENet
\ \ \ MINet
\ \ \ \ \ GCPA
\ \ \ \ \ U\textsuperscript{2}Net
}
\end{minipage}
\caption{Visual comparisons between the proposed method and 10 state-of-the-art networks. * stands for Transformer based networks. More comparisons are listed in the supplementary material. Best view in zoom-in.}
\label{fig:visualization}
\vspace{-0.01\linewidth}
\end{figure*}

\subsubsection{Qualitative Evaluation.}
Visual comparisons are listed in Fig.\ref{fig:visualization}. Compared with other methods, our predictions are more accurate in structural completeness and contain richer details (rows 1, 2, and 6). Prediction completeness demonstrates the effectiveness of the global context branch, while rich details indicate the success of Pixel Shuffle and CRM. Moreover, our network excels in dealing with challenging scenarios like a small object among complex backgrounds (row 3), the unique object among its peers (row 4), and multiple salient objects (row 5).

\vspace{-0.01\linewidth}
\section{Ablation Studies}
We investigate the effectiveness of proposed modules and methods, i.e., global-context branch, CRM and Pixel Shuffle. For more ablation studies, please refer to supplementary materials.
\vspace{-0.01\linewidth}

\subsection{Effectiveness of Global-Context Branch}
We study the impact of the global-context branch by removing it and training the rest of the network, i.e., the first stage of CRM in Fig.\ref{fig:network} will no longer fuse $f_{g}$ with decoder features. The evaluation results on DUTS-TE and PASCAL-S are listed in Table \ref{table:GLC}, and we can observe significant improvement with the presence of a global context branch.
\begin{table}[!h]
\centering
\setlength\tabcolsep{1.25pt}
\caption{Quantitative comparisons for the effectiveness of global-context branch.}
\begin{tabular}{c|cccc|cccc} 
\hline
\hline
\multirow{2}{*}{} & \multicolumn{4}{c|}{\textbf{DUTS-TE}} & \multicolumn{4}{c}{\textbf{PASCAL-S}}  \\
& \textbf{$F_{\beta}\hspace{-1.5mm}\uparrow$} & \textbf{$M\hspace{-1.5mm}\downarrow$} & \textbf{$E_{\xi}\hspace{-1.5mm}\uparrow$} & \textbf{$S_{\alpha}\hspace{-1.5mm}\uparrow$}
& \textbf{$F_{\beta}\hspace{-1.5mm}\uparrow$} & \textbf{$M\hspace{-1.5mm}\downarrow$} & \textbf{$E_{\xi}\hspace{-1.5mm}\uparrow$} & \textbf{$S_{\alpha}\hspace{-1.5mm}\uparrow$}\\ 
\hline
w/o global context  & .912             & .028               & .914              & .904            & .892            & .053                 & .869            & .867          \\
w/ global context   &\textbf{.916}   &\textbf{.026}    &\textbf{.920}   &\textbf{.911}  &\textbf{.894} &\textbf{.050} &\textbf{.872} &\textbf{.874}    \\
\hline
\hline
\end{tabular}
\label{table:GLC}
\vspace{-0.04\linewidth}
\end{table}

\subsection{Effectiveness of CRM}
\begin{figure}[!h]
\centering
\begin{subfigure}{.18\linewidth}
\centering
\includegraphics[width=\linewidth]{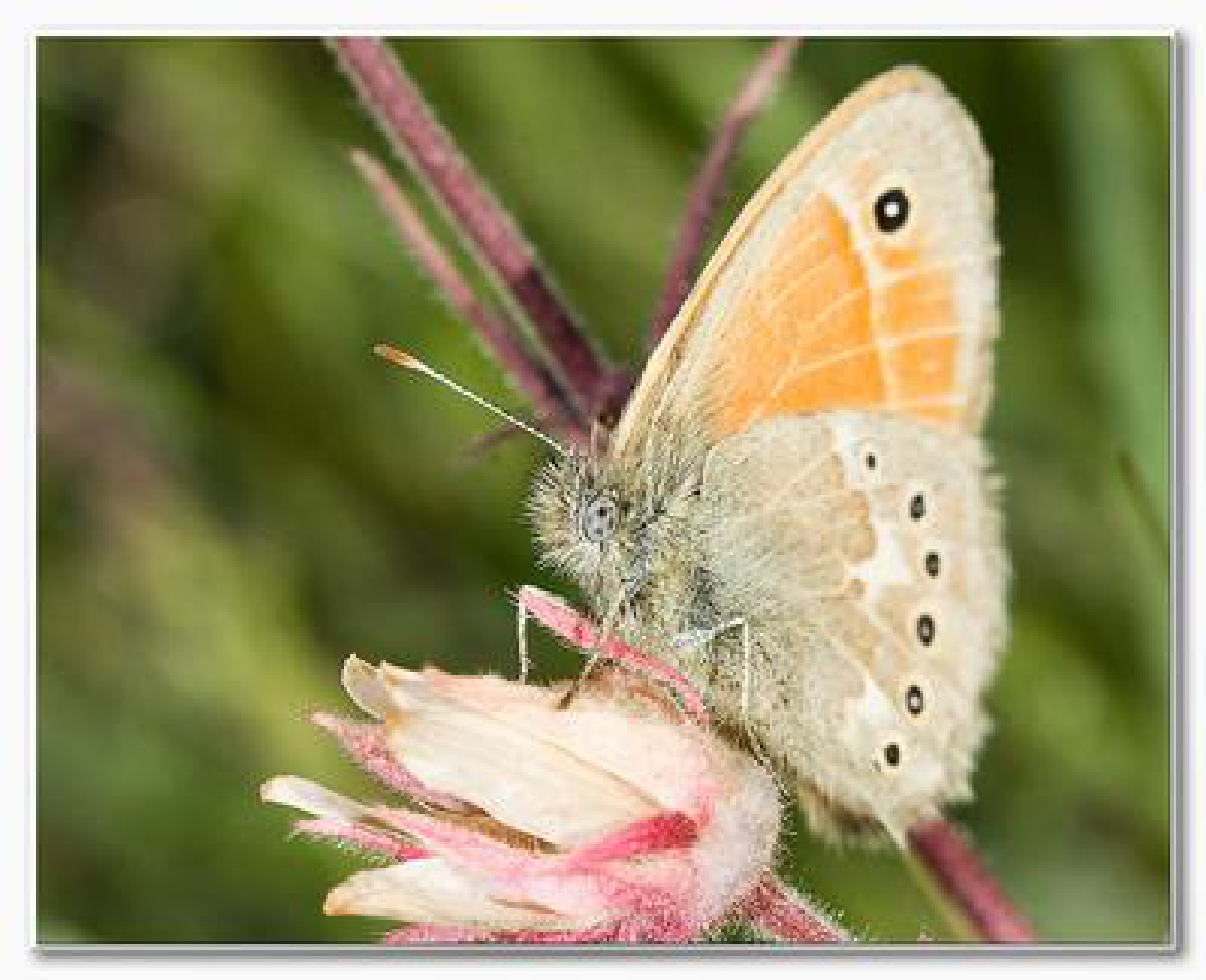}
\includegraphics[width=\linewidth]{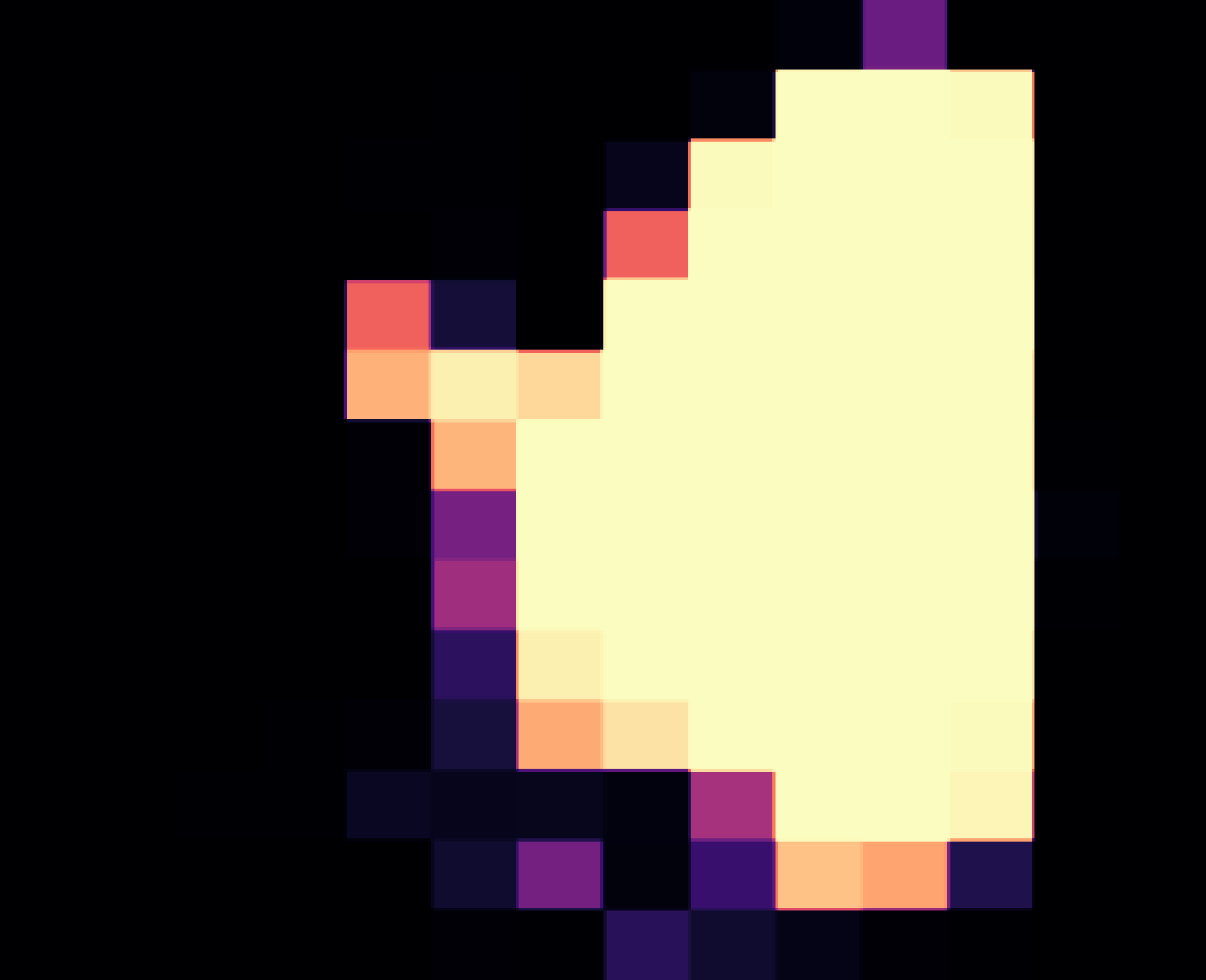}
\includegraphics[width=\linewidth]{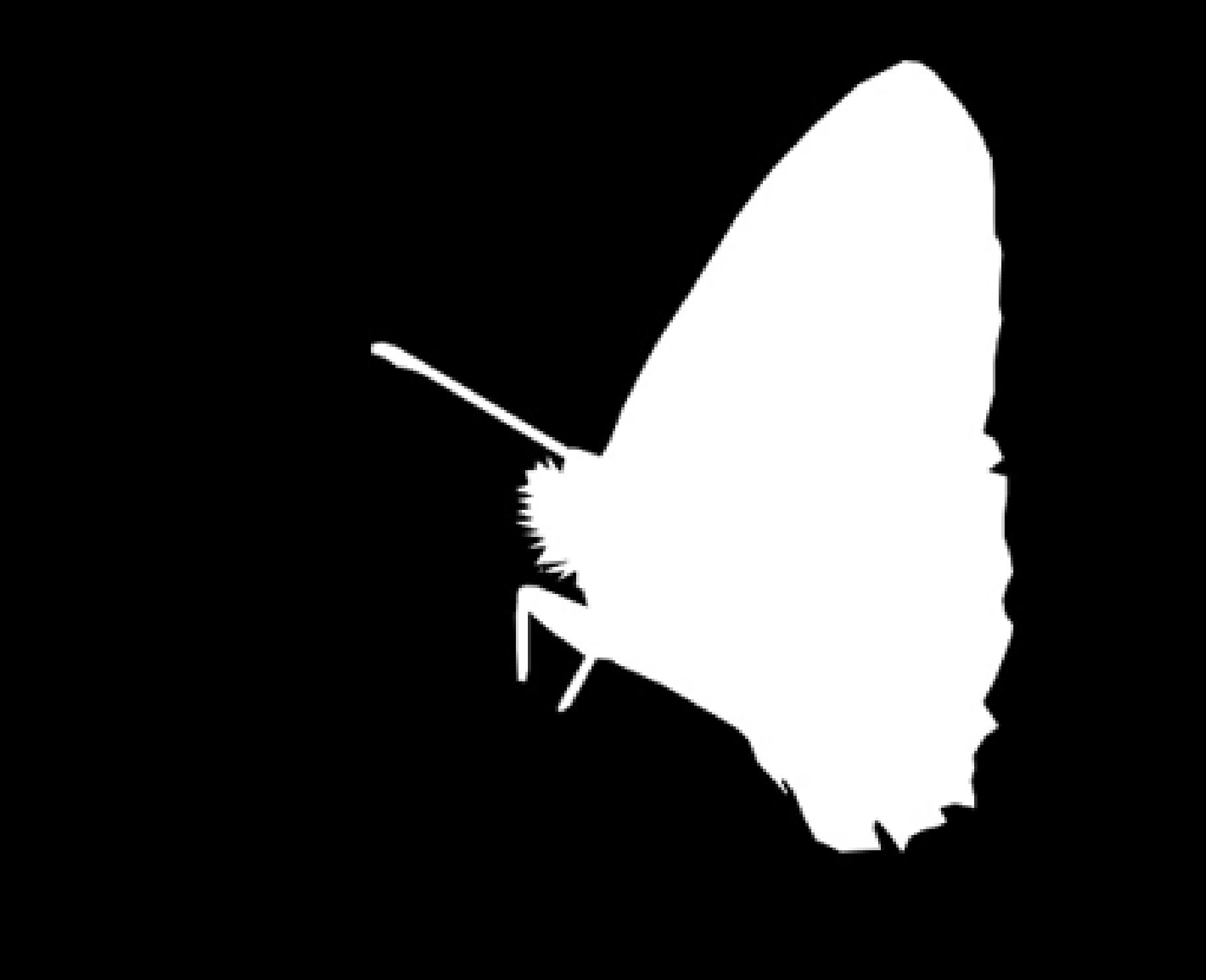}
\caption{}
\end{subfigure}
\begin{subfigure}{.18\linewidth}
\centering
\begin{overpic}[width=\linewidth]{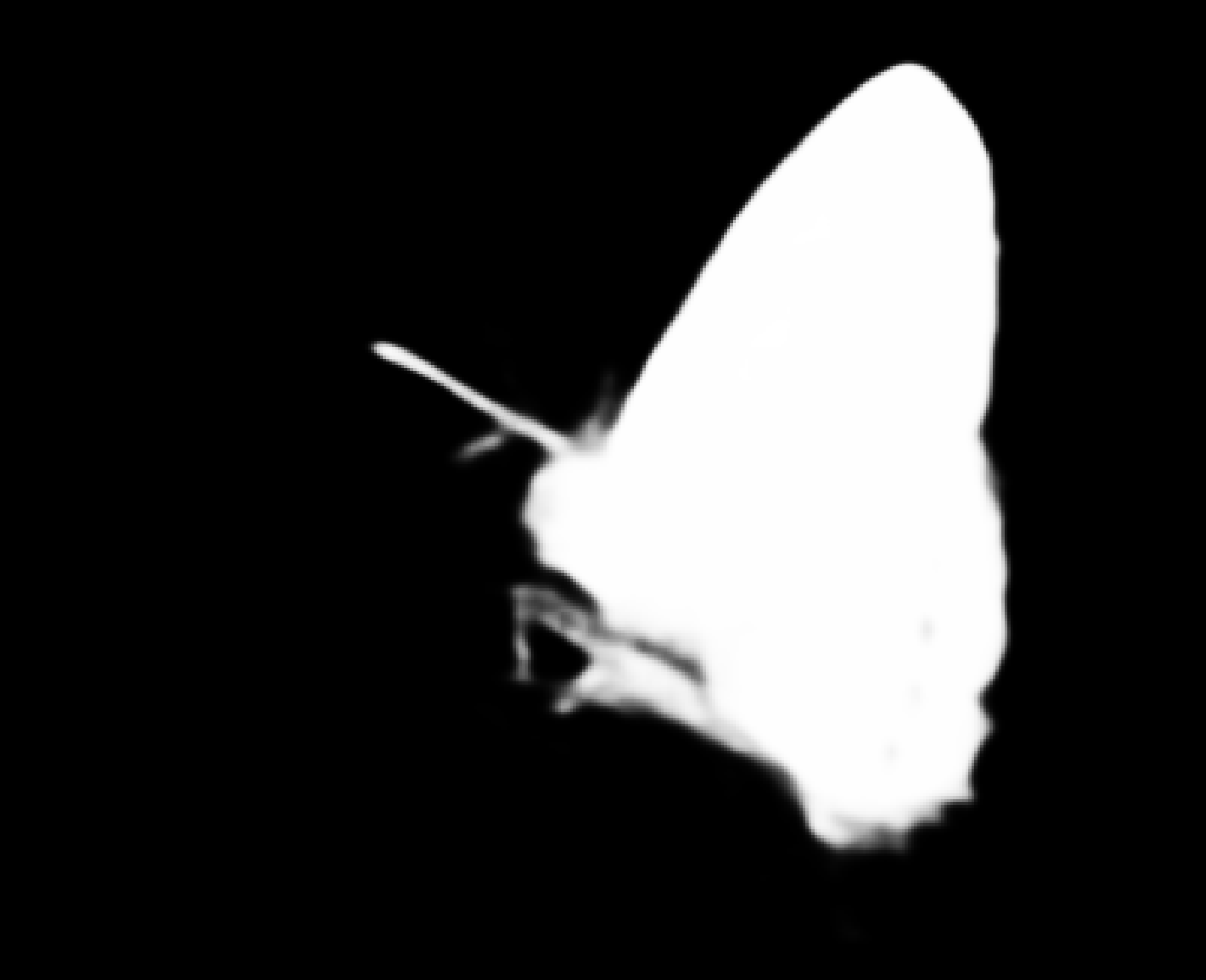} \put (5,63) {\textcolor{white}{\footnotesize.0159}}\end{overpic}
\includegraphics[width=\linewidth]{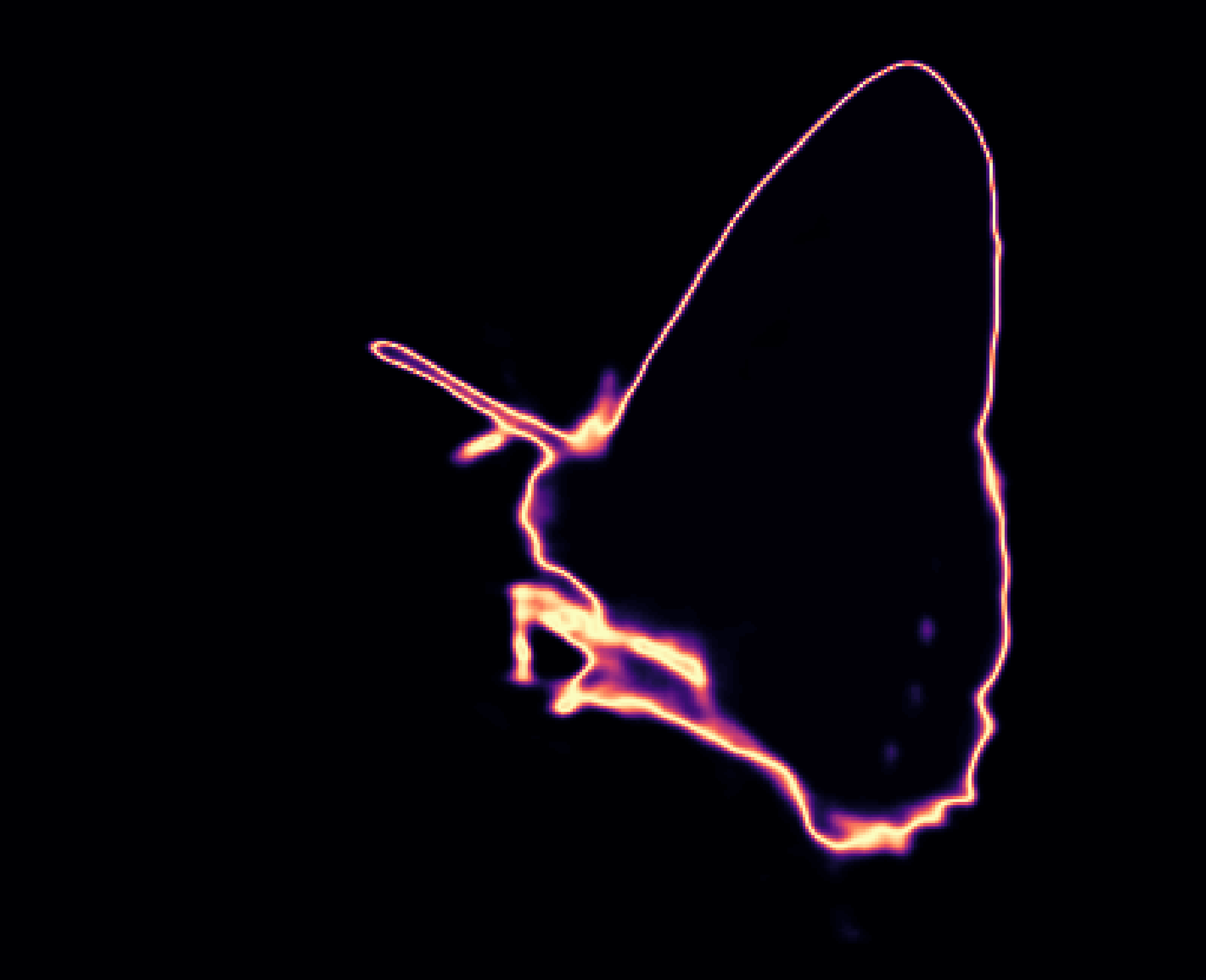}
\begin{overpic}[width=\linewidth]{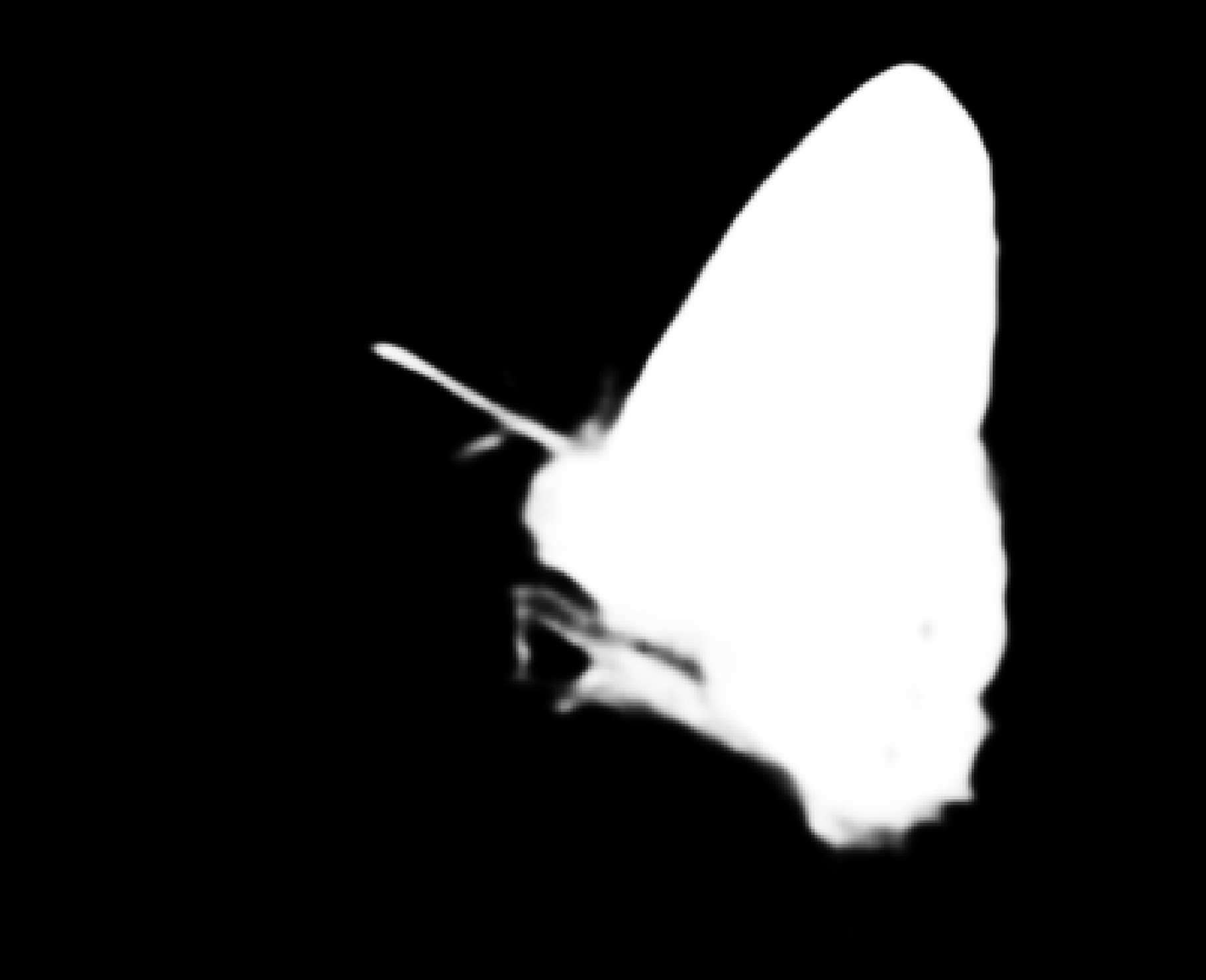} \put (5,63) {\textcolor{white}{\footnotesize.0158}}\end{overpic}
\caption{Stage 4}
\end{subfigure}
\begin{subfigure}{.18\linewidth}
\centering
\begin{overpic}[width=\linewidth]{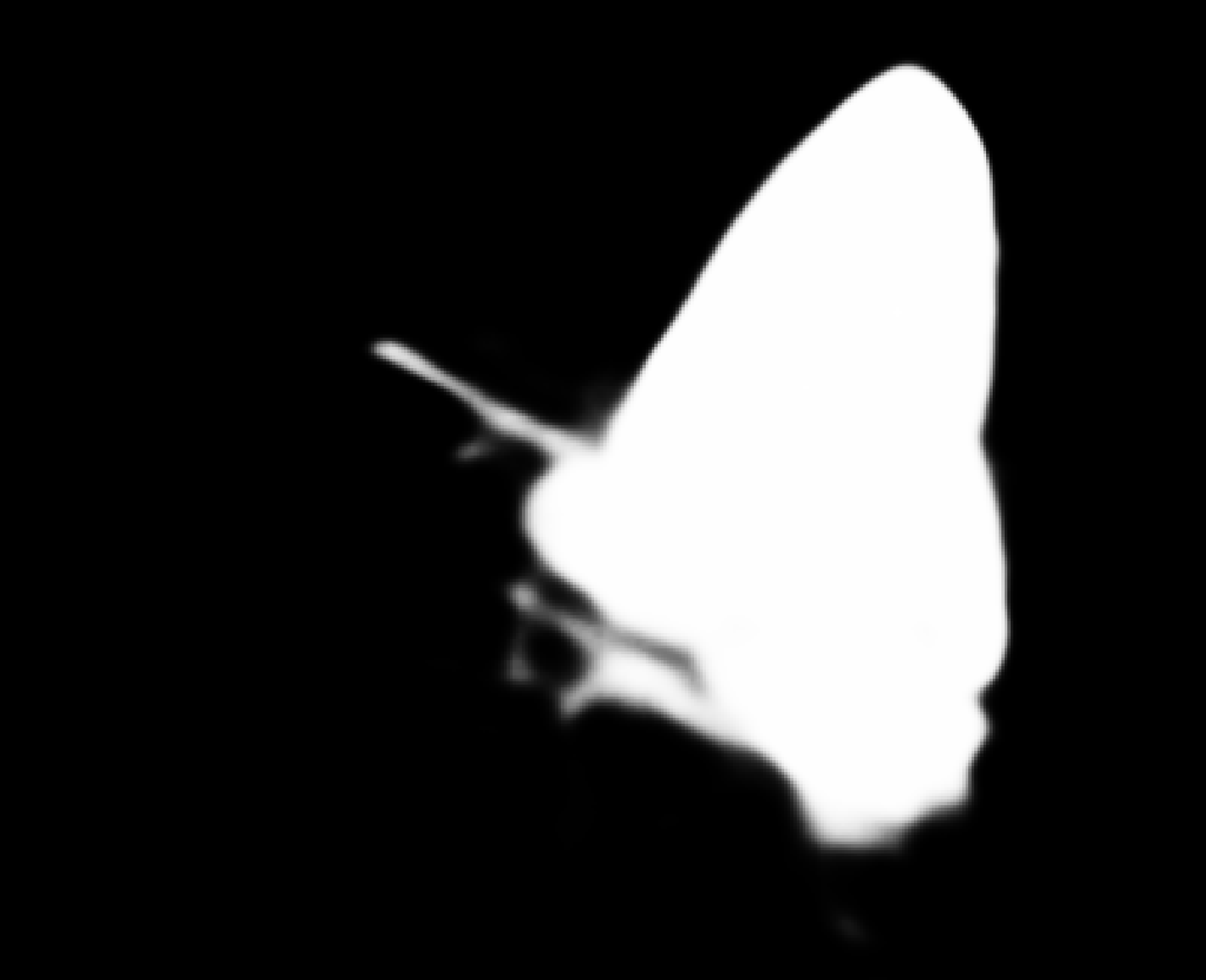} \put (5,63) {\textcolor{white}{\footnotesize.0181}}\end{overpic}
\includegraphics[width=\linewidth]{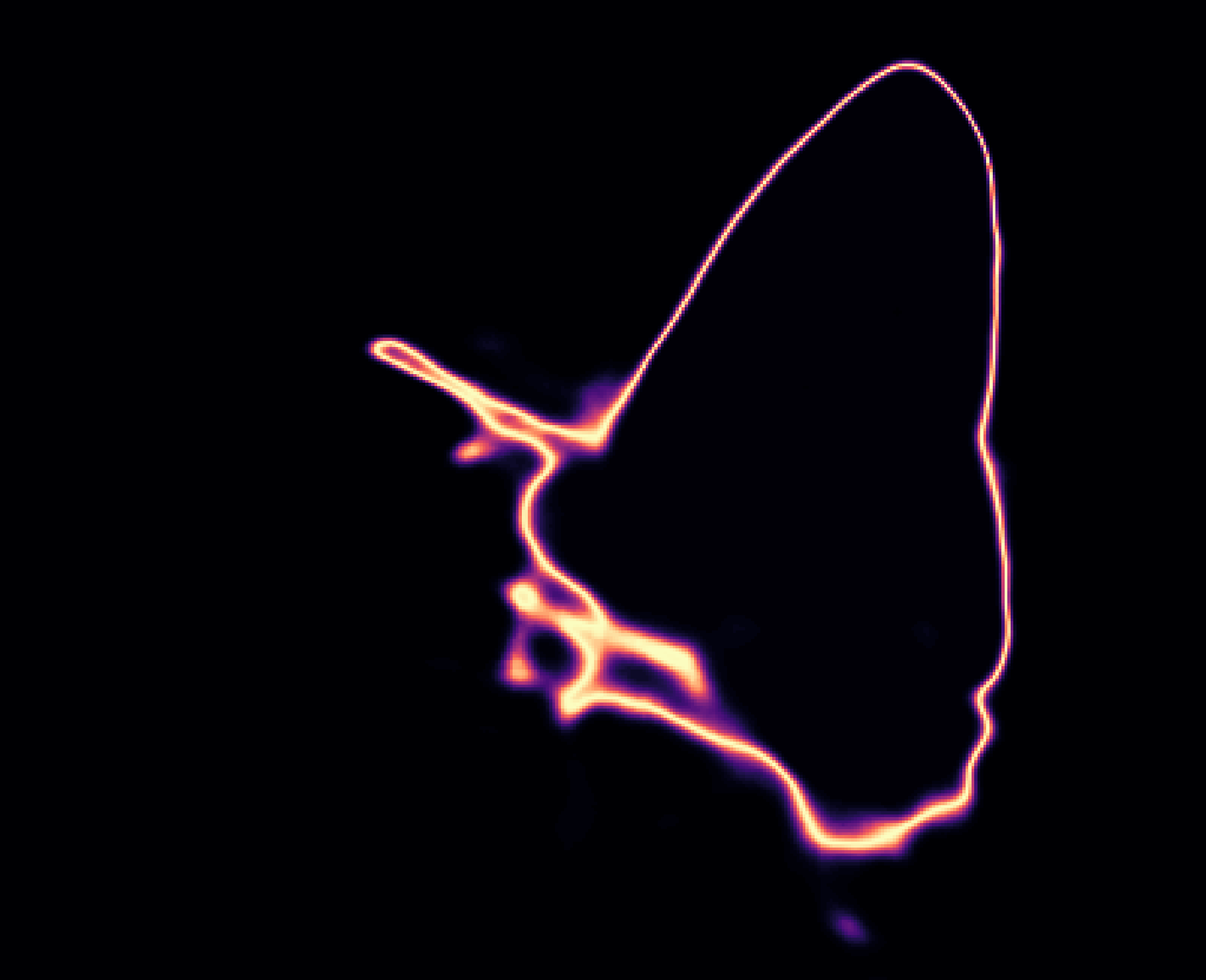}
\begin{overpic}[width=\linewidth]{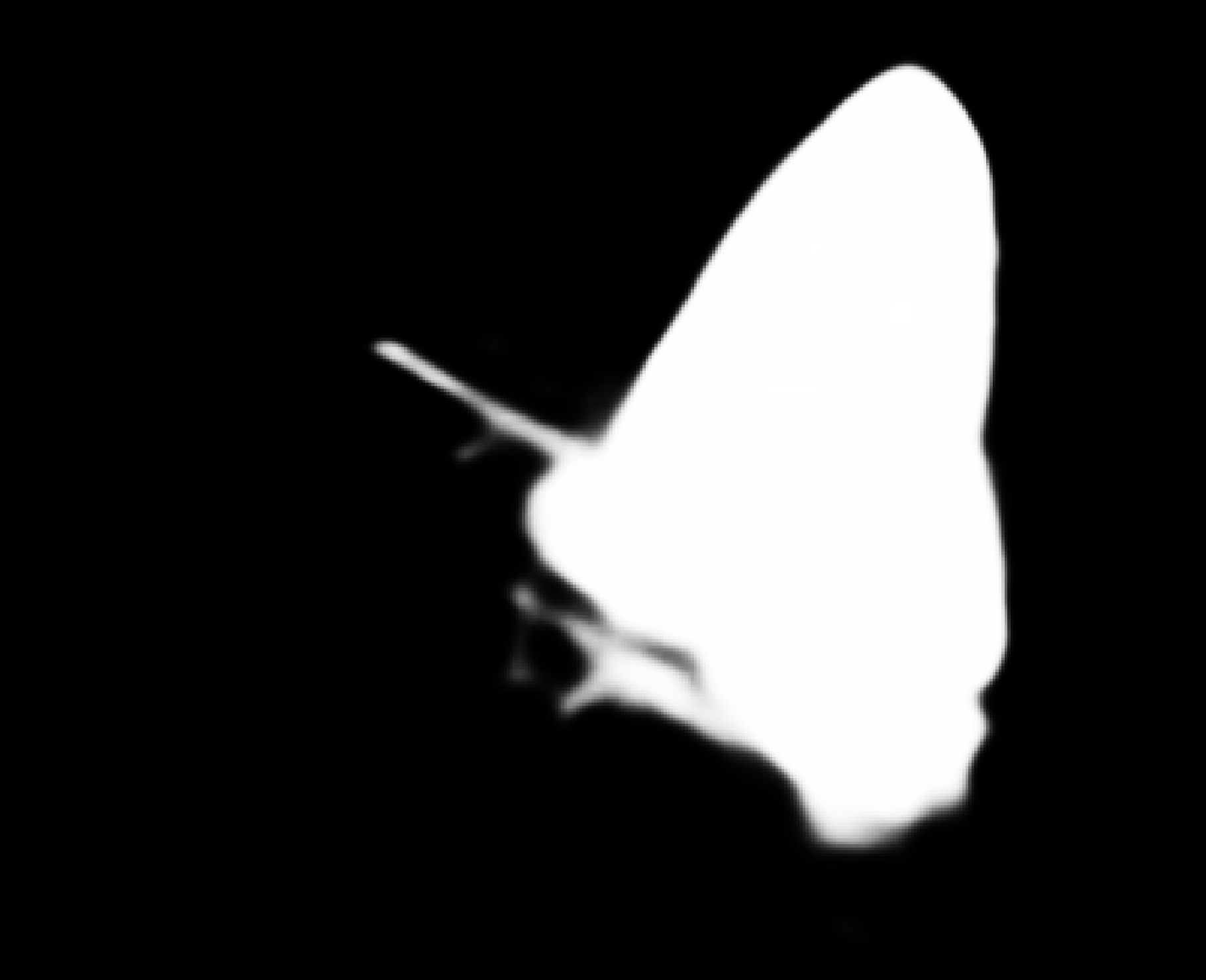} \put (5,63) {\textcolor{white}{\footnotesize.0174}}\end{overpic}
\caption{Stage 3}
\end{subfigure}
\begin{subfigure}{.18\linewidth}
\centering
\begin{overpic}[width=\linewidth]{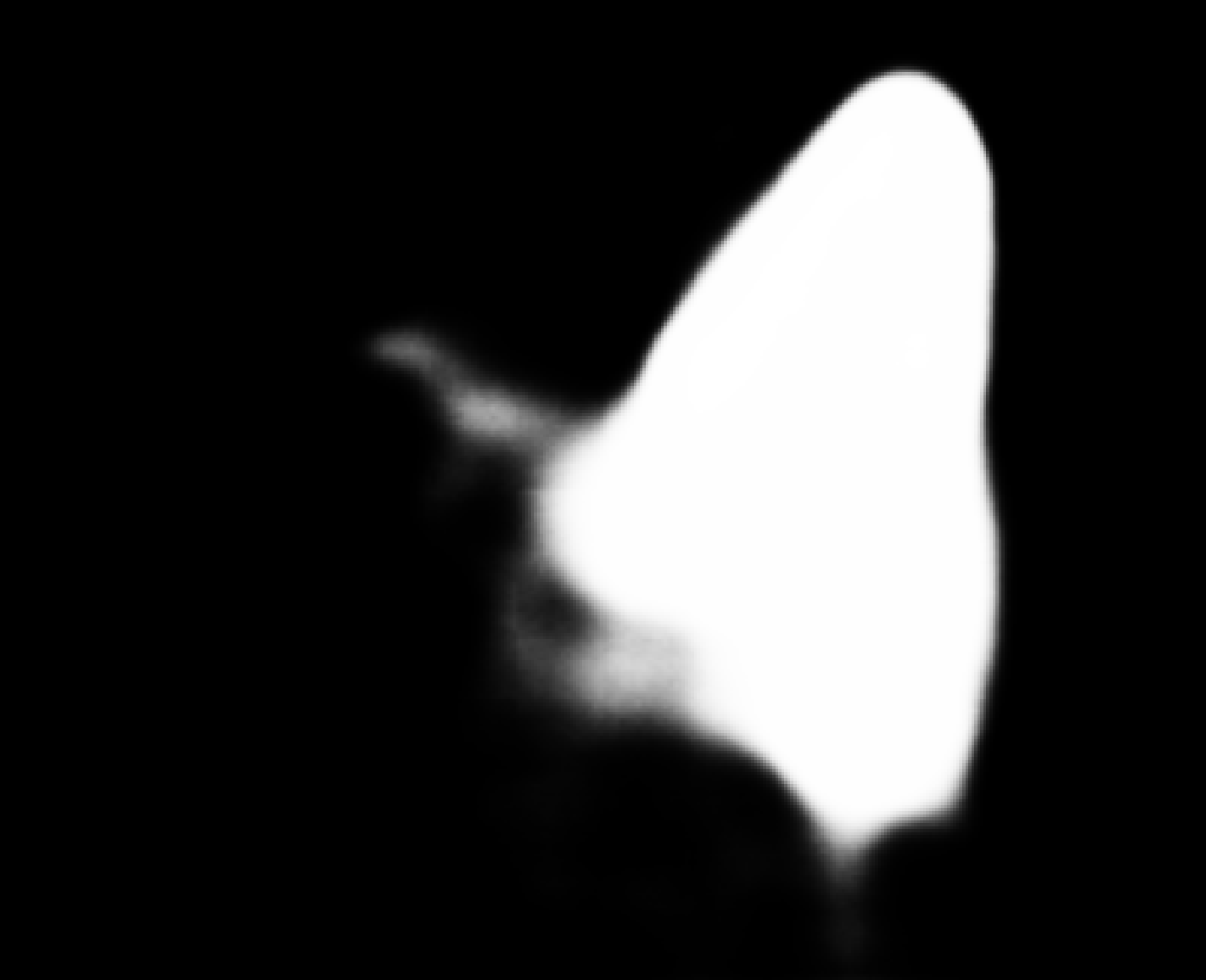} \put (5,63) {\textcolor{white}{\footnotesize.0263}}\end{overpic} 
\includegraphics[width=\linewidth]{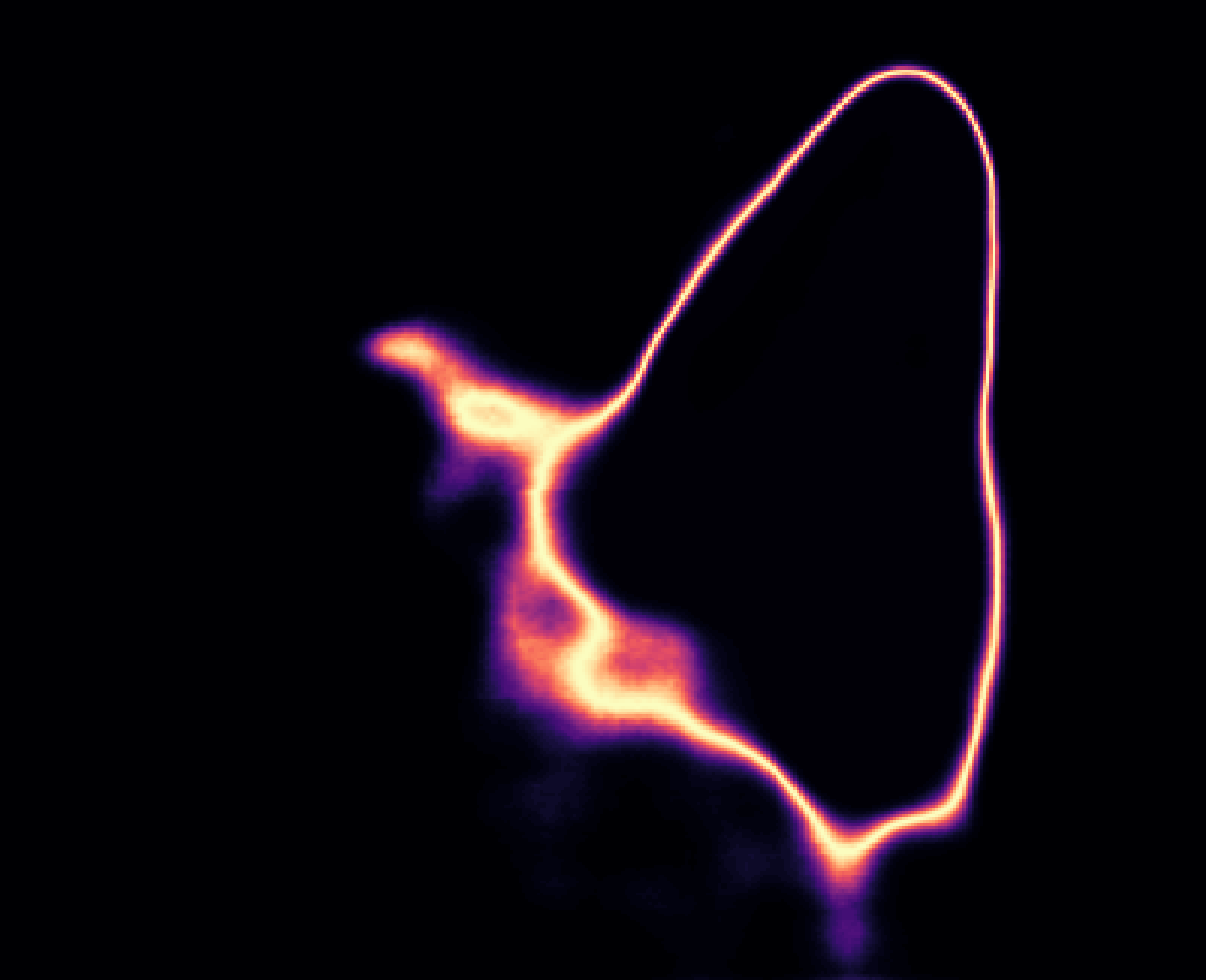}
\begin{overpic}[width=\linewidth]{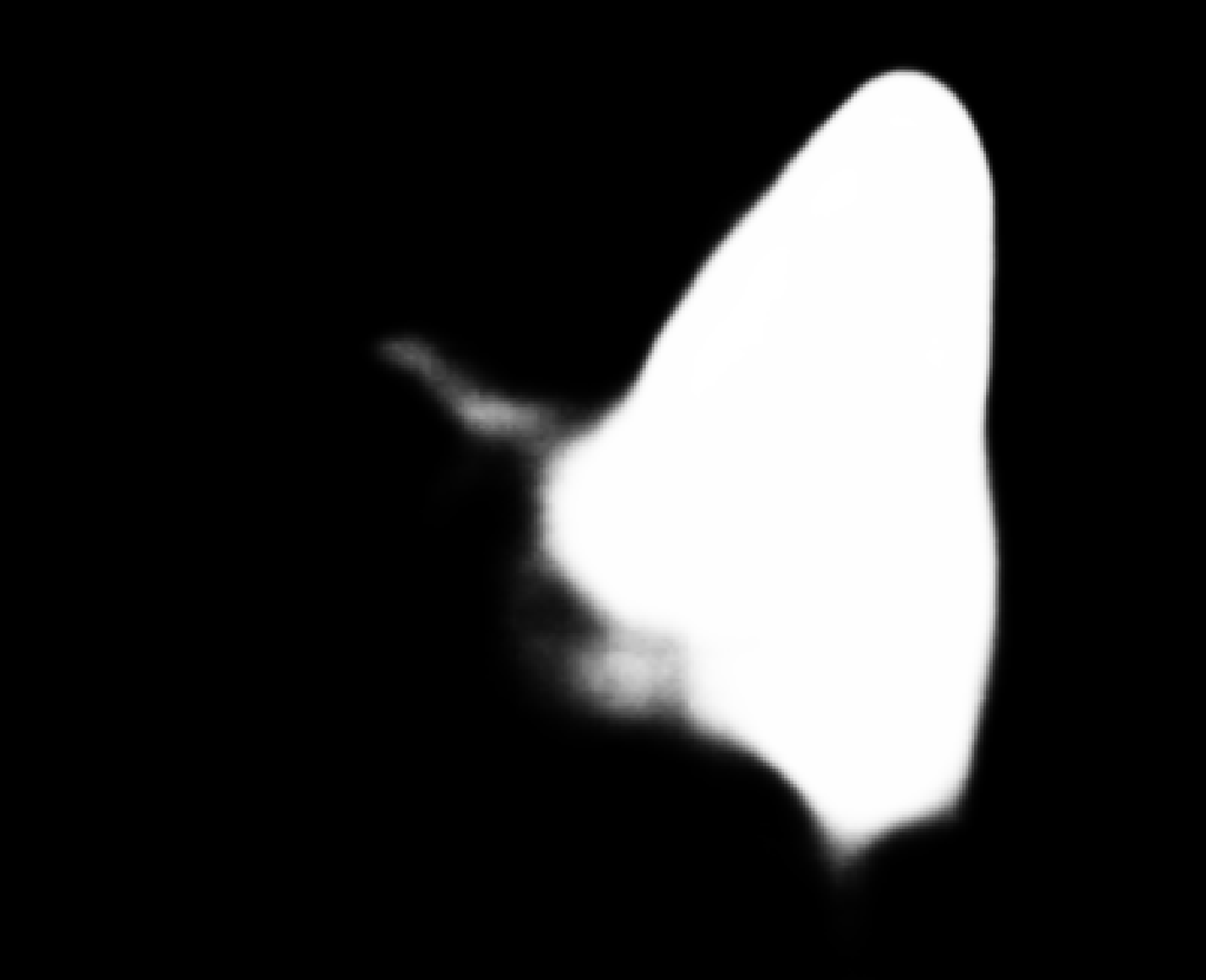} \put (5,63) {\textcolor{white}{\footnotesize.0234}}\end{overpic}
\caption{Stage 2}
\end{subfigure}
\begin{subfigure}{.18\linewidth}
\centering
\begin{overpic}[width=\linewidth]{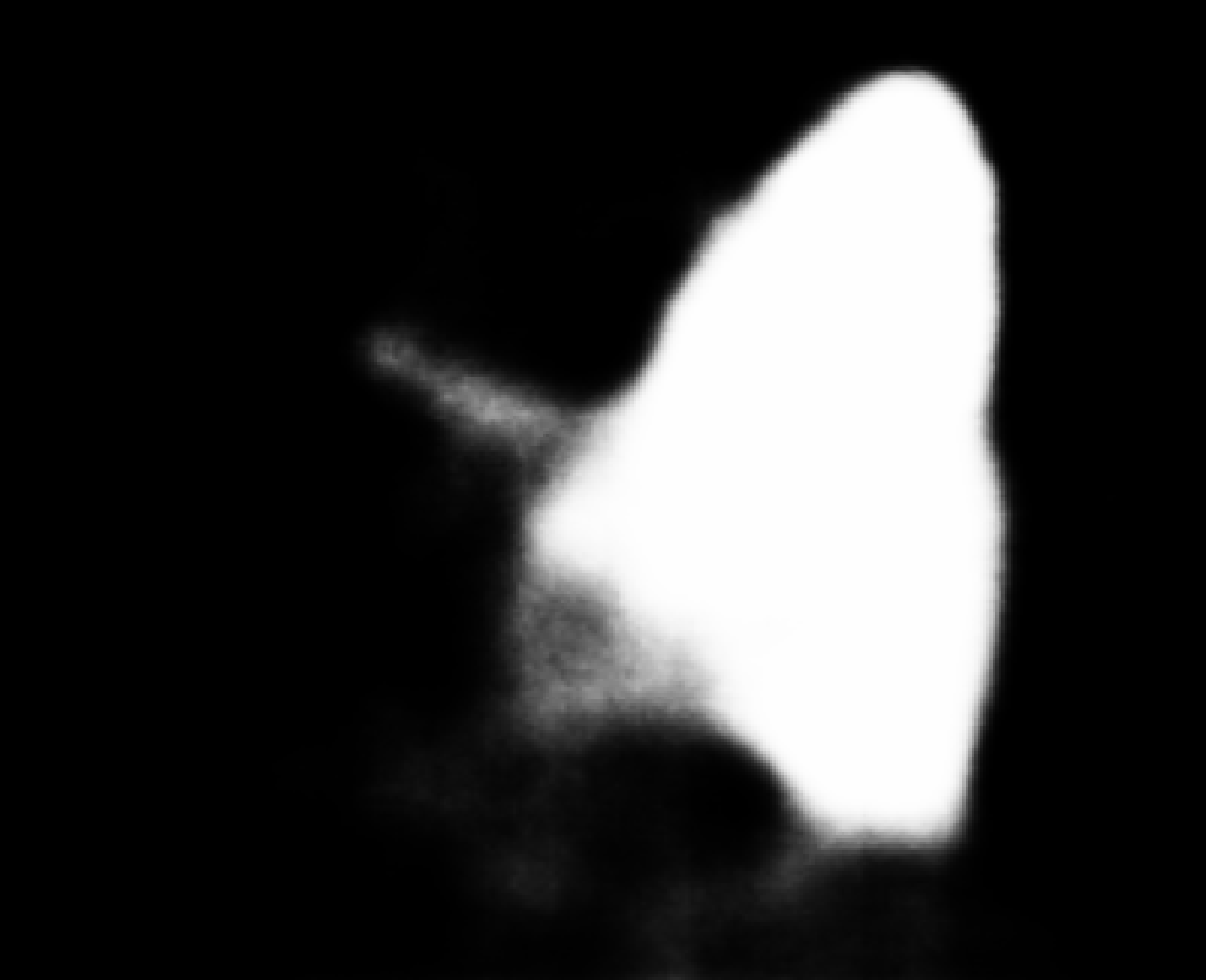} \put (5,63) {\textcolor{white}{\footnotesize.0312}}\end{overpic} 
\includegraphics[width=\linewidth]{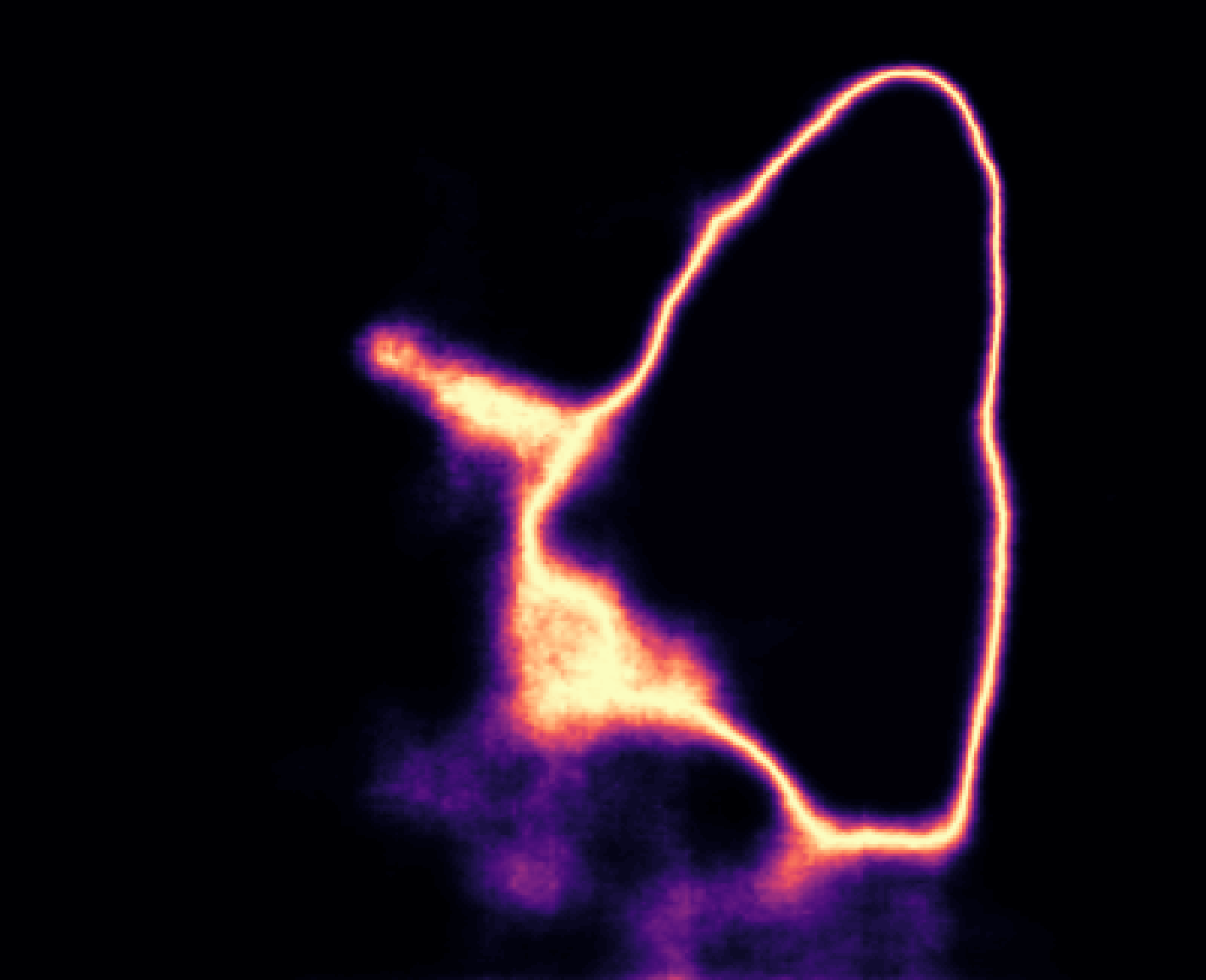}
\begin{overpic}[width=\linewidth]{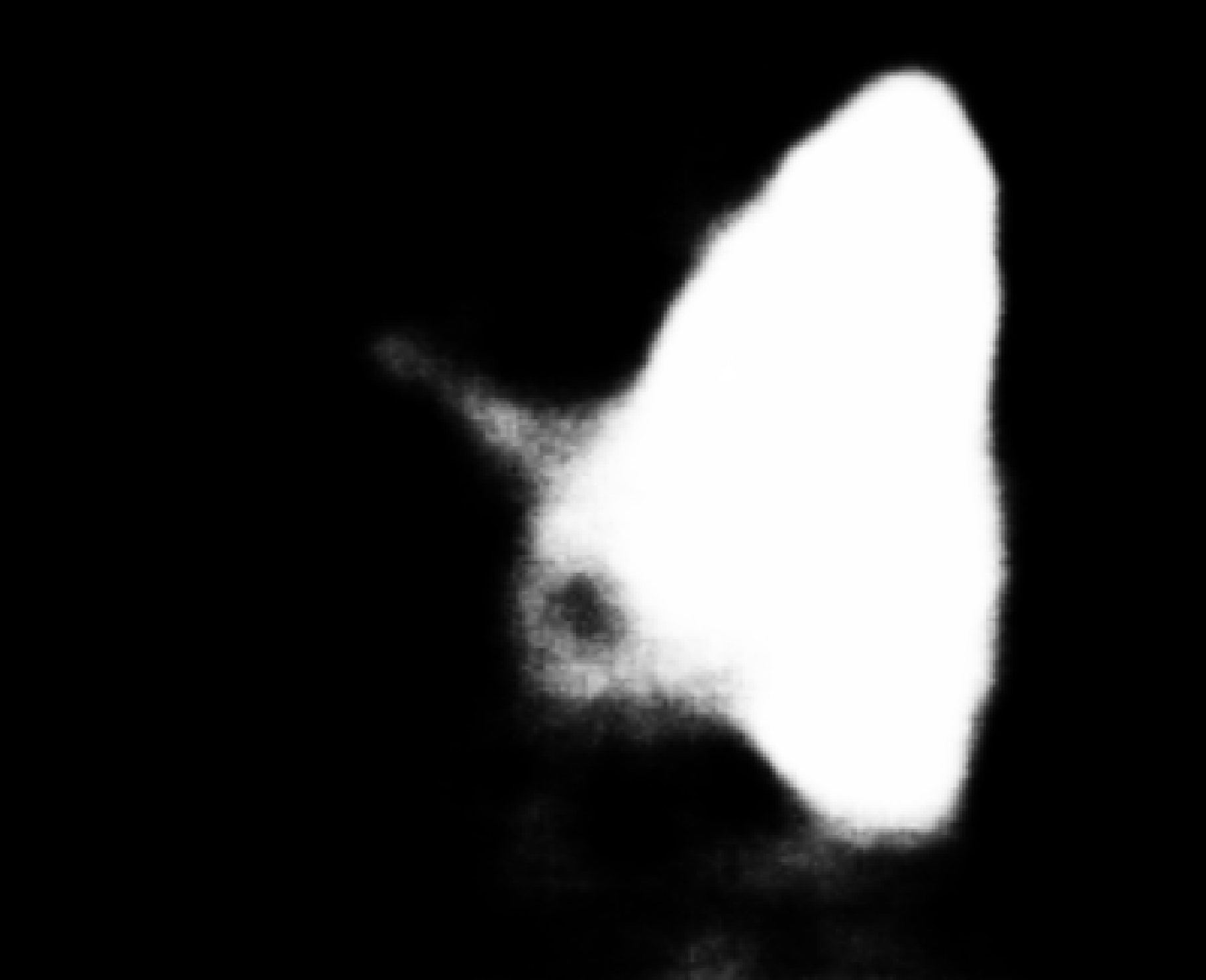} \put (5,63) {\textcolor{white}{\footnotesize.0268}}\end{overpic}
\caption{Stage 1}
\end{subfigure}
\vspace{-0.03\linewidth}
\caption{Qualitative comparisons for CRM. (a) top to bottom: Input, Global-context map, GT, (b)-(e) maps generated from decoder stage 4 to stage 1: first row: first stage predictions, second row: local context maps, last row: second stage predictions. MAE is marked at the top left corner.}
\label{CLC_fig}
\vspace{-0.02\linewidth}
\end{figure}

We compare the two predictions obtained in the CRM visually and quantitatively. Table \ref{table:CLC} reflects the improvement in accuracy between the predictions. In Fig.\ref{CLC_fig}, we can observe that the second prediction is refined by the local context features generated from the first prediction.
\begin{table}[!h]
\centering
\setlength\tabcolsep{2pt}
\caption{Quantitative comparisons for the effectiveness of local-context branch on predictions at decoder stage 2.}
\begin{tabular}{c|cccc|cccc} 
\hline
\hline
\multirow{2}{*}{\begin{tabular}[c]{@{}c@{}}Decoder\\Stage 2\end{tabular}} & \multicolumn{4}{c|}{\textbf{HKU-IS}} & \multicolumn{4}{c}{\textbf{DUT-OMRON}}  \\
& \textbf{$F_{\beta}\hspace{-1.5mm}\uparrow$} & \textbf{$M\hspace{-1.5mm}\downarrow$} & \textbf{$E_{\xi}\hspace{-1.5mm}\uparrow$} & \textbf{$S_{\alpha}\hspace{-1.5mm}\uparrow$}
& \textbf{$F_{\beta}\hspace{-1.5mm}\uparrow$} & \textbf{$M\hspace{-1.5mm}\downarrow$} & \textbf{$E_{\xi}\hspace{-1.5mm}\uparrow$} & \textbf{$S_{\alpha}\hspace{-1.5mm}\uparrow$}\\ 
\hline
First Stage        & .926           & .035           & .945           & .914                 & .817           & .051           & .866            & .847          \\
Second Stage   &\textbf{.927}&\textbf{.033}&\textbf{.947}&\textbf{.915}      &\textbf{.818}&\textbf{.049}&\textbf{.872}&\textbf{.848}    \\
\hline
\hline
\end{tabular}
\label{table:CLC}
\vspace{-0.04\linewidth}
\end{table}

\subsection{Effectiveness of Pixel Shuffle}
\begin{figure}[!h]
\centering
\begin{subfigure}{.15\linewidth}
\centering
\includegraphics[width=\linewidth]{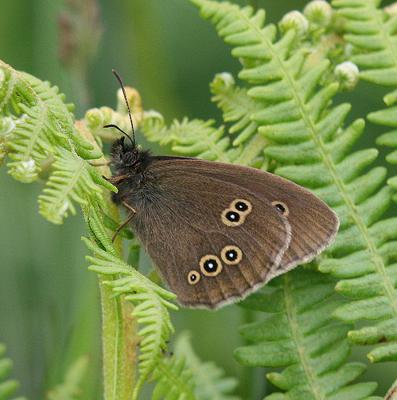}
\includegraphics[width=\linewidth]{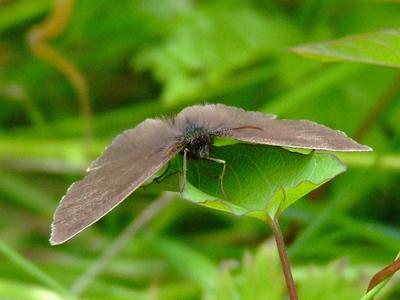}
\end{subfigure}
\begin{subfigure}{.15\linewidth}
\centering
\includegraphics[width=\linewidth]{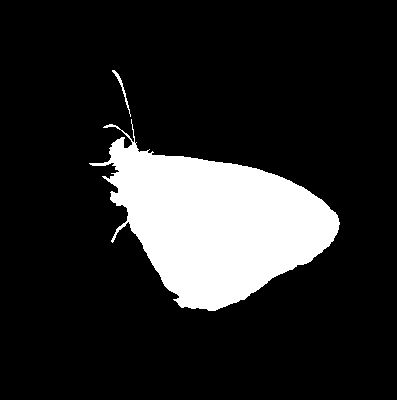}
\includegraphics[width=\linewidth]{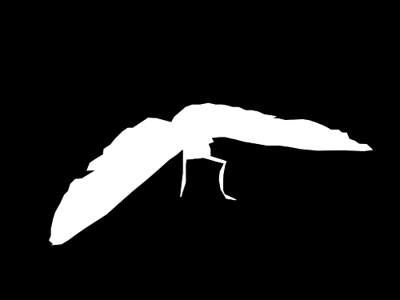}
\end{subfigure}
\begin{subfigure}{.15\linewidth}
\centering
\includegraphics[width=\linewidth]{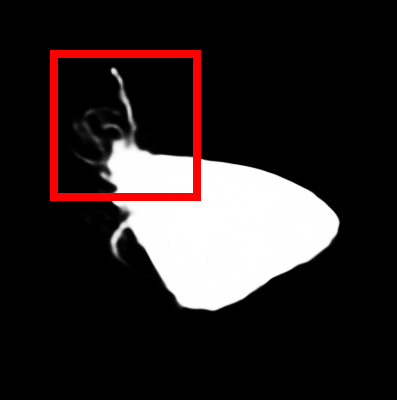}
\includegraphics[width=\linewidth]{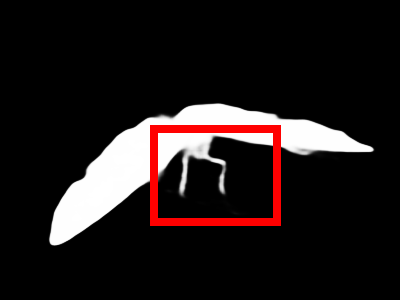}
\end{subfigure}
\begin{subfigure}{.15\linewidth}
\centering
\includegraphics[width=\linewidth]{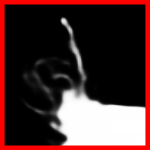}
\includegraphics[width=\linewidth]{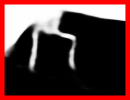}
\end{subfigure}
\begin{subfigure}{.15\linewidth}
\centering
\includegraphics[width=\linewidth]{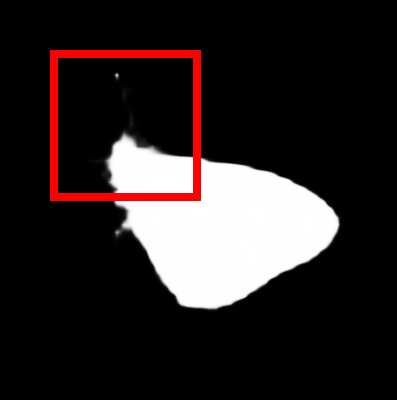}
\includegraphics[width=\linewidth]{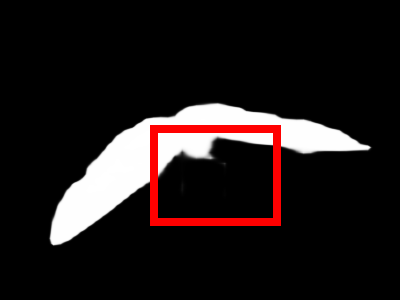}
\end{subfigure}
\begin{subfigure}{.15\linewidth}
\centering
\includegraphics[width=\linewidth]{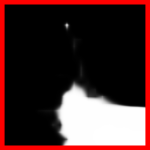}
\includegraphics[width=\linewidth]{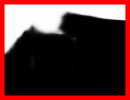}
\end{subfigure}
\begin{minipage}{\linewidth}
{\footnotesize
\ \ \ \ \ \ Image
\ \ \ \ \ \ \ \ GT
\ \ \ \ \ \ \ \ \ \ \ \ PS
\ \ \ \ PS zoomed
\ \ \ \ \ BI
\ \ \ \ \ BI zoomed
}
\end{minipage}
\vspace{-0.02\linewidth}
\caption{Qualitative comparisons of predictions when using BI and PS during training. Pixel Shuffle is effective in restoring more details in the salient object prediction.}
\label{PS_ablation}
\vspace{-0.06\linewidth}
\end{figure}

We replace Pixel Shuffle (PS) with bilinear interpolation (BI) in our network to compare the difference. Better details are restored in the predictions when using PS during the training, as shwon in Fig.\ref{PS_ablation}.

\section{Conclusion}
In this work, we have proposed a novel Transformer-based network named SelfReformer which can guide itself with global and local contexts. In order to obtain a better global context, we framed a supervised patch-wise saliency detection task to obtain the global fature explicitly. Meanwhile, since interpolation or pooling methods damage fine features in the ground truth, we adopted Pixel Shuffle as the up/downsampling method for details preservation. Besides, we developed CRM to guide the decoder with global context information and generate a local context map for better details in predictions. The proposed network demonstrated excellent performance in locating salient objects accurately with rich fine features. Evaluation results indicate the SelfReformer achieved the state-of-the-art across five benchmark datasets in all four related evaluation metrics.


\bibliography{reference}

\end{document}